\makeatletter
\let\my@xfloat\@xfloat
\makeatother

\documentclass[tikz]{uicthesi}
\makeatletter
\def\@xfloat#1[#2]{
	\my@xfloat#1[#2]%
	\def\baselinestretch{1}%
	\@normalsize \normalsize
}
\makeatother

\usepackage[hyphens]{url}

\usepackage{subfigure} 
\usepackage[bookmarks=false]{hyperref}
\usepackage{rotating} 
\usepackage{comment}
\usepackage{array}
\usepackage{amsthm,amsfonts,dsfont}
\usepackage{adjustbox}
\usepackage{bm}
\usepackage{stmaryrd}
\usepackage{xcolor}
\newcommand{\eg}[0]{\textit{e.g.}}

\theoremstyle{definition}
\newtheorem{definition}{Definition}

\usepackage{hyperref}
\usepackage{algorithmicx}
\usepackage{algpseudocode}
\usepackage[misc,geometry]{ifsym}
\usepackage{adjustbox}
\usepackage{stmaryrd}
\usepackage{mathrsfs}
\usepackage{pdfpages}
\usepackage{textcomp}
\usepackage{booktabs,amsmath,epsfig,multirow,CJKutf8,siunitx,balance,dblfloatfix}
\usepackage{color,soul,soulpos,adjustbox,pgfplots}
\usepackage{latexsym,graphicx,amssymb}
\usepackage{subfigure}
\usepackage[font={footnotesize}]{caption}
\usepackage{algorithm}
\usepackage{algpseudocode}
\usepackage{arydshln}
\usepackage{subfigure}
\usepackage{lmodern}

\graphicspath{{./figures/}}

\makeatletter
\newcommand\xleftrightarrow[2][]{%
  \ext@arrow 9999{\longleftrightarrowfill@}{#1}{#2}}
\newcommand\longleftrightarrowfill@{%
  \arrowfill@\leftarrow\relbar\rightarrow}
\makeatother

\begin{document}

\title{Structured Knowledge Discovery from Massive Text Corpus}
\author{CHENWEI ZHANG}
\pdegrees{B.E., Southwest University, 2014}
\degree{Doctor of Philosophy in Computer Science}
\committee{\qquad\qquad Professor Philip S. Yu, Chair and Advisor \\ 
    \qquad\qquad Professor Bing Liu \\
    \qquad\qquad Professor Piotr Gmytrasiewicz \\
    \qquad\qquad Professor Caragea Cornelia \\
    \qquad\qquad Professor Jiawei Zhang, Department of Computer Science, Florida State University}
    
\maketitle

\dedication
{\null\vfil
{\large
\begin{center}
This dissertation is dedicated to my parents and grandparents, \\\vspace{12pt}
for their unconditional love and support.
\end{center}}
\vfil\null}

\acknowledgements
{
\vspace{-0.2in}
\small{
First and foremost, I would like to express my sincere gratitude to my Ph.D. advisor, Prof. Philip S. Yu, for his guidance and support throughout my Ph.D. study and research. 
It has been my privilege to work with you at different aspects of my Ph.D. journey. Your invaluable suggestions, guidance and your passion for research not only help me with my past academic achievements but also will influence my professional career in the future.

Besides my advisor, I would like to thank the rest of my thesis committee: Prof. Bing Liu, Prof. Piotr Gmytrasiewicz, Prof. Cornelia Caragea, and Prof. Jiawei Zhang, for your valuable time serving as my dissertation committee members.

I am grateful to Prof. Yong Deng at Southwest University, for the mentorship during my early research career and enlightening me the first glance of research.
My sincere thank goes to Dr. Wei Fan, who mentored me when I was a research intern at Baidu Research and Tencent America. Without your continuous support, this dissertation would not have been possible. I would like to thank Dr. Nan Du and Dr. Yaliang Li for invaluable suggestions and fruitful discussions during our collaborations in various research projects, which relate to this dissertation.

I would like to express gratitude to my fellow lab mates in the Big Data and Social Computing Lab at the University of Illinois at Chicago, for the stimulating discussions, for the sleepless nights we were working together before deadlines, and for all the fun we have had in the last five years. My warmest thanks extend to all the collaborators, colleagues and friends that I met at the University of Illinois at Chicago.

Last but not least, none of this could have happened without my family. I am grateful for my parents and grandparents, for their unconditional love and encouragement.}
\begin{flushright}
CZ
\end{flushright}
}

\contributionofauthors
Chapter \ref{chapter:c1} is an introduction that outlines my dissertation research. 

Chapter \ref{chapter:c2} presents published papers \cite{zhang2016mining,zhang2017bringing} for which I was the primary author. Dr. Nan Du, Dr. Wei Fan, Dr. Yaliang Li, Dr. Chun-Ta Lu, and Prof. Philip S. Yu contributed to discussions with respect to the work and revising the manuscript. 

Chapter \ref{chapter:c3} presents a published paper \cite{zhang2018joint}, for which I was the primary author. Dr. Yaliang Li, Dr. Nan Du, Dr. Wei Fan, and Prof. Philip S. Yu contributed to discussions with respect to the work and revising the manuscript.

Chapter \ref{chapter:c4} presents a published paper \cite{zhang2018generative} for which I was the primary author. Dr. Yaliang Li, Dr. Nan Du, Dr. Wei Fan, and Prof. Philip S. Yu contributed to discussions with respect to the work and revising the manuscript.

Chapter \ref{chapter:c5} presents a published paper \cite{zhang2018synonymnet} for which I was the primary author. Dr. Yaliang Li, Dr. Nan Du, Dr. Wei Fan, and Prof. Philip S. Yu contributed to discussions with respect to the work and revising the manuscript.

Chapter \ref{chaper:conclusion} concludes this dissertation.
\tableofcontents
\listoftables
\listoffigures

\newcommand{\TaskNameOne}{Structured Intent Detection for Natural Language Understanding}
\newcommand{\TaskNameTwo}{Structure-aware Natural Language Modeling}
\newcommand{\TaskNameThree}{Generative Structured Knowledge Expansion}
\newcommand{\TaskNameFour}{Synonym Refinement on Structured Knowledge}
 
\summary
Nowadays, with the booming development of the Internet, people benefit from its convenience due to its open and sharing nature. 
A large volume of natural language texts is being generated by users in various forms, such as search queries, documents, and social media posts. 
As the unstructured text corpus is usually noisy and messy, it becomes imperative to correctly identify and accurately annotate structured information in order to obtain meaningful insights or better understand unstructured texts. On the other hand, the existing structured information, which embodies our knowledge such as entity or concept relations, often suffers from incompleteness or quality-related issues. Given a gigantic collection of texts which offers rich semantic information, it is also important to harness the massiveness of the unannotated text corpus to expand and refine existing structured knowledge with fewer annotation efforts.

In this dissertation, I will introduce principles, models, and algorithms for effective structured knowledge discovery from the massive text corpus. We are generally interested in obtaining insights and better understanding unstructured texts with the help of structured annotations or by structure-aware modeling. Also, given the existing structured knowledge, we are interested in expanding its scale and improving its quality harnessing the massiveness of the text corpus.
In particular, four problems are studied in this dissertation: \TaskNameOne, \TaskNameTwo, \TaskNameThree, and \TaskNameFour.

\chapter{Introduction} \label{chapter:c1}
\section{Dissertation Outline}
Nowadays, with the booming development of the Internet, people benefit from its convenience due to its open and sharing nature. 
A wide range of user goals is fulfilled on the Internet through various forms of interactions such as web search, web chats, social media postings and so on. The abundant text corpus that is available online embodies rich knowledge that is to be discovered.
Due to the open, sharing nature of the Internet and different linguistic preferences of individuals, the gigantic collection of unstructured text corpus is usually noisy and messy. It is challenging yet rewarding to correctly identify and accurately annotate structured information in order to obtain meaningful insights or better understand the massive unstructured texts.

The structured information summarizes our existing knowledge in a structured manner, which is ubiquitously accessible for both machine and human beings. We may introduce triplets that contain factual relationships among entities as the structured information in knowledge graphs. For example, \texttt{Barack Obama} as an entity has a semantic relation \texttt{president of} with another entity \texttt{U.S.A}. The structured information could be also on the concept level, where the triplets introduce semantic relationships between concepts. For example, we may have \texttt{medicine} as a concept with a relation \texttt{cure} to another concept \texttt{disease}. 
Besides that, the structured information can contain both entity and concept level information in a hierarchical structure, such as \texttt{Barack Obama} as an entity may connect to \texttt{Politician} as a concept. Many researchers in academia and industry are striving to obtain high-quality structured knowledge, such as WordNet \cite{miller1995wordnet}, Yago \cite{fabian2007yago}, Freebase \cite{bollacker2008freebase}, ConceptNet \cite{speer2012representing}, and SenticNet \cite{cambria2018senticnet}.

However, the existing structured knowledge often suffers from incompleteness and quality-related issues. As obtaining high-quality structured information for knowledge discovery is usually time-consuming and labor-intensive, it is thus important to automatically expand and refine the structured information exploiting the massiveness of unannotated text corpus.

The contributions of this dissertation are made toward two strongly correlated, synergistic objectives:
\begin{itemize}
    \item \textbf{Utilizing Structured Information for Natural Language Understanding and Modeling}: Given a massive unannotated text corpus, we are interested in obtaining insights, understanding and modeling the texts with the help of existing structured annotations or by structure-aware modeling.
    \item \textbf{Expanding and Refining Structured Knowledge Harnessing the Massiveness of the Text Corpus}: Given the existing structured knowledge, we are interested in expanding the scale and improving the quality of structured knowledge, where additional human annotation efforts are minimized via harnessing the massive collection of the unannotated text corpus.
\end{itemize}

In particular, four problems are studied in this dissertation: \TaskNameOne, \TaskNameTwo, \TaskNameThree, and \TaskNameFour.
\begin{itemize}
\item To better understand complicated user intents from their diversely expressed natural language utterances, we utilize concept-level structured knowledge and treat intent detection on unannotated text utterances as a structured prediction problem.
\item To extract both word-level and sentence-level semantics while preserving their structural relationships, we provide a structure-aware approach that jointly annotates word-level concept mentions and sentence-level intent labels for each utterance.
\item To expand the scale of high-quality structured knowledge and reduce data preparation efforts, we introduce a generative modeling approach that harnesses word-level semantics learned from the massive text corpus for structured knowledge expansion.
\item To improve the quality of the existing structured knowledge, we refine it by removing synonymous entities. We introduce a framework that detects entity synonyms by comparing among contexts in which entities are mentioned from a massive text corpus.
\end{itemize}

\section{\TaskNameOne}
(Part of this chapter was previously published in~\cite{zhang2016mining,zhang2017bringing}.)

Unstructured texts generated by users in their web search or social media posts are naturally encoded with users' information-seeking intents. To better understand texts generated by users, an intent detection task aims to categorize the text corpus according to intents. Unlike conventional topic classification tasks where the label of the text is highly correlated with some topic-specific concept words, words from different concept categories tend to co-occur in a single piece of information-seeking utterance. When the user tries to express more information in a single piece of utterance, the intent also becomes complicated: the users mention multiple concepts and semantic transitions emerge among multiple concepts.

In Chapter \ref{chapter:c2}, first we formally define the user intent as a semantic transition between two concepts. For complicated utterances, 
we further utilize a concept-level intent graph and formulate intent detection as a structured prediction problem: a structured intent is defined as a sub-graph over the pre-defined concept-level intent graph where each node represents a concept mention and each directed edge indicates a semantic transition. A multi-task neural network model is proposed: one task extracts concept mentions, and the other task infers semantic transitions from the utterance. A customized graph-based mutual transfer loss function is designed to impose explicit constraints over two subtasks for collective inference.

\section{\TaskNameTwo}
(Part of this chapter was previously published in~\cite{zhang2018joint})

Being able to recognize words as slots and detect the intent of an utterance has been a keen issue in natural language understanding. Existing works either treat word-level slot filling and utterance-level intent detection separately in a pipeline manner, or adopt joint models which sequentially label slots while summarizing the utterance-level intent without explicitly preserving the semantic hierarchy among words on the word level, slots on the concept level, and intents on the utterance level. 
In Chapter \ref{chapter:c3}, to exploit the semantic hierarchy for effective natural language modeling, we investigate a structure-aware approach that accomplishes slot filling and intent detection in a bottom-up fashion via a dynamic routing-by-agreement schema. The model does slot filling by learning to assign each word on the word-level to the most appropriate slot on the concept-level via dynamic routing. The dynamic routing also aggregates concept-level slot representations to predict the utterance-level intent. As the intent of the utterance may also help recognize words as different slots, a re-routing schema is proposed that further synergizes the word-level slot filling performance using the inferred utterance-level intent in a top-down fashion. 

\section{\TaskNameThree}
(Part of this chapter was previously published in~\cite{zhang2018generative}.)

When knowledge graph is becoming an indispensable resource that offers rich structured information for numerous knowledge-intensive applications, it often suffers from incompleteness issues. Building a complete, high-quality knowledge graph is time-consuming and requires significant human annotations. Previously, most knowledge graph completion methods use discriminative classifiers that extract triplets directly from corpus where certain relations are expressed. When the knowledge graph is in its infancy, we lack sufficient and high-quality annotations on the text corpus for existing discriminative models to excel.

To reduce human annotation efforts for structured knowledge expansion, in Chapter \ref{chapter:c4} we introduce a generative perspective to increase the scale of high-quality structured knowledge and study the Structured Knowledge Expansion task.
The proposed model explores the generative modeling capacity for entity pairs and harnesses word-level semantics learned from the massive text corpus for structured knowledge expansion. It is able to generate meaningful entity pairs that are not yet observed and efficiently expand the scale of structured knowledge.

\section{\TaskNameFour}
(Part of this chapter was previously published in~\cite{zhang2018synonymnet}.)

Currently, information extraction systems can automatically extract structured knowledge from a large collection of text corpus. However, the task to extract information is challenging and current systems make many mistakes: ambiguous, redundant or conflicting entity information are prevalently observed during the construction of structured knowledge. Given an existing knowledge graph, a lot of human annotation efforts are being made to improve the quality of the extracted knowledge.

To improve the quality of the existing structured knowledge, in Chapter \ref{chapter:c5} we propose to remove duplicated and redundant entity information in an existing knowledge graph. Previous works on detecting synonymous entities focus on learning the similarity between entities using character-level features. These methods work well for synonyms that share a lot of character-level features like \texttt{airplane/aeroplane}. However, a much larger number of synonym entities in the real-world do not share a lot of character-level features, such as \texttt{JD/law degree}. Instead of relying on excessive human annotations, we propose to leverage the free-text contexts in which entities are mentioned in a gigantic collection of text corpus for effective synonym detection. Instead of using entities features, a novel neural network model is proposed which makes use of multiple pieces of contexts in which the entity is mentioned, and compares the context-level similarity via a bilateral matching schema to determine synonymity.


\chapter{\TaskNameOne}\label{chapter:c2}
This chapter was previously published as ``Mining User Intentions from Medical Queries: A Neural Network based Heterogeneous Jointly Modeling Approach'' in WWW'16~\cite{zhang2016mining}, DOI: \url{https://doi.org/10.1145/2872427.2874810}, and ``Bringing Semantic Structures to User Intent Detection in Online Medical Queries'' in BigData'17~\cite{zhang2017bringing}. DOI: \url{https://doi.org/10.1109/BigData.2017.8258025}.
\ulposdef{\hlcyan}[xoffset=1pt]{\mbox{\color{cyan!30}\rule[-.8ex]{\ulwidth}{2.7ex}}}
\ulposdef{\hlred}[xoffset=1pt]{\mbox{\color{red!30}\rule[-.8ex]{\ulwidth}{2.7ex}}}
\ulposdef{\hlyellow}[xoffset=1pt]{\mbox{\color{yellow!30}\rule[-.8ex]{\ulwidth}{2.7ex}}}
\ulposdef{\hlorange}[xoffset=1pt]{\mbox{\color{orange!30}\rule[-.8ex]{\ulwidth}{2.7ex}}}
\ulposdef{\hlgreen}[xoffset=1pt]{\mbox{\color{green!30}\rule[-.8ex]{\ulwidth}{2.7ex}}}

\section{Introduction}
\begin{figure}[htbp]
\centering
\epsfig{file=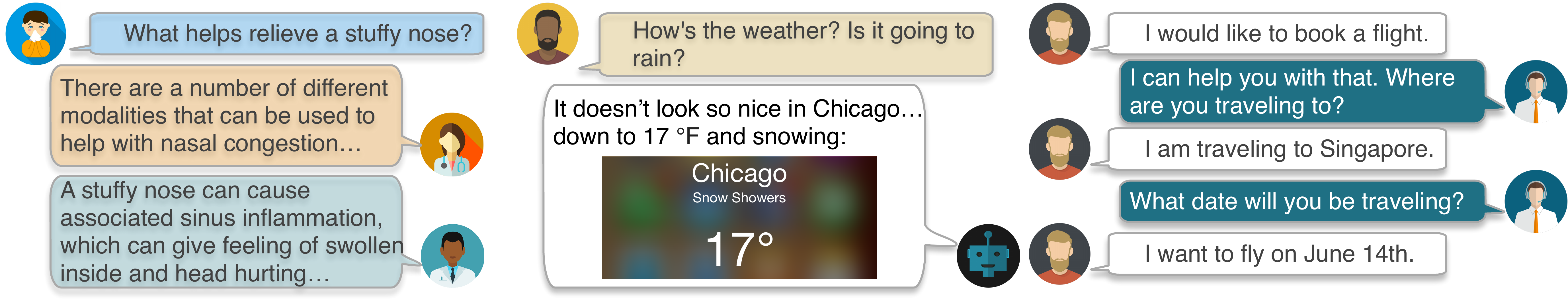,width=6in}
\phantom{0}(a) Community Q\&A \phantom{000000} (b) Voice Assistant\/Chatbot\phantom{00000}(c) Service Center Q\&A\phantom{000000}
\caption{Intent Detection from various types of user-generated utterances.}\label{fig:cases}
\end{figure}

A wide range of user goals is fulfilled on the Internet through various forms of interactions such as web search, web chats and so on. 
For example, online question answering websites are able to offer globally accessible information via human-human interactions. As voice assistants and chat-bots become more and more popular, users may ask smart devices questions via voice commands. In service center question answering systems, customers express their requests and get their tasks resolved. For example, booking a flight with customer service representatives. 
\ref{fig:cases} illustrates three scenarios on community Q\&A, voice assistant/chatbot, and service center Q\&A. 

With various forms of interactions, a huge amount of text corpus are generated by users. The text corpus generated by users, usually consists of declarative statements followed by questions, are naturally encoded with users' information-seeking intentions. An intent detection task tries to model and discover intentions that a user encodes in the text corpus. Unlike conventional text classification tasks where the label of text is highly correlated with some topic-specific words, words from different topic categories tend to co-occur in questions generated by users for information-seeking purposes. Besides the existence of topic-specific words and word order, word correlations and the way words are organized in the corpus are crucial to the intent detection task. Due to different linguistic preferences of individuals, the intentions can be expressed partially, implicitly or diversely, which makes it challenging to accurately understand user intentions from the text corpus. 

\begin{table}[ht]
    \centering
    \begin{tabular}{ll}
Text  &  Intent\\
        \toprule
I have (got) a \hlcyan{fever}, should I take the \hlyellow{Tylenol}? & \textit{\hlcyan{Symptom}}  $\to$ \textit{\hlyellow{Medicine}}\\ 
Which \hlyellow{medicine} should I take if I’m running a \hlcyan{fever}? & \textit{\hlcyan{Symptom}} $\to$ \textit{\hlyellow{Medicine}}\\ 
I've come down with a \hlcyan{fever}, should I take \hlyellow{Aspirin}?  & \textit{\hlcyan{Symptom}}  $\to$ \textit{\hlyellow{Medicine}}\\ 
Is it okay to use \hlyellow{ibuprofen} when I'm running a \hlcyan{temperature}?  & \textit{\hlcyan{Symptom}} $\to$ \textit{\hlyellow{Medicine}}\\ 
My \hlcyan{temperature is 103}, can I use \hlyellow{Advil}?  & \textit{\hlcyan{Symptom}}  $\to$ \textit{\hlyellow{Medicine}}\\ 
  \bottomrule
    \end{tabular}
    \caption{Utterances with the same intent but different expressions.}
    \label{tab::diversified_expressions}
\end{table}

Specifically, the intention we studied in this work is characterized as a directed semantic transition between two concepts: from a concept that is mentioned in declarative statements (\eg~\texttt{Symptom}), to another concept that indicates the user's information need (\eg~\texttt{Medicine}). 
As shown in \ref{tab::diversified_expressions}, each sentence adopts a unique expression but they all share the same intention where users mention symptom concepts and look for related medications.
Moreover, when users try to express more sophisticated information in a single piece of sentence, the semantic transitions also become complicated over multiple concepts, as shown in \ref{tab::medical_queries}.
\begin{table}[ht!]
    \centering
    \begin{tabular}{p{3.5in}l}
        Text  &  Structured Intent\\
        \toprule
        My three-year-old child is \hlcyan{sick with a temperature of 100 degrees} she \hlcyan{can't keep anything down including liquids}. \hlyellow{What kind of medicine} should I give my child, \hlgreen{and how much}? & \hlcyan{Symptom} $\to$ \textit{\hlyellow{Medicine}} $\to$ \textit{\hlgreen{Instruction}}\\ \hline
        Do I have \hlorange{insomnia} if I have \hlcyan{trouble staying asleep}? Any \hlyellow{medication} is recommended to help me \hlcyan{fall asleep} easier? & \textit{\hlorange{Disease}} $\leftarrow$ \textit{\hlcyan{Symptom}} $\to$ \textit{\hlyellow{Medicine}}\\
        \bottomrule
    \end{tabular}
     \caption{Complicated sentences with structured intents.}
    \label{tab::medical_queries}
\end{table}

In this work, we introduce a novel neural network architecture that bring structures to detect complicated user intents in the user-generated text corpus. We observe an appealing property that information-seeking text corpus exhibits a strong coupling between concept mentions and semantic transitions between concepts. 
The proposed model is trained to automatically discover concept mentions and infer semantic transitions from the unstructured text corpus, in contrast to relying on fixed dictionaries for word-concept mapping \cite{chiang2012autodict,godbole2010building,zhang2016mining} or using pre-defined parsing rules \cite{de2010rule} and templates \cite{spink2004study} in prior works. 
A customized graph-based mutual transfer loss function is designed to impose explicit constraints to reduce the conflicts between extracting concept mentions and inferring semantic transitions.
We show that by taking the correlations among concept mentions and semantic transitions into considerations, the proposed model is able to accurately detect complicated user intents from the text corpus.

Experiments are conducted on the text corpus collected from an online question-answering discussion forum. We contrast the performance of the proposed model with other alternatives by an 8\% relative improvement in micro-AUC and an 23\% relative reduction in coverage loss.

\section{Preliminaries}
We now formally define the terminologies and describe the structured intent detection problem for natural language understanding. Also, we provide observations to show appealing coupling properties of concept mentions and semantic transitions in the text corpus \cite{cai2017cnn}, which motivates a graph-based formulation for structured intent.

\subsection{Terminologies}
\begin{definition}[\textbf{Concept}]
Let a concept $c$ be a group or class of objects and/or abstract ideas that share similar fundamental characteristics in a certain domain. $C=\{c_1,c_2,...,c_M\}$ is list of a full spectrum of $M$ concepts in a specific domain. For example, the medical domain contains concepts of diseases, symptoms, medicine and so on. Users can mention concepts in a text corpus by specific object names as explicit mentions (``Tylenol'', ``Ibuprofen'' or ``xxx caplet/capsule/drop/syrup''), or as implicit mentions by abstract ideas (``remedy'' or ``which medication/medicine/drug'').
\end{definition}

\begin{definition}[\textbf{Semantic Transition}]
Let a semantic transition ${t}_{{i}\to{j}}$ defines a transition of a user information-seeking intention from a concept $c_i$ to a concept $c_j$. A semantic transition ${t}_{{i}\to{j}}$ exists in the text corpus when two concepts $c_i$, $c_j\in{C}$ are mentioned (either explicitly or implicitly) with a semantic transition between them. For example, a concept transition $t_{Symptom\to{Medicine}}$ in the healthcare domain usually starts with patients describing their symptoms and asking for related information about medications that help them alleviate their symptoms.

$T$ contains the full spectrum of $N$ semantic transitions in a certain domain, which can be indexed as flat labels $T=\{t_1,t_2,...,t_N\}$ for simplicity instead of $\{t_{i\to{j}}\}$. Those two index notations are used interchangeably in this work. Multiple semantic transitions can co-exist in a single piece of text corpus and the direction of a semantic transition does not necessarily follow the order of concept occurrence in the text corpus. Multiple semantic transitions may follow certain structures such as a chain-like path, like $Symptom{\to}Medicine{\to}Instruction$.
\end{definition}

An intent is defined as a semantic transition between two concepts. Formally, we have:
\begin{definition}[\textbf{Intent}]
Considering a basic case, where each text corpus consists of some declarative sentences followed by questions. For each information-seeking text corpus $Q$, the resulting intent is denoted as a tuple pair $\langle{s, n}\rangle$:
where $s$ is the concept being mentioned in declarative sentences as the information known to the user, while $n$ indicates a concept being mentioned in questions indicating user's information needs. 
\end{definition}

When a user tries to express complicated information needs, a single piece of text corpus may embody multiple intents. We observe that multiple semantic transitions in a single piece of text are often correlated with each other, coupled with some shared concept mentions.

To effectively model complicated semantic transitions among multiple concept mentions, we first define an intent graph that bring structures to concept mentions and semantic transitions.
\begin{definition}[\textbf{Intent Graph}]
Let $G=\langle{C},{T}\rangle$ be an intent graph where each node represents a concept $c_m\in{C}$ and each directed edge $t_{i,j}\in{T}$ be a semantic transition from node $c_i$ to $c_j$. An intent graph $G$ is a graph representation that indicates all possible concept mentions and semantic transitions in a certain domain. Note that the domain-specific intent graph can be obtained from domain experts or constructed as a concept-level graph from large text corpora using existing techniques \cite{hasegawa2004discovering,yan2009unsupervised,zhang2016heer}.
\end{definition}

\begin{definition}[\textbf{Structured Intent}]
Let a Structured Intent ${\hat G}_Q=\langle{{\hat C}_Q, {\hat T}_Q}\rangle$ be a sub-graph of $G=\langle{C},{T}\rangle$, indicating concepts $ {\hat C}_Q\subseteq C$ mentioned by the text corpus $Q$ and semantic transitions ${\hat T}_Q \subseteq T$ inferred from $Q$.
\end{definition}

\begin{figure}[!ht]
\centering
\begin{minipage}[b]{0.45\textwidth}
\epsfig{file=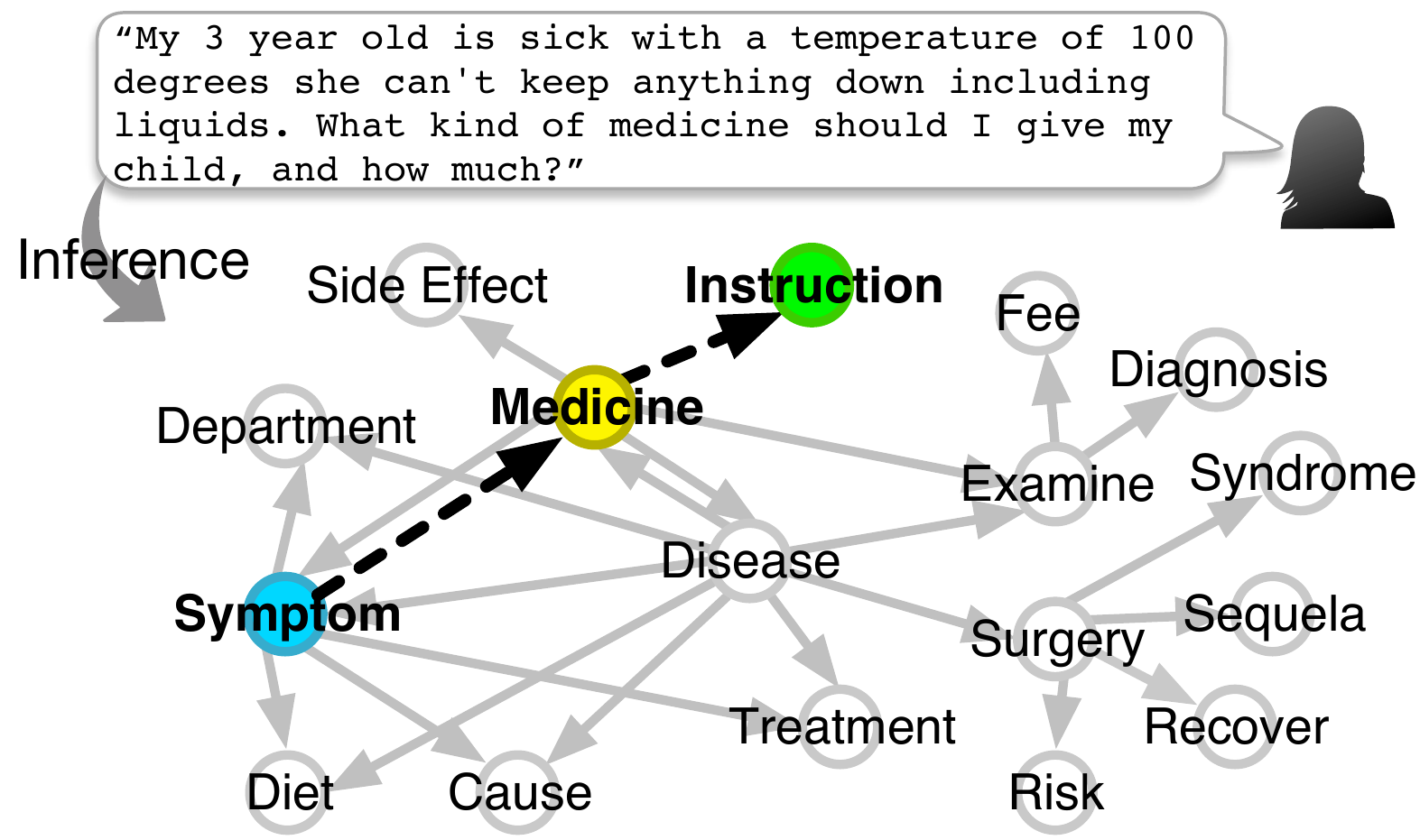, width=3in}
\caption{\TaskNameOne.}
\label{fig::concept_transition_inference_problem}
\end{minipage}
\hfill
\begin{minipage}[b]{0.45\textwidth}
\epsfig{file=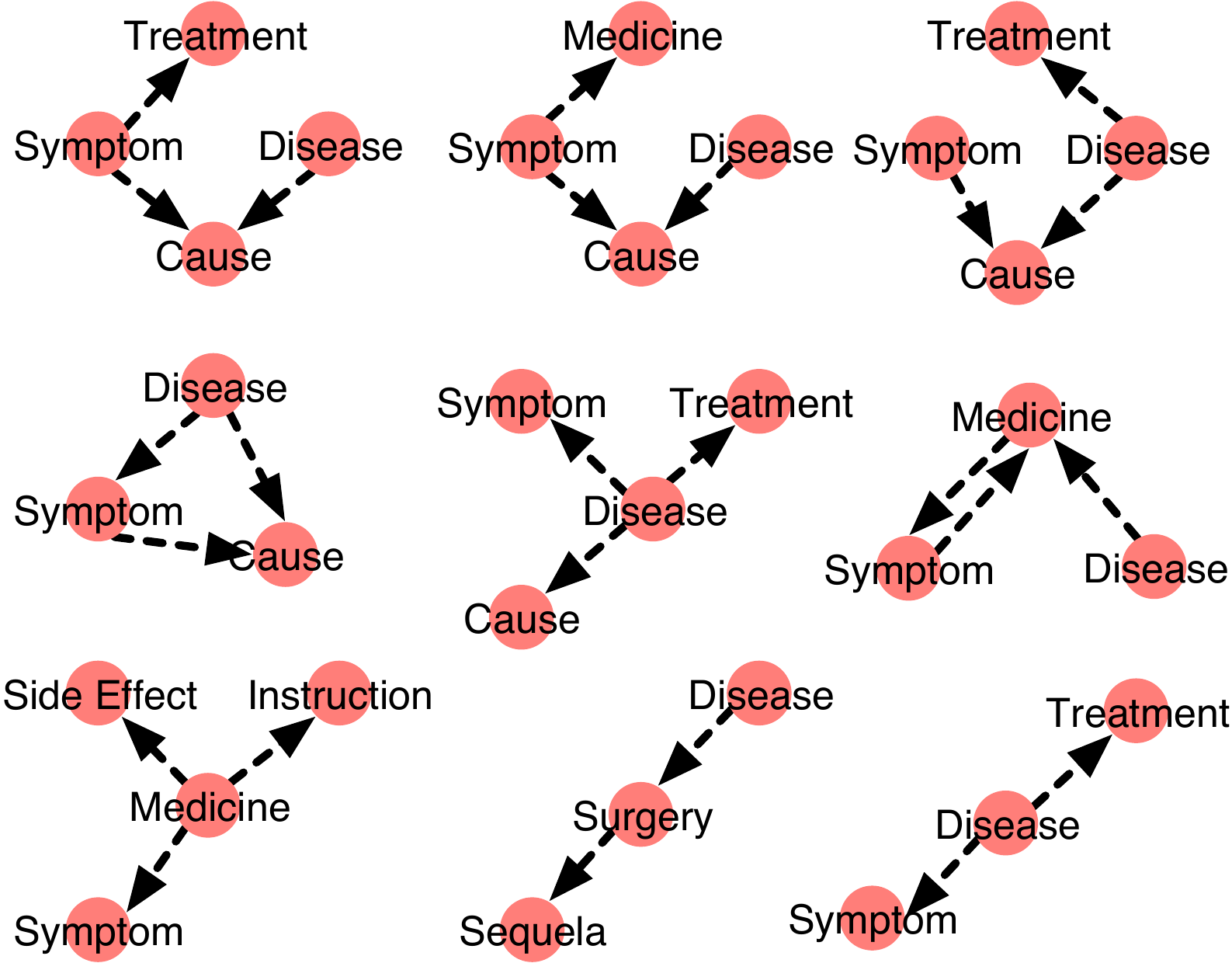, width=2.5in}
\caption{Frequent structured intents.}\label{fig::top_concept_transitions}
\end{minipage}
\end{figure}

\subsection{Problem Statement}
\begin{definition}[\textbf{\TaskNameOne}]
Given 1) an information-seeking text corpus $Q$ which consists of $K$ elements $\{q_1, q_2, ..., q_K\}$, where each element is a word or a phrase and 2) an intent graph $G=\langle C, T\rangle$, where $C$ denotes all possible concept mentions and $T$ indicates all possible semantic transitions, the Structured Intent Detection problem tries to effectively infer the Structured Intent $\hat{G}_Q=\langle \hat{C}_Q,\hat{T}_Q\rangle$ as a sub-graph of the intent graph $G$, where $ {\hat C}_Q\subseteq C$ and ${\hat T}_Q \subseteq T$. \ref{fig::concept_transition_inference_problem} illustrates this idea, where ${\hat C}_Q$ are shown as colored nodes and ${\hat T}_Q$ are shown as black arrows with dashed lines.
\end{definition}

\subsection{Observations}\label{sec::ob}
We sample 10,000 pieces of text corpus from an online medical question answering discussion forum and label them with structured intents. We end up having 17 unique types of concepts and 23 unique types of semantic transitions (Details in Section \ref{sec::data}).

We show the top-9 frequent structured intents being annotated, as shown in \ref{fig::top_concept_transitions}. By characterizing complicated user intentions with a graph-based formulation, the Structured Intent Detection task maps each text corpus with diverse expressions into a graph structure that indicates users' information needs in a clear and structural way.

More importantly, we found that the Structured Intents rarely have disconnected components, from a perspective of the graph theory. This not only shows that users tend to express multiple semantic transitions in a single piece of information-seeking text corpus but also indicates that multiple semantic transitions in a single piece of text corpus are expressed and developed together, coupled with some shared concept mentions.
In summary, the connectivity patterns of frequent Structured Intents further imply that by taking advantages of the semantic structure, the correlations between nodes and edges in the Structured Intent can be jointly inferred with a synergistic effect.
\section{Proposed Approach}
\begin{figure}[ht!]
\centering
\epsfig{file=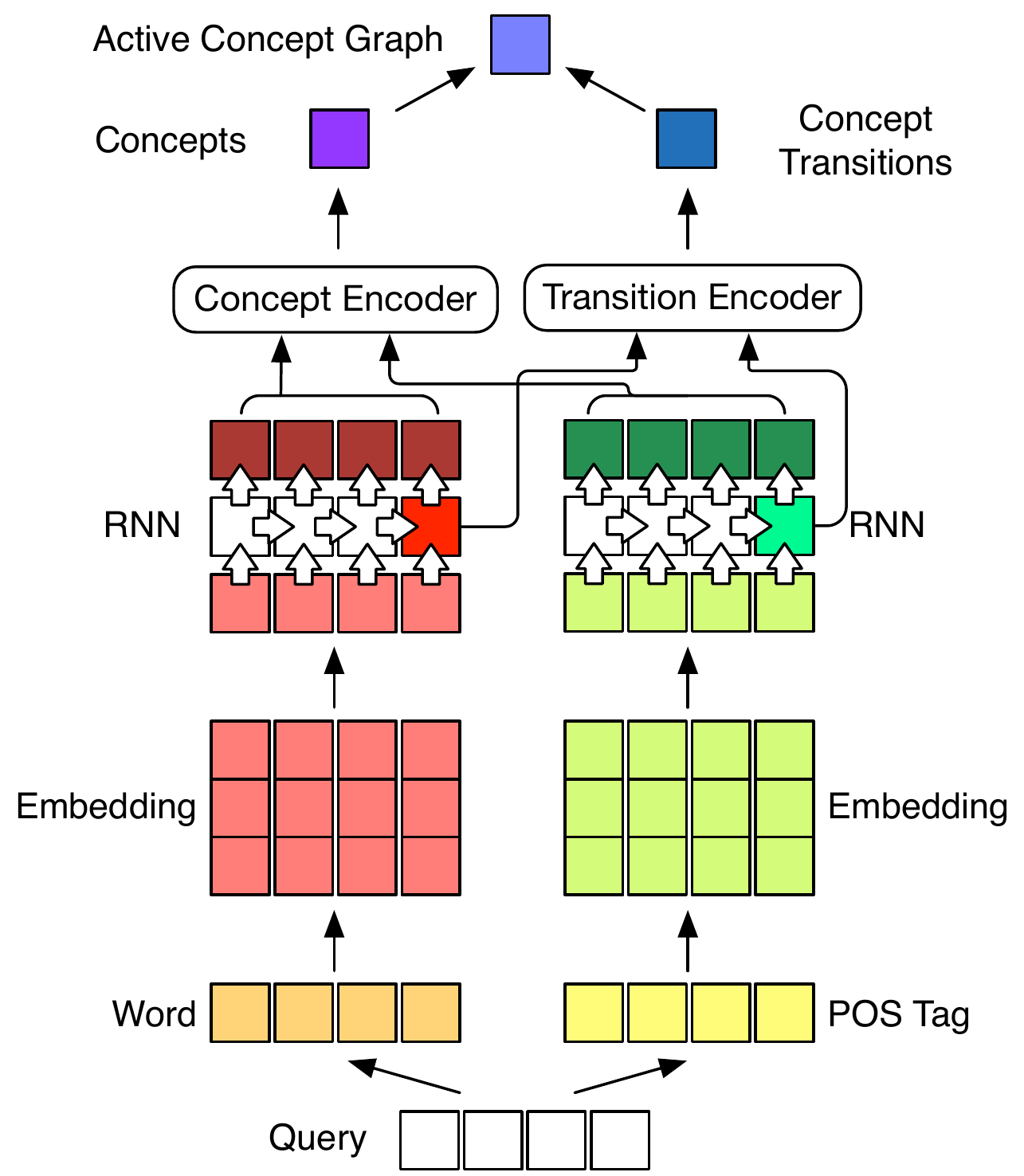, width=3in}
\caption{The proposed neural network architecture.}\label{fig::architecture}
\end{figure}

In this section, a neural network structure is introduced to provide an end-to-end solution to the Structured Intent Detection problem where the input is a text corpus and the output is a Structured Intent inferred from the corpus. The model utilizes word representations to deal with the lexical diversities. Also, part-of-speech embedding of each word is used to further capture the syntax information. Recurrent neural networks are adopted to model the sequential information from distributed representations of word and POS tag sequences in each query simultaneously.
In the graph-based co-inference procedure, concept mentions and semantic transitions are inferred collectively. A Concept Extractor is proposed to utilize the joint outputs of two RNNs to encode each element into a concept vector.
Especially, the Concept Extractor is able to learn an attention weight as a confidence score that indicates the contribution of each element to each concept. While for inferring semantic transitions, a transition encoder learns to summarize the semantics and construct a transition vector, from which we infer a probability distribution over all possible semantic transitions.
The loss of the neural network structure not only incorporates prediction errors between the inferred semantic transitions and the true semantic transitions but also exploits a mutual transfer loss indicating the conflicts between the extracted concepts and the semantic transitions. A Structured Intent is presented with the inferred concepts and semantic transitions, by collectively minimizing a graph-based mutual transfer loss based on the intent graph. \ref{fig::architecture} gives an overview of the proposed method.

\subsection{Lexical-Syntax Representations}
Unlike traditional methods which ignore the sequential information of the input text corpus and treat it as a bag-of-words (BoW), in this work a text corpus $Q$ is considered as a sequence of elements $\{q_1, q_2, ..., q_K\}$, where each element $q_k$ can be a word or a phrase. $K$ is the length of the text corpus, which varies in different corpora. For each element $q_k$ in a text corpus $Q$, we utilize both word representations indicating the lexical information, as well as its corresponding Part-of-Speech (POS) tag as the syntax information.

Part-of-speech (POS) tags bring useful syntax information about general word categories (such as noun, verb, adjective, etc.), which is helpful in dealing with ambiguous words and diversified expressions. For example, \texttt{fever} can be either a noun or a verb. The word \texttt{fever} with a POS tag ``noun'' is defined as a disease that causes an increase in body temperature and the fever with a POS tag ``verb'' can be considered as someone in a fever, as a symptom. In this work, an existing POS tagger\footnote{https://github.com/fxsjy/jieba} is utilized to give general POS tags to each element in the text corpus. The lexical-syntax joint representation consists of words along with POS tags are shown to be effective in modeling both lexical (words) and syntax (POS tags) from the natural language text corpus in various tasks \cite{legrand2015joint,zhang2016mining}. In this work, each element $q_k$ of a text corpus $Q$ is represented by words and POS tags as a tuple:
\begin{equation}\label{eq::joint_feature}
{q_k} = \left( {{w_k},{p_k}}\right)~s.t.~w_k \in \mathbb{R}^{V_{word}}, p_k \in \mathbb{R}^{V_{pos}},
\end{equation}
where $w_k$ is the one-hot representation of the $k$-th word in the corpus $Q$ and $V_{word}$ is the number of unique words, namely the vocabulary size. Similarly, $p_k$ is the one-hot representation of the $k$-th word's POS tag in the corpus. $V_{POS}$ is the POS vocabulary size.

\subsection{Word Embedding}
The one-hot representation suffers from the curse of dimensionality since the representation becomes extremely sparse as the vocabulary becomes large. The word embedding is used to transfer one-hot representation of each word $w_k$ and POS tag $p_k$ into a dense representation:
${w\_{embed}}_{k} \in \mathbb{R}^{D_{word}}, {p\_{embed}}_{k} \in \mathbb{R}^{D_{pos}}$, where $V_{word}$ usually can be large up to millions while $D_{word}$ is reduced to several hundreds. Note that ${D_{word}}$ and $D_{pos}$ are usually set empirically. In this work, we set ${D_{word}}=100$ and $D_{pos}=20$.
The embedded representation of each $w_k$ and $p_k$ are learned respectively by a linear mapping via a skip-gram model \cite{mikolov2013distributed}:
\begin{equation}
{{embed}\_w}_{k} = \mathbf{W}_{word}~w_{k}, \quad {{embed}\_p}_{k} = \mathbf{W}_{pos}~p_{k},
\end{equation}
where ${\mathbf{W}_{word}} \in \mathbb{R}^{D_{word}\times V_{word}}$ and ${\mathbf{W}_{pos}} \in \mathbb{R}^{D_{pos}\times V_{pos}}$ are weights.

In this work, the embedding is initialized with word vectors pre-trained from 64 million text corpus and updated with the model during training.
After the word embedding, the $k$-th element in the text corpus $q_k$ has a lexical-syntax representation, represented by a tuple: ${e}_k = ({embed\_w}_k, {embed\_p}_k)$.
\subsection{Recurrent Neural Network}
Once we obtained a representation $e_k$ for each element $q_k$ in a text corpus $Q$, the ${embed\_w}_k$ sequence and the ${embed\_p}_k$ sequences are fed into two recurrent neural networks, namely $\operatorname{RNN_W}$ and $\operatorname{RNN_P}$, to capture the sequential semantics respectively. 

In general, a recurrent neural network keeps hidden states over a sequence of elements and updates the hidden state $h_k$ by the current input $x_k$ as well as the previous hidden state $h_{k-1}$ where $k>1$ by a recurrent function:
$h_k = \operatorname{RNN}({x}_k, h_{k-1})$.
The Gated Recurrent Unit (GRU) \cite{cho2014learning} is proposed to address the gradients decay or exploding problem \cite{bengio1994learning,hochreiter1998vanishing} over long sequences in the vanilla RNN. The GRU has been attracting great attention since it overcomes the vanishing gradient in traditional RNNs and is more efficient than LSTM \cite{hochreiter1997long} on certain tasks \cite{chung2014empirical}. The GRU is designed to learn from previous time stamps with long time lags of unknown size between important time stamps. 

In this work, two separate RNN with GRU cells, namely $\operatorname{RNN}_W$ and $\operatorname{RNN}_P$, are adopted to model the sequential information in the sequence of embedded words ${embed\_w}_k$ and the sequence of embedded POS tags ${embed\_p}_k$:
\begin{equation}
{h\_w}_k, {o\_w}_k = \operatorname{RNN_W}({embed\_w}_k, {h\_w}_{k-1}),\quad 
{h\_p}_k, {o\_p}_k = \operatorname{RNN_P}({embed\_p}_k, {h\_p}_{k-1}),    
\end{equation}

\subsection{Graph-based Co-inference}
In order to fully exploit the correlations of concept mentions and semantic transitions, instead of inferring concepts and semantic transitions separately, a collective inference schema is adopted. The Concept Extractor aims to select a subset of concepts $\hat{C}_Q \subseteq C$ that are mentioned in a the corpus $Q$. A transition encoder is introduced to infer semantic transitions $\hat{T}_Q \subseteq T$ over the Intent Graph $G$. The concepts $\hat{C}_Q$ and transitions $\hat{T}_Q$ are inferred collectively, by minimizing a mutual transfer loss which indicates the conflicts within the inferred Structured Intent $\hat{G}_Q=<\hat{C}_Q,\hat{T}_Q>$.

\begin{figure}[btp]
\centering
\epsfig{file=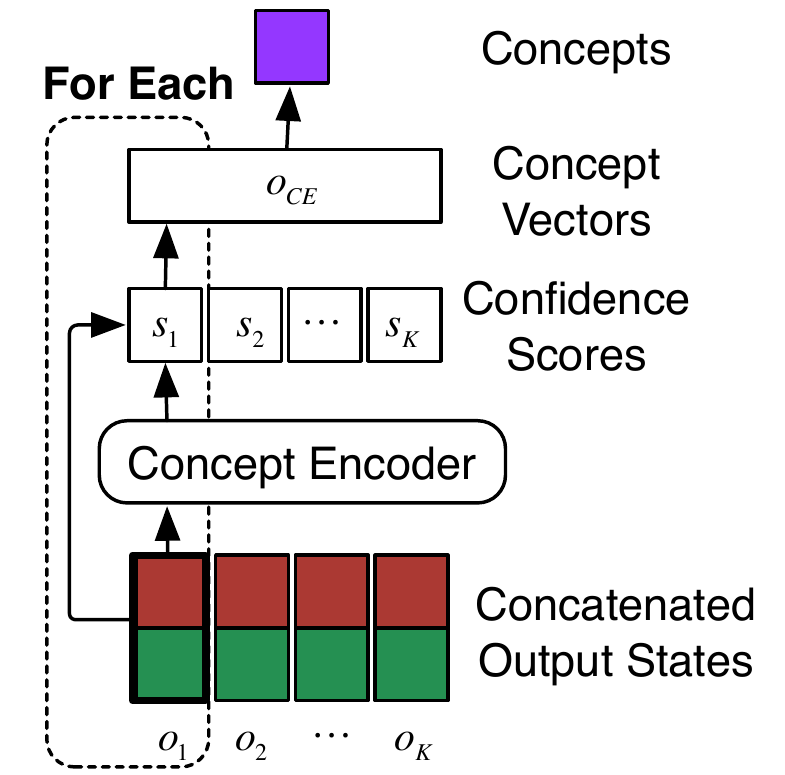, height=2in}
\caption{The Concept Extractor.}\label{fig::concept_encoder}
\end{figure}

\noindent\textbf{Concept Extractor}
The Concept Extractor encodes all concept mentions from a sequence of output states of an $\operatorname{RNN}$ to a single concept vector. Since some words in the text corpus may contribute more to a concept, the Concept Extractor itself learns to assign a confidence score to each output state.
Let ${o}_k$ be the $k$-th output vector of an $\operatorname{RNN}$, while in this work we concatenate the output vectors of $\operatorname{RNN}_W$ and $\operatorname{RNN}_P$:
\begin{equation}
{o}_k = [{o\_w}_k, {o\_p}_k], {o\_w}_k\in\mathbb{R}^{1\times{D_{o_w}}},{o\_p}_k\in\mathbb{R}^{1\times{D_{o_p}}},
\end{equation} where $D_{o_w}$ and $D_{o_p}$ are the output dimensions of output vectors in $\operatorname{RNN}_W$ and $\operatorname{RNN}_P$.
The Concept Extractor assigns a score $s_k$ for each $o_k$ indicating the degree of confidence, parameterized by $\theta$:
\begin{equation}
{s_k} = \frac{{CE\left( {{o_k};\theta } \right)}}{{\sum\limits_{k' \in K} {CE\left( {{o_{k'}};\theta } \right)} }}
\quad {\text{s.t. }}\sum\nolimits_k {{s_k} = 1} ,\forall {s_k} \in [0,1].
\end{equation}
The $s_k$ scores for all elements in a text corpus are normalized to sum up to one. 
We implement the $CE(\cdot)$ function as a single layer neural network with a non-linear activation function ReLU. Thus $\theta$ consists of $\{\mathbf{W}_{\theta}\in \mathbb{R}^{(D_{o_w}+D_{o_p})\times1}, b_{\theta}\in{\mathbb{R}}\}$. Note that although weights and biases are applied on each of the $o_k$, they are shared among all ${o_1,o_2,...,o_K}$.
\ref{fig::concept_encoder} shows the architecture of the Concept Extractor, which is used to determine confidence scores for each joint output state. This figure shows an example of a score $s_1$ learned from the Concept Extractor for $o_1$.
The $o_{CE}\in\mathbb{R}^{(D_{o_w}+D_{o_p})\times{1}}$ is a representation of encoded concepts from the text corpus, which is calculated as the weighted sum on the output vectors as ${O_{CE}} = \sum\limits_k {{s_k}{o_k}}$.

The probability of a concept $c_i \in C$ being mentioned in any part of the text corpus $Q$ is defined by a softmax function over logits on all concepts, where the logit for each concept is learned by a logistic function:
\begin{equation}
({\hat{C}_{Q}})_m = P(c_m|c_m\in{C};\theta)=\frac{1}{{1 + {e^{ - {W_{CE}}{O_{CE}} + {b_{CE}}}}}},
\end{equation}
where $\mathbf{W}_{CE} \in \mathbb{R}^{1 \times (D_{o_w}+D_{o_p})}$, $b_{CE} \in \mathbb{R}$ are weights and biases for each type of concept $m \in M$. 
We feed logits $\hat{C}_Q\in\mathbb{R}^{1\times{M}}$ to the softmax layer and get the probability distribution over all $M$ types of concepts being mentioned in the given text corpus $Q$. 

\noindent\textbf{Transition Encoder}
In the field of machine translation, a novel recurrent neural network encoder-decoder has gained attention \cite{sutskever2014sequence}, where the encoder recurrent neural network encodes the global information spanning over the whole input sentence in its last hidden state. Inspired by the effectiveness of the last hidden states in modeling natural language sequences in applications like dialog systems \cite{serban2016building}, we propose a transition encoder which leverages the last hidden state of the neural network for both $\operatorname{RNN}_W$, $\operatorname{RNN}_P$ to make inferences on semantic transitions, where the transition vector $o_{TE}$ is constructed by
$O_{TE} = [{h\_w}_K,{h\_p}_K]$, where $K$ is the length of the query.
The logit of a transition $t_{n}\in{T}$ is quantified as:
\begin{equation}
(\hat{T}_{Q})_{n}=P(t_{n}|t_{n}\in{T};\phi)=\frac{1}{{1 + {e^{ - {W_{TE}}{O_{TE}} + {b_{TE}}}}}},
\end{equation}
where $\phi=\{\mathbf{W}_{TE} \in \mathbb{R}^{1 \times (D_{ow}+D_{op})}, b_{TE} \in \mathbb{R}\}$ parameterizes weights and biases for each type of transition. Similarly, $\hat{T}_Q\in\mathbb{R}^{1\times{N}}$ is fed to the softmax layer and we get the inferred probability distribution over all $N$ semantic transitions.

\subsection{Mutual Transfer Loss}
The idea of mutual transfer loss is to characterize the loss caused by transferring the inferred semantic transitions to their corresponding concept mentions, and the other way around. Since for each semantic transition $t_{{i}\to{j}}\in{T}$, two concepts $c_i$ and $c_j$ are involved. If a semantic transition $t_{{i}\to{j}}$ is inferred with a high probability while its corresponding concepts $c_i$, $c_j$ have low probabilities, then that indicates conflicts in the final Structured Intent. The mutual transfer loss is proposed in the co-inference procedure to minimize the conflicts between the inferred concepts and semantic transitions so that the resulting Structured Intent can be more reasonable.

The graph-based formulation for the Structured Intent gives an appealing property that transitions and their proximate concepts can be clearly characterized by a transfer matrix $A\in\mathbb{R}^{M\times{N}}$ over the Intent Graph $G=\langle{C},{T}\rangle$. Each entry $a_{mn}=1$ if and only if the concept $c_m$ involves in at least one end of a semantic transition $t_{m\to{\cdot}}$ or $t_{{\cdot}\to{m}}$.
The mutual transfer loss is defined on $\hat{C}_Q, \hat{T}_Q, \mathcal{T}_Q$ as:
\begin{equation}
\mathcal{L}_{MTL}(\hat{C}_Q, \hat{T}_Q, \mathcal{T}_Q) = H(\mathcal{T}_Q, \hat{T}_Q) + E(\hat{C}_Q, \hat{T}_Q),
\end{equation}
where $\mathcal{T}_Q$ is a ground truth one-hot indicator for semantic transitions given the corpus $Q$. $\hat{C}_Q$ and $\hat{T}_Q$ are extracted concepts and inferred semantic transitions with the proposed method. $H(\cdot, \cdot)$ calculates the cross entropy \cite{tsoumakas2009mining}. $E(\hat{C}_Q, \hat{T}_Q)$ is an energy-based function on inferred transitions $\hat{T}_Q$ and extracted concepts $\hat{C}_Q$. Each combination of $\hat{C}_Q$ and $\hat{T}_Q$ corresponds with an energy value, the lower energy level a combination of $\hat{C}_Q$ and $\hat{T}_Q$ has indicates less conflicts among the inferred concepts and transitions. In this work, an energy-based function for $E(\hat{C}_Q, \hat{T}_Q)$ is proposed as:
\begin{equation}
 E(\hat{C}_Q, \hat{T}_Q) = \mathcal{L}_R(\hat{C}_Q, \hat{T}_Q{{A}^T}) + \mathcal{L}_R(\hat{T}_Q,\hat{C}_Q{A}),
\end{equation}
where $\mathcal{L}_R$ is implemented by a ranking loss function \cite{murphy2012machine} that penalizes cases where the inferred concepts/transitions after transformation by matrix $A$ have high probabilities but order below the ranking of the originally inferred concepts/transitions in the same corpus. $\mathcal{L}_R$ has a general form:
\begin{equation}
\begin{gathered}
  \mathcal{L}_{R}(\hat X,\hat Y) = {\frac{1}{{\left| {{{\hat X}}} \right|(L - \left| {{{\hat X}}} \right|)}}\left| {} \right.} 
\{ \left( {p,q} \right):{{\hat Y}_{p}} < {{\hat Y}_{q}},{\hat X_{p}} \geq{\hat X_{q}} \},  \hfill \\
\end{gathered}
\end{equation}
where $\hat X\in{\mathbb{R}^{1\times{L}}}$ is the originally inferred labels and $\hat Y\in{\mathbb{R}^{1\times{L}}}$ is the inferred labels from the transformation with $A$. $\left|\cdot\right|$ denotes the number of ground truth labels being assigned. $L$ is the label size, where we have $M$ for concepts and $N$ for semantic transitions.
\section{Evaluation}
\subsection{Dataset}\label{sec::data}
We collect text corpora from an online medical question answering discussion forum\footnote{http://club.xywy.com}, on which user posted their healthcare related questions and medical professionals give online suggestions or advice. 
The obtained corpora are in Chinese. Due to the fact that sentences in Chinese are not naturally split by spaces, word segmentation is performed using a Chinese word segmentation package\footnote{https://github.com/fxsjy/jieba}.

After preprocessing and annotation, we obtain 10,000 pieces of text corpora. We end up having 17 unique types of concepts and 23 unique types of semantic transitions,
among which 11,531 unique words and 60 unique POS tags are observed. 
The POS tagging uses ICTCLAS annotation \cite{zhang2003hhmm}. 
The average length of the text corpus is 13.8, with a standard variation of $\pm$6.1. The average number of concepts in the labeled corpus is 3.6020$\pm$0.8. The average number of semantic transitions is 2.4723$\pm$0.7.

Word embeddings are pre-trained using a skip-gram model \cite{mikolov2013distributed} on 64 million unlabeled text corpus separately.
Context window size is set to 8 and we specify a minimum occurrence count of 5. The vocabulary contains 100-dimension vectors on 382,216 words. Words not presented in the set of pre-trained words are initialized as random vectors. All word vectors will be updated during training.

\subsection{Experiment Settings}
To show the advantages of the proposed method in addressing the concept transition inference problem, we compare it with the following baseline models. 
\begin{itemize}
\item \textbf{LR}: a logistic regression model applied with POS tagging features and word representations.
\item \textbf{NNID-JM} \cite{zhang2016mining}: the neural network intent detection model with joint modeling. Both words and POS tags are used to characterize the words in the corpus. Domain-specific POS tags, such as ``noun\_medicine'', are used in NNID-JM instead of ``noun'' for word ``Tylenol''. The NNID-JM doesn't explicitly exploit label correlations on the output level.
\item \textbf{CI}: the Concept Inference model which only infers mention of concepts from the corpus with the Concept Extractor. $H(\mathcal{C}_Q, \hat{C}_Q)$ is used as the loss function for the CI task.
\item \textbf{CTI}: the Concept Transition Inference model without co-inference. Only semantic transitions are inferred from the corpus. The last output states of two RNNs are concatenated to predict the semantic transitions. $H(\mathcal{T}_Q, \hat{T}_Q)$ is used as the loss function.
\item \textbf{coCTI}: the Concept Transition Inference model with co-inference. $H(\mathcal{T}_Q, \hat{T}_Q) + H(\mathcal{C}_Q, \hat{C}_Q)$ is used as the loss function. This variation can be seen as a multi-task learning model for extracting concepts and inferring semantic transitions, where two tasks share the lower-level neural network structure for word representation.
\item \textbf{coCTI-MTL}: the proposed model with co-inference and a mutual transfer loss $\mathcal{L}_{MTL}$, where the CI task and CTI task not only share the neural network structure, but also adopt the mutual transfer loss.
\end{itemize}

\noindent\textbf{Evaluation Metrics}: 
Each directed edge in the Intent Graph is considered as an individual label and we evaluate inferred Structured Intent as a multi-class, multi-label classification problem.
\textit{Receiver operating characteristic} (ROC), the \textit{micro/macro-average area under the curve} (micro-AUC, macro-AUC), \textit{coverage error} and \textit{label ranking average precision} (LRAP) are used to evaluate the effectiveness of the proposed model in inferring Structured Intents from the text corpus. 
The ROC and AUCs focus on the quality of prediction, while the coverage error and LRAP are introduced to evaluate the completeness/ranking of the prediction. ROC is the curve created by plotting the true positive rate (TPR) against the false positive rate (FPR) at various threshold settings. Micro-AUC computes the averaged area under the ROC curve over all the labels. Coverage error computes the average number of labels that we need to have in the final prediction in order to predict all true labels. LRAP score favors better rank to labels that are associated to each sample and is usually used in multi-label ranking problems.

\noindent\textbf{Experiment Settings}:
The embeddings for word and POS tagging have a dimension of 100 and 20, respectively. The hidden layer and the output layer of the GRU unit have a dimension of 100. For training the proposed neural network structure, 70\% of the labeled data are used for training and 10\% samples are served as the validation set to tune for the best parameter set. The remaining data are used for testing. Cross-validation is used and we combine test data in each fold to report the test performance.
The optimization is performed in a mini-batch fashion with a batch size of 32. The Adam Optimizer \cite{kingma2014adam} is applied to train the neural network and the initial learning rate is set to $10^{-4}$. Weight variables are initialized with the Xavier initializer \cite{glorot2010understanding} and bias variables are initialized as zeros.

\subsection{Experiment Results}
\begin{figure}[h]
\centering
\epsfig{file=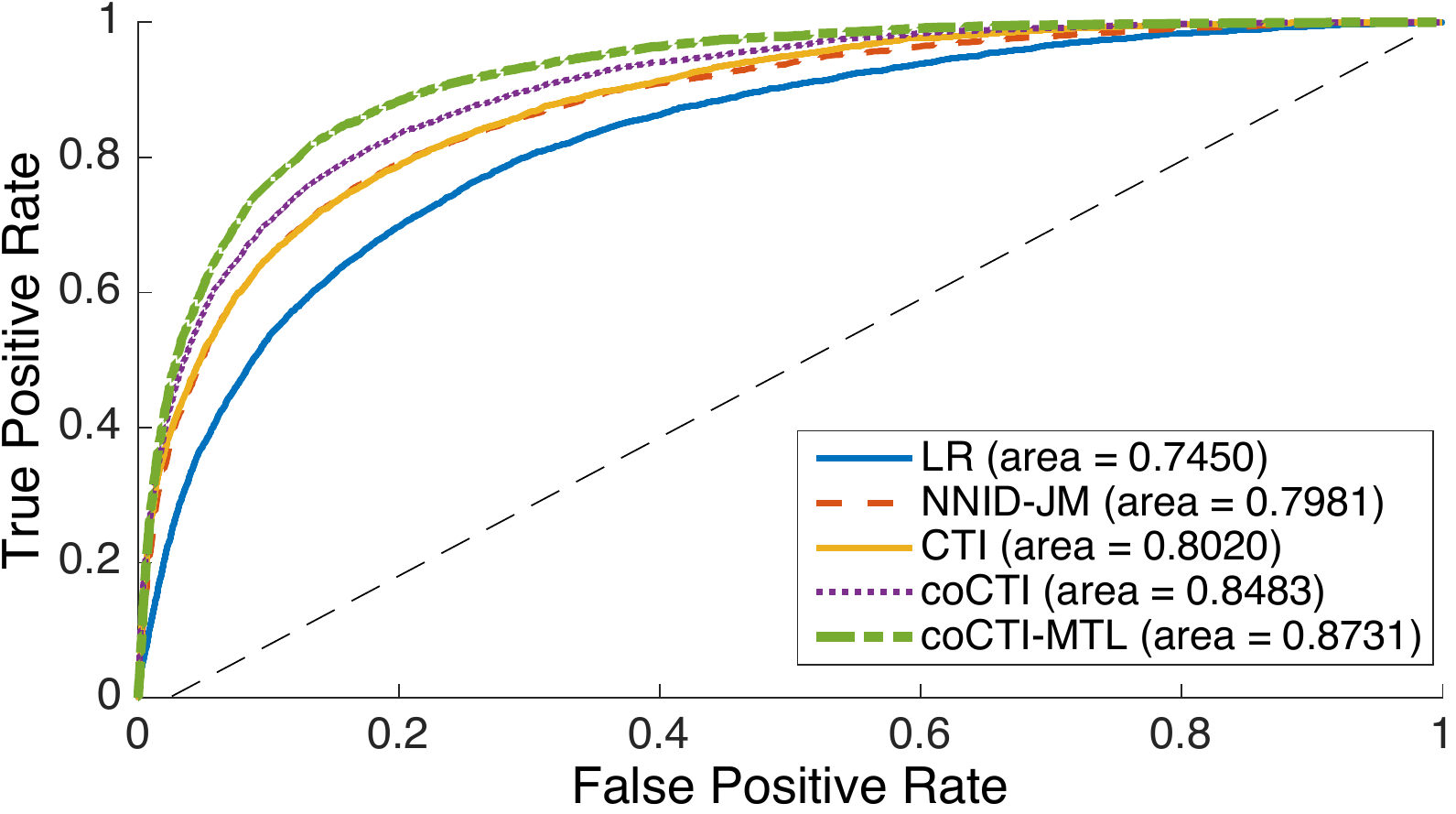, width=3.3in}
\caption{Micro-AUC scores and ROC curves.}\label{fig::mroc}
\end{figure}

\ref{fig::mroc} shows the effectiveness of the proposed model by micro-AUC and ROC curves. Generally, neural network based models (NNID-JM, CTI, coCTI, coCTI-MTL) outperform traditional logistic regression model (LR) consistently. 
For NNID-JM, in order to make a fair comparison, domain specific POS tags (such as noun\_disease, noun\_medicine, noun\_symptom) are maintained as an external knowledge base. Those POS tags are used by the POS tagger in NNID-JM as its default setting. When compared with NNID-JM, the proposed CTI model achieves similar performance on micro-AUC, while it doesn't rely on any other external knowledge like domain-specific POS tags in NNID-JM. 

From \ref{fig::mroc} we can further observe that CTI-MTL achieves the best performance (0.8731 in micro-AUC) among all the comparison methods in correctly inferring semantic transitions from the text corpus. The CTI-MTL model has a nearly 2.5\% improvement on micro-AUC when compared with coCTI and a nearly 7.5\% improvement with CTI. This demonstrates that the mutual transfer loss which penalizes conflicts between the extracted concept mentions and inferred semantic transitions can indeed improve the structured intent detection performance.

\begin{figure}[h]
\centering
\epsfig{file=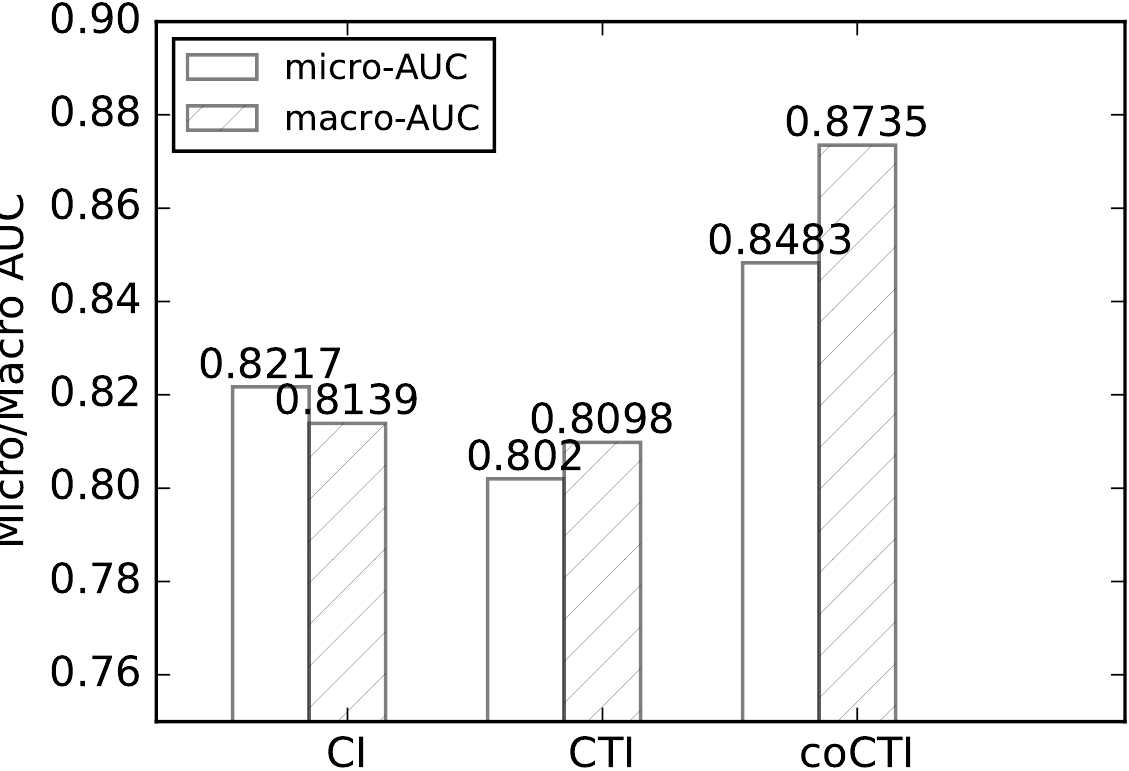, width=4in}
\caption{Micro/Macro-AUC scores on collective inference and separate inference.}\label{fig::coinferornot}
\end{figure}

\begin{table}[h]
\centering
\resizebox{4.5in}{!}{%
\begin{tabular}{lccccc}
\toprule
Concept Transition & LR& NNID-JM& CTI& coCTI & coCTI-MTL\\ \hline
\textit{Symptom}$\to$\textit{Diet} & 0.6544 {\color{blue}(5)}& 0.7755 {\color{blue}(4)}& 0.7669 {\color{blue}(3)}& 0.7959 {\color{blue}(2)}& 0.8495 {\color{blue}(1)}\\
\textit{Symptom}$\to$\textit{Medicine} & 0.7022 {\color{blue}(5)} & 0.7893 {\color{blue}(4)} & 0.8242 {\color{blue}(3)} & 0.8571 {\color{blue}(2)} & 0.8624 {\color{blue}(1)}\\
\textit{Symptom}$\to$\textit{Cause}  & 0.7600 {\color{blue}(5)} & 0.8549 {\color{blue}(4)} & 0.8786 {\color{blue}(3)} & 0.8911 {\color{blue}(1)} & 0.8880 {\color{blue}(2)}\\
\textit{Disease}$\to$\textit{Diet}  & 0.7818 {\color{blue}(5)} & 0.8670 {\color{blue}(4)} & 0.8681 {\color{blue}(3)} & 0.9059 {\color{blue}(2)}& 0.9458 {\color{blue}(1)}\\
\textit{Disease}$\to$\textit{Treatment}  & 0.7181 {\color{blue}(5)} & 0.7787 {\color{blue}(3)} & 0.7482 {\color{blue}(4)} & 0.8456 {\color{blue}(2)}& 0.8836 {\color{blue}(1)}\\
\textit{Disease}$\to$\textit{Examine}  & 0.6397 {\color{blue}(5)} & 0.6707 {\color{blue}(4)} & 0.7838 {\color{blue}(3)} & 0.8221 {\color{blue}(2)} & 0.8480 {\color{blue}(1)}\\
\textit{Disease}$\to$\textit{Medicine} & 0.7623 {\color{blue}(5)} & 0.8726 {\color{blue}(4)} & 0.8749 {\color{blue}(3)} & 0.8873 {\color{blue}(2)} & 0.9015 {\color{blue}(1)}\\
\textit{Surgery}$\to$\textit{Recover}  & 0.8117 {\color{blue}(5)} & 0.9126 {\color{blue}(3)} & 0.9012 {\color{blue}(4)} & 0.9239 {\color{blue}(2)} & 0.9396 {\color{blue}(1)}\\
\textit{Surgery}$\to$\textit{Sequela}  & 0.7385 {\color{blue}(5)} & 0.8031 {\color{blue}(4)} & 0.8214 {\color{blue}(3)} & 0.8417 {\color{blue}(2)} & 0.8972 {\color{blue}(1)}\\
\textit{Surgery}$\to$\textit{Syndrome}  & 0.7896 {\color{blue}(5)} & 0.7994 {\color{blue}(4)} & 0.8634 {\color{blue}(2)} & 0.8619 {\color{blue}(3)} & 0.9172 {\color{blue}(1)}\\
\textit{Surgery}$\to$\textit{Risk}  & 0.6613 {\color{blue}(5)} & 0.8063 {\color{blue}(4)} & 0.8688 {\color{blue}(3)} & 0.8715 {\color{blue}(2)} & 0.9099 {\color{blue}(1)}\\
\textit{Medicine}$\to$\textit{Symptom}  & 0.6861 {\color{blue}(5)} & 0.8275 {\color{blue}(3)} & 0.7553 {\color{blue}(4)} & 0.8294 {\color{blue}(2)} & 0.8598 {\color{blue}(1)}\\
\textit{Medicine}$\to$\textit{Side Effect}  & 0.6652 {\color{blue}(5)} & 0.8162 {\color{blue}(3)} & 0.7771 {\color{blue}(4)} & 0.8135 {\color{blue}(2)} & 0.8814 {\color{blue}(1)}\\
\textit{Medicine}$\to$\textit{Disease}  & 0.6806 {\color{blue}(4)} & 0.6514 {\color{blue}(5)} & 0.8081 {\color{blue}(3)} & 0.8126 {\color{blue}(2)} & 0.8678 {\color{blue}(1)}\\
\textit{Medicine}$\to$\textit{Instruction}  & 0.7090 {\color{blue}(5)} & 0.7761 {\color{blue}(3)} & 0.7603 {\color{blue}(4)} & 0.8170 {\color{blue}(2)} & 0.8820 {\color{blue}(1)}\\
\textit{Examine}$\to$\textit{Fee}  & 0.7576 {\color{blue}(5)} & 0.9049 {\color{blue}(3)} & 0.8981 {\color{blue}(4)} & 0.9425 {\color{blue}(2)} & 0.9482 {\color{blue}(1)}\\
\textit{Examine}$\to$\textit{Diagnosis}  & 0.6832 {\color{blue}(5)} & 0.7956 {\color{blue}(3)} & 0.7445 {\color{blue}(4)} & 0.8383 {\color{blue}(2)} & 0.8822 {\color{blue}(1)}\\
\textit{Symptom}$\to$\textit{Treatment}  & 0.6817 {\color{blue}(5)} & 0.7640 {\color{blue}(3)} & 0.7313 {\color{blue}(4)} & 0.8130 {\color{blue}(2)} & 0.8531 {\color{blue}(1)}\\
\textit{Symptom}$\to$\textit{Department}  & 0.5978 {\color{blue}(5)} & 0.6460 {\color{blue}(3)} & 0.6013 {\color{blue}(4)} & 0.6738 {\color{blue}(2)} & 0.8080 {\color{blue}(1)}\\
\textit{Disease}$\to$\textit{Cause}  & 0.7306 {\color{blue}(5)} & 0.8206 {\color{blue}(4)} & 0.8515 {\color{blue}(3)} & 0.8608 {\color{blue}(2)} & 0.8634 {\color{blue}(1)}\\
\textit{Disease}$\to$\textit{Symptom}  & 0.6936 {\color{blue}(4)} & 0.7552 {\color{blue}(3)} & 0.6845 {\color{blue}(5)} & 0.7554 {\color{blue}(2)} & 0.8372 {\color{blue}(1)}\\
\textit{Disease}$\to$\textit{Department}  & 0.6931 {\color{blue}(5)} & 0.7387 {\color{blue}(4)} & 0.7431 {\color{blue}(3)} & 0.7652 {\color{blue}(2)} & 0.8290 {\color{blue}(1)}\\
\textit{Disease}$\to$\textit{Surgery}  & 0.7801 {\color{blue}(5)} & 0.8795 {\color{blue}(4)} & 0.9029 {\color{blue}(3)} & 0.9236 {\color{blue}(2)} & 0.9380 {\color{blue}(1)}\\\bottomrule
\end{tabular}
}
\caption{Fine-grained AUC scores for all semantic transitions.}\label{tab::fineauc}
\end{table}

\ref{fig::coinferornot} shows the effectiveness of the co-inference procedure by comparing the performance of CTI with coCTI. The CI infers concept mentions so we can't simply compare its performance with CTI/coCTI where semantic transitions are inferred. However, for CTI and coCTI, the improved performance on both micro-AUC and macro-AUC validates the effectiveness of inferring concepts and semantic transitions collectively than inferred separately. The coCTI model can be considered as a multi-task learning model where the lower-level text representations are learned jointly and shared between two sub-tasks.

Furthermore, the fine-grained AUC scores on all semantic transitions without micro/macro-averaging are shown in \ref{tab::fineauc}. A general observation we can draw from the results is that the coCTI-MTL model is able to outperform other baselines in almost all types of semantic transitions. 
\begin{figure}[tbh!]
\centering
\epsfig{file=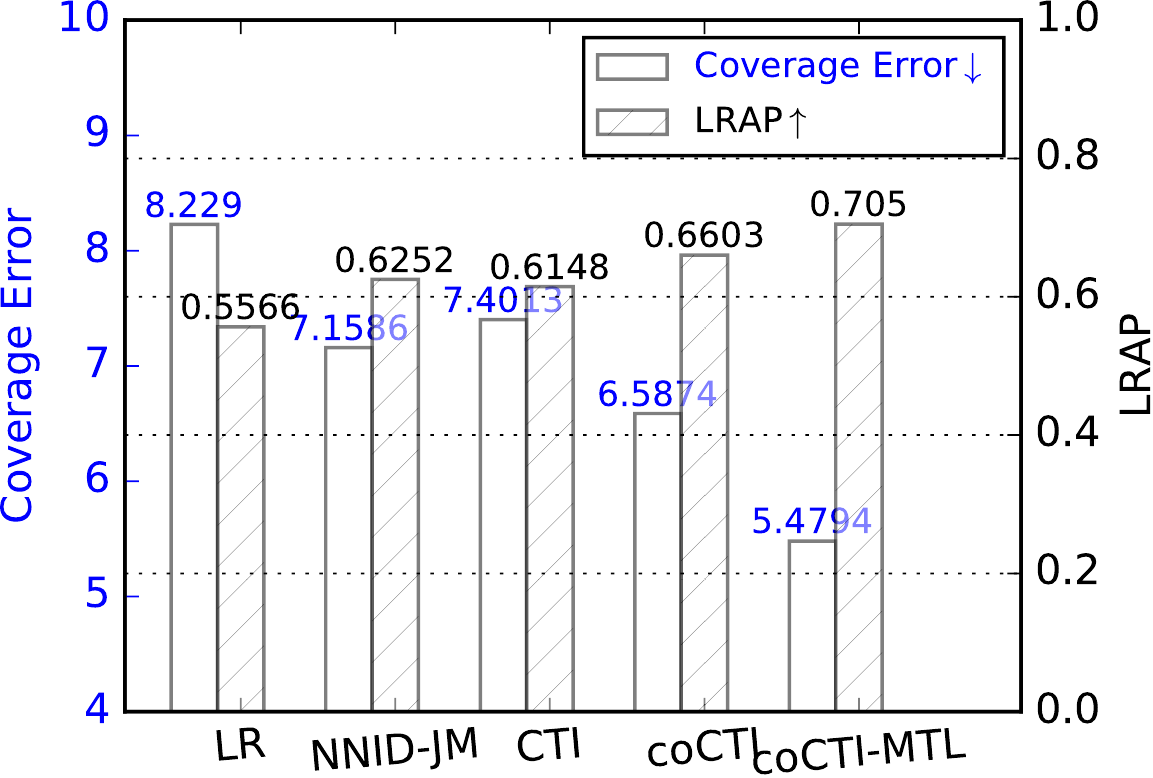, width=4.5in}
\caption{Coverage Loss and Label Ranking Average Precision (LRAP).}\label{fig::cl_lrap}
\end{figure}

\ref{fig::cl_lrap} shows the coverage loss and LRAP over proposed methods and other baselines, where the coCTO-MTL model is able to achieve the lowest coverage error and the highest label ranking average precision score.
\section{Related Works}
\noindent\textbf{Query Analysis}
As the number of people posting questions or searching for information online is growing rapidly, researchers have been focusing on new problems and applications based on the user-generated text corpus, such as queries or search queries. \cite{limsopatham2013inferring} analyzes the conceptual relationship in online web documents records for a better web search. \cite{stanton2014circumlocution} focuses on the circumlocution problem in diagnostic questions in the healthcare domain, where users are not able to express their ideas effectively. \cite{zhang2016mining} tries to model user intentions as a classification task for text queries. 
\cite{liu2015context} proposes a technique to detect whether users express their own experiences in the generated text corpus. 
\cite{li2016extracting} introduces a knowledge discovery model for the online question-answering corpus.
In \cite{liu2016augmented}, authors introduce a neural network model to understand users questions and try to generate answers appropriately. Being able to infer concept transitions from noisy, user-generated questions may further facilitate various applications in domains like healthcare, such as healthcare question-answering, medical dialog systems or recommendation. For example, once we extracted the concept transition $Symptom \to Medicine$ from a question \texttt{Any medication is recommended to help me fall asleep easier?}, we may follow up by recommending the user to the nearest pharmacy for further medical consultations on corresponding OTC medicines on Insomnia.

\noindent\textbf{Text Classification}
Recently, lots of neural network models are developed for classifying natural language text corpus into different categories \cite{Grefenstette2014,Lai2015}. Those methods achieve decent performance on general text classification tasks.  
The proposed Structured Intent Detection task can be seen as a multi-class multi-label classification problem. Unlike traditional text classification tasks like news classification where the existence of some topic words may easily dominate the label for a news title, users tend to mention multiple concepts in a single piece of text corpus. It is crucial to accurately infer semantic transitions among those concepts, besides extracting concept mentions only.

Also, the aforementioned methods consider the textual information only. With a graph-based formation in this work, our model seamlessly incorporates an existing Intent Graph for effective intent detection on complicated information-seeking text corpora. More specifically, we propose to predict concept mentions as nodes and semantic transitions as links collectively, while most existing works have been focusing on predicting links among concrete entities, e.g. among users in social networks \cite{liben2007link}, or predicting links among entities on a knowledge graph \cite{nickel2016review,bordes2013translating}.

\chapter{\TaskNameTwo}\label{chapter:c3}
Part of this chapter was published as ``Joint Slot Filling and Intent Detection via Capsule Neural Networks'', in ACL'19~\cite{zhang2018joint}: \url{https://arxiv.org/abs/1812.09471}.
\newcommand{\ModelName}{{\textsc{Capsule-NLM}}}
\newcommand{\IntentModelName}{{\textsc{IntentCapsNet}}}
\newcommand{\UnseenCapsule}{Zero-shot DetectionCaps}
\newcommand{\FirstCapsule}{WordCaps}
\newcommand{\SecondCapsule}{SlotCaps}
\newcommand{\ThirdCapsule}{IntentCaps}
\graphicspath{{arxiv18/figures}}
\section{Introduction}
\begin{figure}[htbp]
    \centering
    \includegraphics[width=0.7\linewidth]{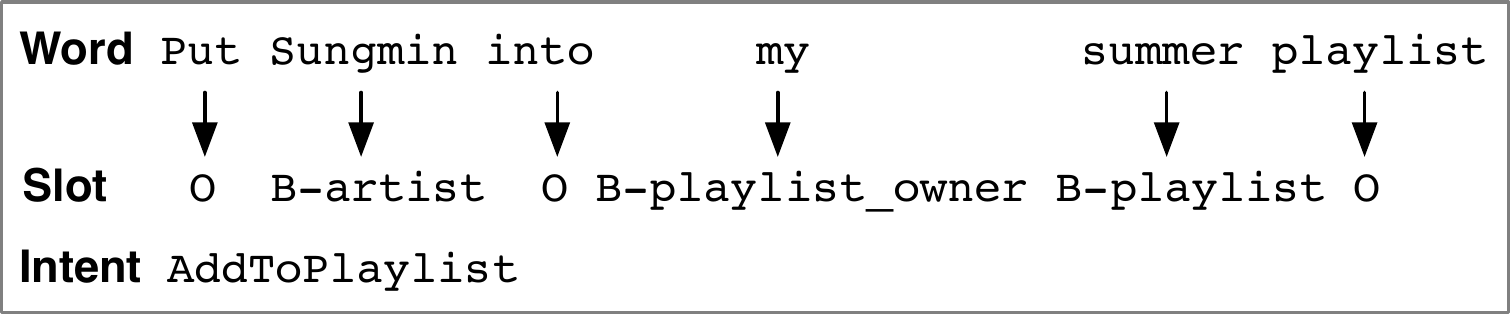}
    \caption{An example of an utterance with BIO format annotation.}\label{fig::illustration}
\end{figure}

With the ever-increasing accuracy in speech recognition and complexity in user-generated utterances, it becomes a critical issue for mobile phones or smart speaker devices to understand the natural language in order to give informative responses. Slot filling and intent detection play important roles in Natural Language Understanding systems.
For example, given an utterance from the user, the slot filling annotates the utterance on a word-level, indicating the slot type mentioned by a certain word such as the slot \texttt{artist} mentioned by the word \texttt{Sungmin}, while the intent detection works on the utterance-level to give categorical intent label(s) to the whole utterance. \ref{fig::illustration} illustrates this idea.

To deal with diversely expressed utterances without additional feature engineering, deep neural network based user intent detection models \cite{hu2009understanding,xu2013convolutional,zhang2016mining,liu2016attention,zhang2017bringing,chen2016end,xia2018zero} are proposed to classify user intents given their utterances in the natural language. 

Currently, the slot filling is usually treated as a sequential labeling task. A neural network such as a recurrent neural network (RNN) or a convolution neural network (CNN) is used to learn context-aware word representations, along with sequence tagging methods such as conditional random field (CRF) \cite{lafferty2001conditional} that infer the slot type for each word in the utterance.

Word-level slot filling and utterance-level intent detection can be conducted simultaneously to achieve a synergistic effect.
The recognized slots, which possess word-level signals, may give clues to the utterance-level intent of an utterance. For example, with a word \texttt{Sungmin} being recognized as a slot \texttt{artist}, the utterance is more likely to have an intent of \texttt{AddToPlayList} than other intents such as \texttt{GetWeather} or \texttt{BookRestaurant}. 

Some existing works learn to fill slots while detecting the intent of the utterance \cite{xu2013convolutional,hakkani2016multi,liu2016attention,goo2018slot}: a convolution layer or a recurrent layer is adopted to sequentially label word with their slot types: the last hidden state of the recurrent neural network, or an attention-weighted sum of all convolution outputs are used to train an utterance-level classification module for intent detection. Such approaches achieve decent performances but do not explicitly consider the task taxonomy on two tasks, nor the hierarchical relationship between words, slots, and intents: intents are sequentially summarized from the word sequence.
As the sequence becomes longer, it is risky to simply rely on the gate function of RNN to compress all contexts in a single vector \cite{cheng2016long}.

In this work, we make the very first attempt to bridge the gap between word-level slot modeling and the utterance-level intent modeling via a hierarchical capsule neural network structure \cite{hinton2011transforming,sabour2017dynamic} that is aware of the task taxonomy.
A capsule houses a vector representation of a group of neurons.
The capsule model learns a hierarchy of feature detectors via a routing-by-agreement mechanism: capsules for detecting low-level features send their outputs to high-level capsules only when there is a strong agreement of their predictions to high-level capsules.

\begin{figure*}[htbp]
    \centering
    \includegraphics[width=0.9\linewidth]{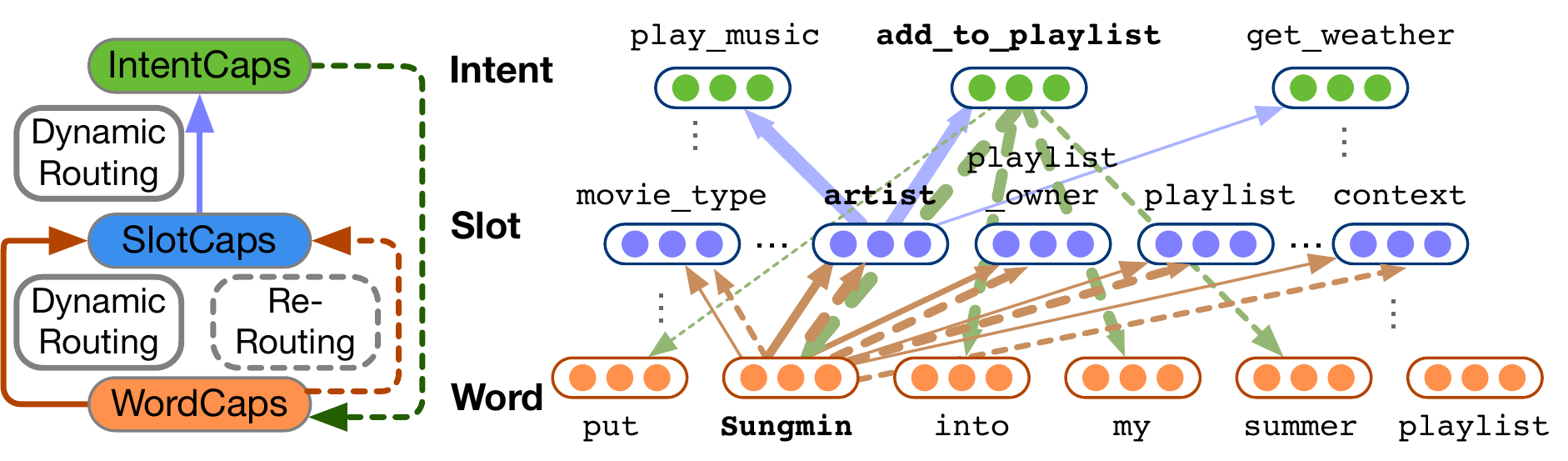}
    \caption{Illustration of the proposed {\ModelName} model.}
    \label{fig::overall}
\end{figure*}

The aforementioned properties of capsule models are appealing for natural language understanding from a hierarchical perspective: words such as \texttt{Sungmin} are routed to concept-level slots such as \texttt{artist}, by learning how each word matches the slot representation. Concept-level slot features such as \texttt{artist}, \texttt{playlist owner}, and \texttt{playlist} collectively contribute to an utterance-level intent \texttt{AddToPlaylist}. The dynamic routing-by-agreement assigns a larger weight from a lower-level capsule to a higher-level when the low-level feature is more predictive to one high-level feature, than other high-level features. \ref{fig::overall} illustrates this idea. The model does slot filling by learning to assign each word in the {\FirstCapsule} to the most appropriate slot in {\SecondCapsule} via dynamic routing. The weights learned via dynamic routing indicate how strong each word in {\FirstCapsule} belongs to a certain slot type in {\SecondCapsule}. The dynamic routing also learns slot representations using {\FirstCapsule} and the learned weight. The learned slot representations in {\SecondCapsule} are further aggregated to predict the utterance-level intent of the utterance. Once the intent label of the utterance is determined, a novel re-routing process is proposed to help improve word-level slot filling by the inferred utterance-level intent label. The solid lines indicate the dynamic-routing process and dash lines indicate the re-routing process. 

The inferred utterance-level intent is also helpful in refining the slot filling result. For example, once an \texttt{AddToPlaylist} intent representation is learned in {\ThirdCapsule}, the slot filling may capitalize on the inferred intent representation and recognize slots that are otherwise neglected previously.
To achieve this, we propose a re-routing schema for capsule neural networks, which allows high-level features to be actively engaged in the dynamic routing between {\FirstCapsule} and {\SecondCapsule}, which improves the slot filling performance.

To summarize, the contributions of this work are as follows:
\begin{itemize}
    \item Encapsulating the hierarchical relationship among word, slot, and intent in an utterance by a hierarchical capsule neural network structure.
    \item Proposing a dynamic routing schema with re-routing that achieves synergistic effects for joint slot filling and intent detection.
    \item Showing the effectiveness of our model on two real-world datasets, and comparing with existing models as well as commercial natural language understanding services.
\end{itemize}
\section{Proposed Approach}
We propose to model the hierarchical relationship among each word, the slot it belongs to, and the intent label of the whole utterance by a hierarchical capsule neural network structure called {\ModelName}. The proposed architecture consists of three types of capsules: 1) {\FirstCapsule} that learn context-aware word representations, 2) {\SecondCapsule} that categorize words by their slot types via dynamic routing, and construct a representation for each type of slot by aggregating words that belong to the slot, 3) {\ThirdCapsule} determine the intent label of the utterance based on the slot representation as well as the utterance contexts. Once the intent label has been determined by {\ThirdCapsule}, the inferred utterance-level intent helps re-recognizing slots from the utterance by a re-routing schema. 

\subsection{WordCaps}
Given an input utterance \(x = \left (  \mathbf{w}_{1}, \mathbf{w}_{2}, ..., \mathbf{w}_{T}\right )\) of $T$ words, where each word is initially represented by a vector of dimension $D_W$. Here we simply trained word represenations from scratch. Various neural network structures can be used to learn context-aware word representations. For example, a recurrent neural network such as a bidirectional LSTM \cite{hochreiter1997long} can be applied to learn representations of each word in the utterance:
\begin{equation}
     \vec{\mathbf{h}}_{t} = \textrm{LSTM}_{fw} (\mathbf{w}_{t}, \vec{\mathbf{h}}_{t-1}),\quad
     {\mathord{\buildrel{\lower3pt\hbox{$\scriptscriptstyle\leftarrow$}}\over {\mathbf{h}}} }_{t} = \textrm{LSTM}_{bw} (\mathbf{w}_{t}, {\mathord{\buildrel{\lower3pt\hbox{$\scriptscriptstyle\leftarrow$}}\over {\mathbf{h}}} }_{t+1}).\\
\end{equation}
For each word \(\mathbf{w}_t\), we concatenate each forward hidden state \(\vec{\mathbf{h}}_{t}\) obtained from the forward $\textrm{LSTM}_{fw}$ with a backward hidden state \({\mathord{\buildrel{\lower3pt\hbox{$\scriptscriptstyle\leftarrow$}}\over 
{\mathbf{h}}} }_{t}\) from $\textrm{LSTM}_{bw}$ to obtain a hidden state \(\mathbf{h}_t\). The whole hidden state matrix can be defined as \(\mathbf{H} = \left(\mathbf{h}_{1}, \mathbf{h}_{2}, ..., \mathbf{h}_{T}\right) \in \mathbb{R}^{T \times 2D_{H}}\), where $D_H$ is the number of hidden units in each LSTM. 
In this work, the parameters of WordCaps are trained with the whole model, while sophisticated pre-trained models such as ELMo \cite{peters2018deep} or BERT \cite{devlin2018bert} may also be integrated.
 
\subsection{\SecondCapsule}
Traditionally, the learned hidden state \(\mathbf{h}_t\) for each word \(\mathbf{w}_t\) is used as the logit to predict its slot tag. When $\mathbf{H}$ for all words in the utterance is learned, sequential tagging methods like the linear-chain CRF models the tag dependencies by assigning a transition score for each transition pattern between adjacent tags to ensure the best tag sequence of the utterance from all possible tag sequences.

Instead of doing slot filling via sequential labeling which does not directly consider the dependencies among words, the {\SecondCapsule} learn to recognize slots via dynamic routing. The routing-by-agreement explicitly models the hierarchical relationship between capsules to address the task taxonomy explicitly. For example, the routing-by-agreement mechanism send a low-level feature, e.g. a word representation in {\FirstCapsule}, to high-level capsules, e.g. {\SecondCapsule}, only when the word representation has a strong agreement with a slot representation.

The agreement value on a word may vary when being recognized as different slots. For example, the word \texttt{three} may be recognized as a \texttt{party\_size\_number} slot or a \texttt{time} slot.  The {\SecondCapsule} first convert the word representation obtained in {\FirstCapsule} with respect to each slot type. We denote \(\mathbf{p}_{k|t}\) as the resulting prediction vector of the $t$-th word when being recognized as the $k$-th slot:
\begin{equation}
    \mathbf{p}_{k|t} = \sigma(\mathbf{W}_{k}\mathbf{h}_{t}^{T} + \mathbf{b}_{k}),
\end{equation}
where $k\in\{1, 2, ..., K\}$ denotes the slot type and $t\in\{1, 2, ..., T\}$. $\sigma$ is the activation function such as $tanh$. $\mathbf{W}_{k}\in \mathbb{R}^{{D_P}\times{2D_H}}$ and $\mathbf{b}_{k} \in \mathbb{R}^{D_P\times 1}$ are the weight and bias matrix for the $k$-th capsule in {\SecondCapsule}, and ${D_P}$ is the dimension of the prediction vector.

\noindent\textbf{Slot Filling by Dynamic Routing-by-agreement}
We propose to determine the slot type for each word by dynamically route prediction vectors of each word from {\FirstCapsule} to {\SecondCapsule}.
The dynamic routing-by-agreement learns an agreement value \(c_{kt}\) that determines how likely the $t$-th word agrees to be routed to the $k$-th slot capsule.  \(c_{kt}\) is calculated by the dynamic routing-by-agreement algorithm \cite{sabour2017dynamic}, which is briefly recalled in Algorithm \ref{al::routing}.

\begin{algorithm}[ht]
\caption{Dynamic routing-by-agreement}\label{al::routing}
\begin{minipage}{\linewidth}
\begin{algorithmic}[1]
\Procedure{Dynamic\_Routing}{$\mathbf{p}_{k|t}$, $iter$}
    \State {for each {\FirstCapsule} t and {\SecondCapsule} k: ${b}_{kt}\leftarrow0$.}
    
    \For {$iter$~iterations}
        \State {for all {\FirstCapsule} $t$: $\mathbf{c}_t\leftarrow\operatorname{softmax}(\mathbf{b}_t)$}
        
        \State {for all {\SecondCapsule} k: $\mathbf{s}_k \leftarrow \Sigma_r c_{kt}\mathbf{p}_{k|t}$}
        
        \State {for all {\SecondCapsule} k: $\mathbf{v}_k= \operatorname{squash}(\mathbf{s}_{k})$}
        
        \State {for all {\FirstCapsule} t and {\SecondCapsule} k: ${b}_{kt} \leftarrow \text{b}_{kt} + \mathbf{p}_{k|t}\cdot\mathbf{v}_{k}$}
        
	    \EndFor
	\State Return $\mathbf{v}_k$
\EndProcedure
\end{algorithmic}
\end{minipage}
\end{algorithm}

The above algorithm determines the agreement value $c_{kt}$ between {\FirstCapsule} and {\SecondCapsule} while learning the slot representations $\mathbf{v}_k$ in an unsupervised, iterative fashion. $\mathbf{c}_t$ is a vector that consists of all $c_{kt}$ where $k\in{K}$.
${b}_{kt}$ is the logit (initialized as zero) representing the log prior probability that the $t$-th word in {\FirstCapsule} agrees to be routed to the $k$-th slot capsule in {\SecondCapsule} (Line 2). During each iteration (Line 3), each slot representation $\mathbf{v}_k$ is calculated by aggregating all the prediction vectors for that slot type $\{{\mathbf{p}_{k|t}} | t{\in}T\}$, weighted by the agreement values $c_{kt}$ obtained from $b_{kt}$ (Line 5-6):
\begin{equation}
    \mathbf{s}_{k} = \sum_{t}^{T}c_{kt}\mathbf{p}_{k|t},
\end{equation}
\begin{equation}
    \mathbf{v}_{k} = \operatorname{squash}(\mathbf{s}_{k}) = \frac{\left \| \mathbf{s}_{k} \right \|^{2}}{1 + \left \| \mathbf{s}_{k}\right \|^{2} }\frac{\mathbf{s}_{k}}{\left \| \mathbf{s}_{k} \right \|},
\end{equation}
where a squashing function $\operatorname{squash}(\cdot)$ is applied on the weighted sum \(\mathbf{s}_{k}\) to get \(\mathbf{v}_{k}\) for each slot type.
Once we updated the slot representation $\mathbf{v}_k$ in the current iteration, the logit $b_{kt}$ becomes larger when the dot product $\mathbf{p}_{k|t}\cdot\mathbf{v}_{k}$ is large. That is, when a prediction vector $\mathbf{p}_{k|t}$ is more similar to a slot representation $\mathbf{v}_k$, the dot product is larger, indicating that it is more likely to route this word to the $k$-th slot type (Line 7). An updated, larger $b_{kt}$ will lead to a larger agreement value $c_{kt}$ between the $t$-th word and the $k$-th slot in the next iteration.
On the other hand, it assigns low $c_{kt}$ when there is inconsistency between $p_{k|t} $ and $\mathbf{v}_k$.
The agreement values learned via the unsupervised, iterative algorithm ensures the outputs of the {\FirstCapsule} get sent to appropriate subsequent {\SecondCapsule} after $iter_{\operatorname{slot}}$ iterations.

\noindent\textbf{Cross Entropy Loss for Slot Filling}\\
For the $t$-th word in an utterance, its slot type is determined as follows:
\begin{equation}
{\hat y}_t = \mathop {\arg \max }\limits_{k \in K} ({{c}_{kt}}).
\end{equation}
The slot filling loss is defined over the utterance as the following cross-entropy function:
\begin{equation}
\mathcal{L}_{slot} = - \sum\limits_t {\sum\limits_k {y_t^k\log (\hat y_t^k)} },
\end{equation}
where $y_t^k$ indicates the ground truth slot type for the $t$-th word. $y_t^k=1$ when the $t$-th word belongs to the $k$-th slot type.

\subsection{\ThirdCapsule}
The {\ThirdCapsule} take the output $\mathbf{v}_k$ for each slot $k \in \{1,2,...,K\}$ in {\SecondCapsule} as the input, and determine the utterance-level intent of the whole utterance.
The {\ThirdCapsule} also convert each slot representation in {\SecondCapsule} with respect to the intent type:
\begin{equation}
\mathbf{q}_{l|k} = \sigma(\mathbf{W}_{l}\mathbf{v}_k^T +b_{l}),
\end{equation}
where $l \in \{1,2,...,L\}$ and $L$ is the number of intents. $\mathbf{W}_l \in \mathbb{R}^{D_L\times D_P}$ and $\mathbf{b}_l \in \mathbb{R}^{D_L\times 1}$ are the weight and bias matrix for the $l$-th capsule in {\ThirdCapsule}.

{\ThirdCapsule} adopt the same dynamic routing-by-agreement algorithm, where:
\begin{equation}
\mathbf{u}_l = \textsc{Dynamic\_Routing}(\mathbf{q}_{l|k}, iter_{\operatorname{intent}}).
\end{equation}

\noindent\textbf{Max-margin Loss for Intent Detection}\\
Based on the capsule theory, the orientation of the activation vector $\mathbf{u}_l$ represents intent properties while its length indicates the activation probability. 
The loss function considers a max-margin loss on each labeled utterance:
\begin{align}\label{eq:loss}
\mathcal{L}_{intent}&= \sum_{l=1}^{L}\{
\left[\kern-0.15em\left[ {{z} = z_l} \right]\kern-0.15em\right]\cdot \max (0,m^ +  - {{\left\| {{\mathbf{u}_l}} \right\|}})^2 + \lambda \left[\kern-0.15em\left[ {{z} \ne z_l} \right]\kern-0.15em\right]\cdot \max (0,{{\left\| {{\mathbf{u}_l}} \right\|}} - m^ - )^2\},
\end{align}
where $\|\mathbf{u}_l\|$ is the norm of $\mathbf{u}_l$ and $\left[\kern-0.15em\left[  \right]\kern-0.15em\right]$ is an indicator function, $z$ is the ground truth intent label for the utterance $x$. $\lambda$ is the weighting coefficient, and $m^ +$ and $m^ -$ are margins.

The intent of the utterance can be easily determined by choosing the activation vector with the largest norm ${{\hat z}} = \mathop {\arg \max }\limits_{l \in \{ 1,2,...,L\} } \left\| {{{\mathbf{u}}_l}} \right\|$.

\subsection{Re-Routing}
The {\ThirdCapsule} not only determine the intent of the utterance by the length of the activation vector, but also learn discriminative intent representations of the utterance by the orientations of the activation vectors.
Previously, the dynamic routing-by-agreement shows how low-level features such as slots help construct high-level ideas such as intents. While the high-level features also work as a guide that helps learn low-level features. For example, the \texttt{AddToPlaylist} intent activation vector in {\ThirdCapsule} also helps strength the existing slots such as \texttt{artist\_name} during slot filling on the words \texttt{Sungmin} in {\SecondCapsule}.

Thus we propose a re-routing schema for {\SecondCapsule} where the dynamic routing-by-agreement is realized by the following equation that replaces the Line 7 in Algorithm \ref{al::routing}:
\begin{equation}
\text{b}_{kt} \leftarrow \text{b}_{kt} + \mathbf{p}_{k|t}\cdot\mathbf{v}_{k} + \alpha \cdot \mathbf{p}_{k|t}^{T}{\mathbf{W}_{RR}}\mathbf{\hat u}_{\hat z}^{T},
\end{equation}
where $\mathbf{\hat u}_{\hat z}$ is the intent activation vector with the largest norm.
$\mathbf{W}_{RR}\in\mathbb{R}^{D_P \times D_L}$ is a bi-linear weight matrix, and $\alpha$ as the coefficient.
The routing information for each word is updated toward the direction where the prediction vector not only coincides with representative slots, but also towards the most-likely intent of the utterance.
As a result, the re-routing makes {\SecondCapsule} obtain updated routing information as well as updated slot representations.
\section{Evaluation}
To demonstrate the effectiveness of our proposed models, we compare the proposed model {\ModelName} with existing alternatives, as well as commercial natural language understanding services.

\subsection{Datasets}
For each task, we evaluate our proposed models by applying it on two real-word datasets: SNIPS Natural Language Understanding benchmark\footnote{\url{https://github.com/snipsco/nlu-benchmark/}} (SNIPS-NLU) and the Airline Travel Information Systems (ATIS) dataset \cite{tur2010left}. The statistical information on two datasets are shown in \ref{data_statistics}.

\begin{table}[h!]
\centering
\begin{tabular}{l|ll}
\hline
\textbf{} & \textbf{SNIPS-NLU} & \textbf{ATIS} \\\hline   
Vocab Size   &  11,241 & 722\\
Average Sentence Length   &  9.05    & 11.28\\
\#Intents  &  7   & 21\\
\#Slots  &  72   & 120\\
\#Training Samples   &  13,084    &  4,478\\
\#Validation Samples   &  700    &  500\\
\#Test Samples   &  700    &  893\\\hline
\end{tabular}
\caption{Dataset statistics.}
\label{data_statistics}
\end{table}

SNIPS-NLU contains natural language corpus collected in a crowdsourced fashion to benchmark the performance of voice assistants.
ATIS is a widely used dataset in spoken language understanding, where audio recordings of people making flight reservations are collected.

\subsection{Experiment Settings}
\noindent\textbf{Baselines}
We compare the proposed capsule-based model {\ModelName} with other alternatives:

\begin{itemize}
    \item \textbf{CNN TriCRF} \cite{xu2013convolutional} introduces a Convolution Neural Network (CNN) based sequential labeling model for slot filling. The hidden states for each word are summed up to predict the utterance intent. We adopt the performance with lexical features.
    \item \textbf{Joint Seq.} \cite{hakkani2016multi} adopts a Recurrent Neural Network (RNN) for slot filling and the last hidden state of the RNN is used to predict the utterance intent. 
    \item \textbf{Attention BiRNN} \cite{liu2016attention} further introduces a RNN based encoder-decoder model for joint slot filling and intent detection. An attention weighted sum of all encoded hidden states is used to predict the utterance intent.
    \item \textbf{Slot-gated Full Atten.} \cite{goo2018slot} utilizes a slot-gated mechanism as a special gate function in Long Short-term Memory Network (LSTM) to improve slot filling by the learned intent context vector. The intent context vector is used for intent detection.
    \item \textbf{DR-AGG} \cite{gong2018information} aggregates word-level information for text classification via dynamic routing. The high-level capsules after routing are concatenated, followed by a multi-layer perceptron layer that predicts the utterance label. We used this capsule-based text classification model for intent detection only.
    \item \textbf{IntentCapsNet} \cite{xia2018zero} adopts a multi-head self-attention to extract intermediate semantic features from the utterances, and uses dynamic routing to aggregate semantic features into intent representations for intent detection. We use this capsule-based model for intent detection only.
\end{itemize}
 
We also compare our proposed model {\ModelName} with existing commercial natural language understanding services, including api.ai (Now called DialogFlow)\footnote{\url{https://dialogflow.com/}}, Waston Assistant\footnote{\url{https://www.ibm.com/cloud/watson-assistant/}}, Luis\footnote{\url{https://www.luis.ai/}}, wit.ai\footnote{\url{https://wit.ai/}}, snips.ai\footnote{\url{https://snips.ai/}}, recast.ai\footnote{\url{https://recast.ai/}}, and Amazon Lex\footnote{\url{https://aws.amazon.com/lex/}}.

\noindent\textbf{Implementation Details}
The hyperparameters used for experiments are shown in \ref{hyperparameter}.

\begin{table}[ht!]
\centering
\begin{tabular}{l|llllll}
\hline
\textbf{DATASET}    & $D_W$ & $D_H$ & $D_P$ & $D_L$  &  $iter_{\operatorname{slot}}$ &  $iter_{\operatorname{intent}}$ \\ \hline
SNIPS-NLU &  1024   & 512   &  512 &  128 &   2   &   2    \\ 
ATIS &   1024 &  512  & 512   &  256 &  3  &    3    \\ \hline
\end{tabular}
\caption{Hyperparameter settings.}
\label{hyperparameter}
\end{table}

\begin{table*}[tb!]
\centering
\resizebox{1.0\textwidth}{!}{
\begin{tabular}{l|ccc|ccc}
\hline
\multirow{2}{*}{\textbf{MODEL}} & \multicolumn{3}{c|}{\textbf{SNIPS-NLU}} & \multicolumn{3}{c}{\textbf{ATIS}} \\ \cline{2-7} 
                      & Slot (F1) &Intent (Acc) & Overall (Acc) 
                      & Slot (F1) &Intent (Acc) & Overall (Acc)           \\ \hline
CNN TriCRF \cite{xu2013convolutional} 	&-		&-		&-	&0.944   &-	&-	\\
Joint Seq. \cite{hakkani2016multi}  	&0.873	&0.969	&0.732	&0.942   &0.926	&0.807	\\
Attention BiRNN \cite{liu2016attention} &0.878	&0.967	&0.741	&0.942   &0.911	&0.789	\\
Slot-Gated Full Atten. \cite{goo2018slot}	&0.888	&0.970	&0.755	&0.948   &0.936	&0.822	\\
DR-AGG \cite{gong2018information}       & -		&0.966  	& -		& -   	 &0.914 & -	\\
IntentCapsNet \cite{xia2018zero}  		& -		&0.974	& -		& -   	 &0.948 & -	\\
\hline
{\ModelName} &\textbf{0.918}&0.973 &\textbf{0.809}   &\textbf{0.952} &\textbf{0.950} & \textbf{0.834}\\
{\ModelName} w/o Intent Detection &0.902	& -	& -	&{0.948} &-	& -	\\ 
{\ModelName} w/o Joint Training     &0.902	& \textbf{0.977}	&0.804	&0.948  &0.847	&0.743\\ 
\hline
\end{tabular}
}
\vspace{0.15in}
\caption{Slot filling and intention detection results.}
\label{tab::overall_caps}
\end{table*}
\begin{figure*}[tb!]
    \centering
    \includegraphics[width=\linewidth]{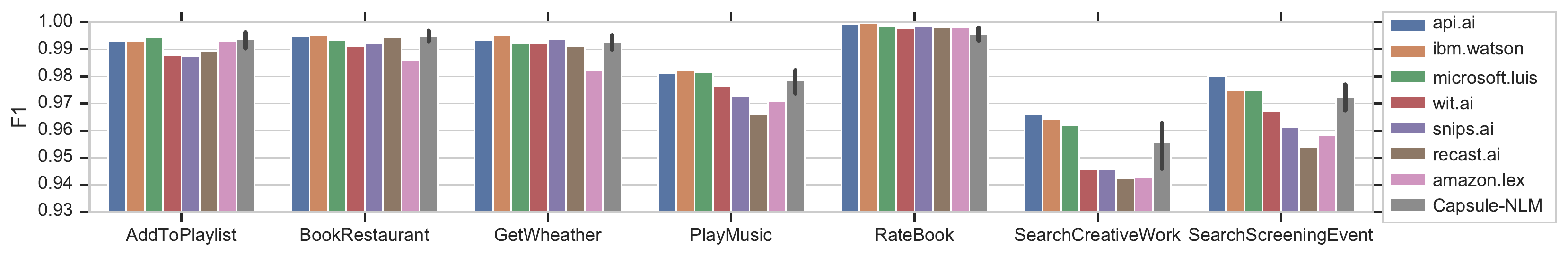}
    \caption{Benchmarking with existing NLU services.}\label{fig::benchmark}
\end{figure*}

We use the validation data to choose hyperparameters.
For both datasets, we randomly initialize word embeddings using Xavier initializer and let them train with the model.
In the loss function, the down-weighting coefficient $\lambda$ is 0.5, margins $m^ +$ and $m^ -$ are set to 0.8 and 0.2 for all the existing intents. $\alpha$ is set as 0.1.
RMSProp optimizer \cite{tieleman2012lecture} is used to minimize the loss.  To alleviate over-fitting, we add the dropout to the LSTM layer with a  dropout rate of 0.2.

\subsection{Experiment Results}
\noindent\textbf{Quantitative Evaluation} The intent detection results on two datasets are reported in \ref{tab::overall_caps}, where the proposed capsule-based model performs consistently better than current learning schemes for joint slot filling and intent detection, as well as capsule-based neural network models that only focuses on intent detection. These results demonstrate the novelty of the proposed capsule-based model {\ModelName} in jointly modeling the hierarchical relationships among words, slots and intents via the dynamic routing between capsules.

Also, we benchmark the intent detection performance of the proposed model with existing natural language understanding services\footnote{https://www.slideshare.net/KonstantinSavenkov/nlu-intent-detection-benchmark-by-intento-august-2017} in \ref{fig::benchmark}.
Since the original data split is not available, we report the results with stratified 5-fold cross validation. From \ref{fig::benchmark} we can see that the proposed model {\ModelName} is highly competitive with off-the-shelf systems that are available to use. Note that, our model archieves the performance without using pre-trained word representations: the word embeddings are simply trained from scratch.

\noindent\textbf{Ablation Study}
To investigate the effectiveness of {\ModelName} in joint slot filling and intent detection, we also report ablation test results in \ref{tab::overall_caps}. ``w/o Intent Detection'' is the model without intent detection: only a dynamic routing is performed between {\FirstCapsule} and {\SecondCapsule} for the slot filling task, where we minimize $\mathcal{L}_{slot}$ during training; ``w/o Joint Training'' adopts a two-stage training where the model is first trained for slot filling by minimizing $\mathcal{L}_{slot}$, and then use the fixed slot representations to train for the intent detection task which minimizes $\mathcal{L}_{intent}$. From the lower part of \ref{tab::overall_caps} we can see that by using a capsule-based hierarchical modeling between words and slots, the model {\ModelName} w/o Intent Detection is already able to outperform current alternatives on slot filling that adopt a sequential labeling schema. The joint training of slot filling and intent detection is able to give each subtask further improvements when the model parameters are updated jointly.

\noindent\textbf{Visualizing Agreement Values between Capsule Layers}
Thanks to the dynamic routing-by-agreement schema, the dynamically learned agreement values between different capsule layers naturally reflect how low-level features are collectively aggregated into high-level ones for each input utterance. In this section, we harness the intepretability of the proposed capsule-based model via hierarchical modeling and provide case studies and visualizations.

\begin{figure}[h]
    \centering
    \includegraphics[width=0.6\linewidth]{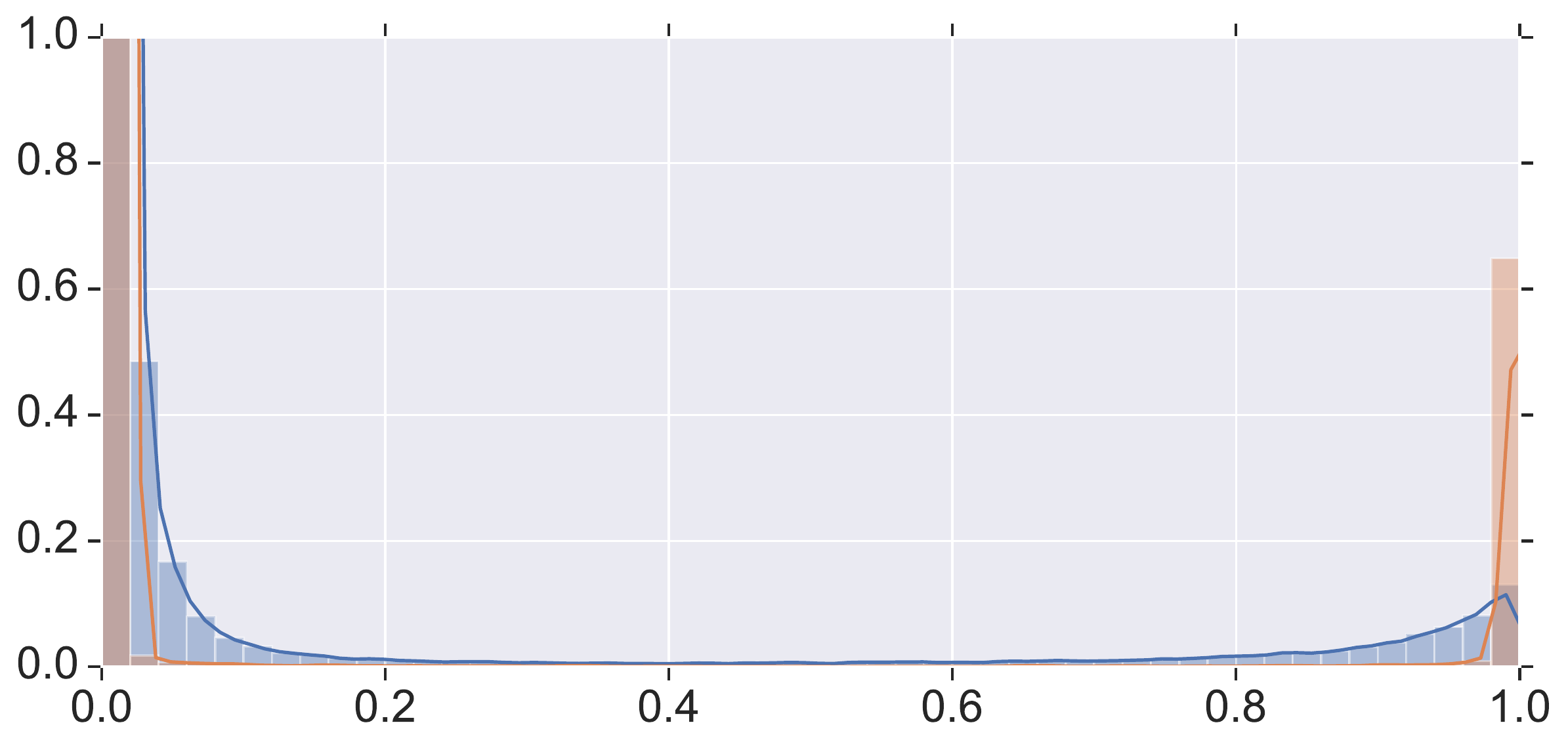}
    \caption{The distribution of agreement values between WordCaps \& SlotCaps.}
    \label{fig:w2s_dist}
\end{figure}

\textbf{Between WordCaps and SlotCaps}
First we study the agreement value $c_{kt}$ between the $t$-th word in the WordCaps and the $k$-th slot capsule in SlotCaps.
\ref{fig:w2s_dist} shows the distribution of all agreement values between WordCaps and SlotCaps on the test split of SNIPS-NLU dataset. Blue bars indicate the distribution of values after the first iteration and orange bars indicate the distribution after the second iteration.
We observe that the dynamic routing-by-agreement is able to converge to an agreement quickly after the first iteration (shown in blue bars). It is able to assign a confident probability assignment close to 0 or 1. After the second iteration (shown in orange bars), the model is more certain about the routing decisions: probabilities are more leaning towards 0 or 1 as the model is confident about routing a word in WordCaps to its most appropriate slot in SlotCaps.

\begin{figure}[ht!]
    \centering
    \includegraphics[width=0.8\linewidth]{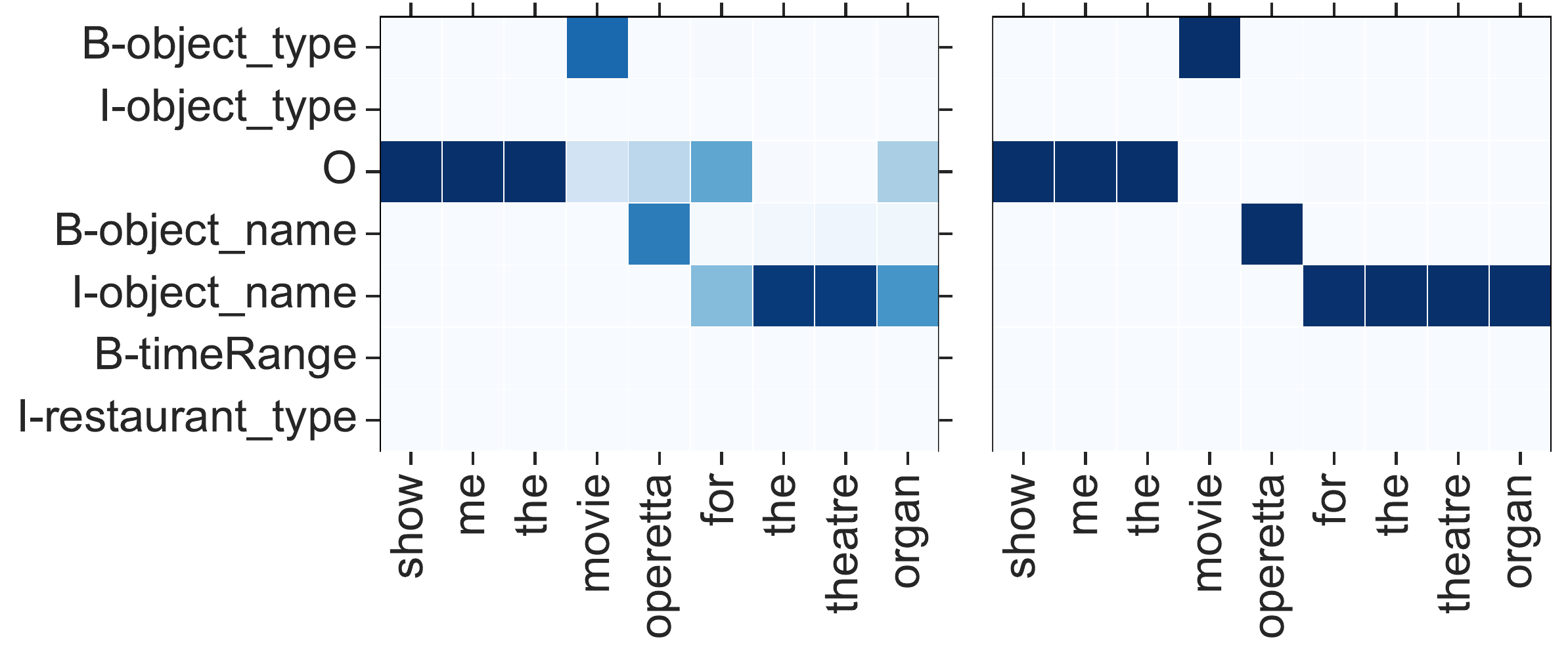}
    \caption{Agreement values between WordCaps (x-axis) and SlotCaps (y-axis).}
    \label{fig:w2s}
\end{figure}

However, we do find that when unseen slot values like new object names emerge in utterances like \texttt{show me the movie operetta for the theatre organ} with an intent of \texttt{SearchCreativeWork}, the iterative dynamic routing process would be even more appealing.
\ref{fig:w2s} shows the agreement values learned by dynamic routing-by-agreement. A sample from the test split of SNIPS-NLU dataset is shown (Left: after the fist routing iteration. Right: after the second iteration). Since the dynamic routing-by-agreement is an iterative process controlled by the variable $iter_{slot}$, we show the agreement values after the first iteration in the left part of \ref{fig:w2s}, and the values after the second iteration in the right part. Due to space limitations, only part of slots (7/72) are shown on the y-axis.
 
From the left part of \ref{fig:w2s}, we can see that after the first iteration, the model considers the word \texttt{operetta} itself alone is likely to be an object name, probably because the following word \texttt{for} is usually a context word being annotated as \texttt{O}. Thus it tends to route word \texttt{for} to both the slot \texttt{O} and the slot \texttt{I-object\_name}. However, from the right part of \ref{fig:w2s} we can see that after the second iteration, the dynamic routing found an agreement and is more certain to have \texttt{operetta for the theatre organ} as a whole for the slot \texttt{B-object\_name} and \texttt{I-object\_name}.

\begin{figure}[ht!]
    \centering
    \includegraphics[width=0.8\linewidth]{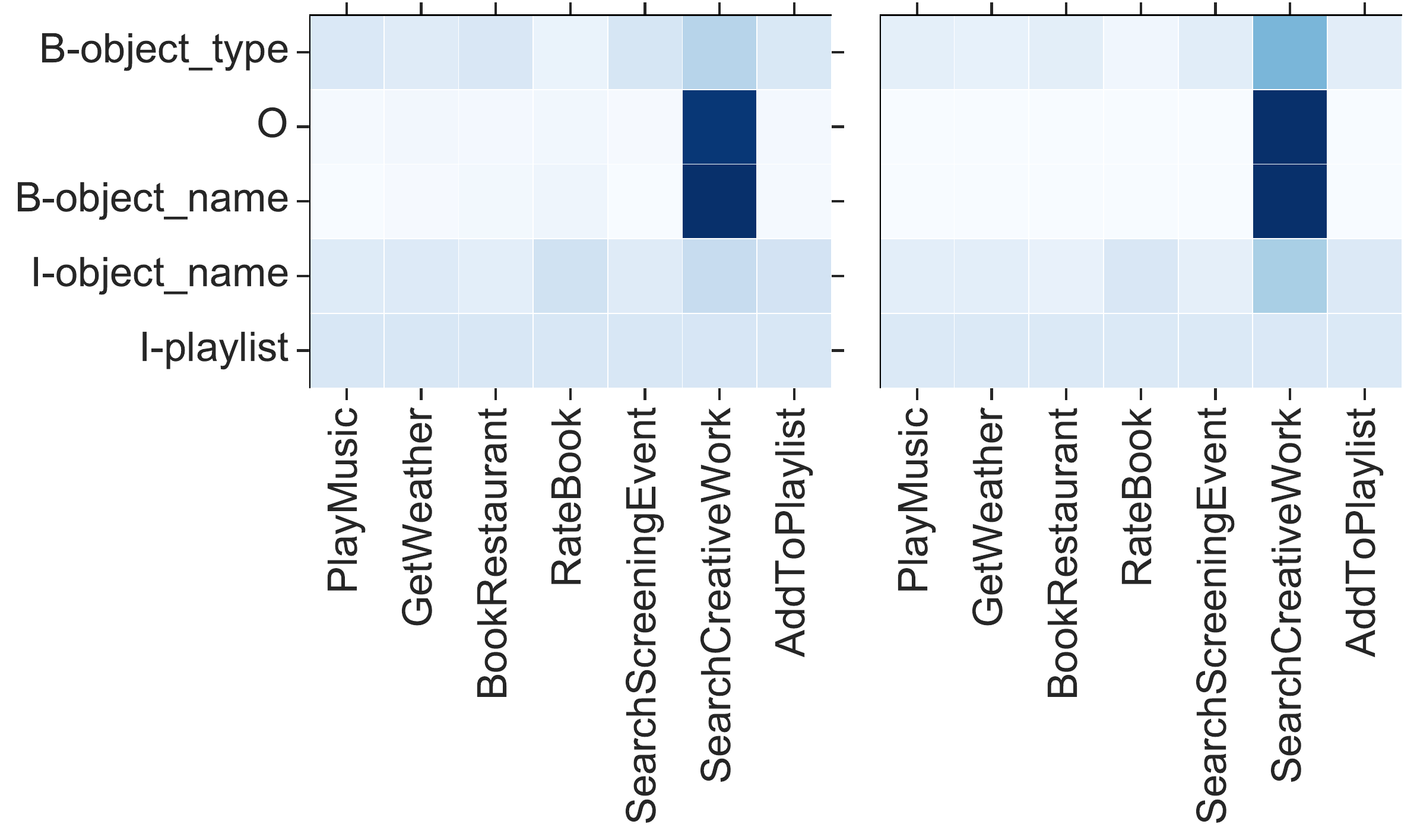}
    \caption{Agreement values between SlotCaps (y-axis) and IntentCaps (x-axis).}
    \label{fig:s2i}
    \vspace{-0.3in}
\end{figure}

\textbf{Between SlotCaps and IntentCaps} Similarly, we visualize the agreement values between each slot capsule in SlotCaps and each intent capsule in IntentCaps.
 The left part of \ref{fig:s2i} shows that after the first iteration, since the model is not able to correctly recognize \texttt{operetta for the theatre organ} as a whole, only the context slot \texttt{O} (correspond to the word \texttt{show me the}) and \texttt{B-object\_name} (correspond to the word \texttt{operetta}) contribute significantly to the final intent capsule. From the right part of \ref{fig:s2i}, we found that with the word  \texttt{operetta for the theatre organ} being recognized in the lower capsule, the slots \texttt{I-object\_name} and \texttt{B-object\_type} contribute more to the correct intent capsule \texttt{SearchCreativeWork}, when comparing with other routing alternatives to other intent capsules.
 
\section{Related Works}
\noindent\textbf{Intent Detection}
With recent developments in deep neural networks, user intent detection models \cite{hu2009understanding,xu2013convolutional,zhang2016mining,liu2016attention,zhang2017bringing,chen2016end,xia2018zero} are proposed to classify user intents given their diversely expressed utterances in the natural language. As a text classification task, the decent performance on utterance-level intent detection usually relies on hidden representations that are learned in the intermediate layers via multiple non-linear transformations. 

Recently, various capsule based text classification models are proposed that aggregate word-level features for utterance-level classification via dynamic routing-by-agreement \cite{gong2018information,zhao2018investigating,xia2018zero}. Among them, \cite{xia2018zero} adopts self-attention to extract intermediate semantic features and uses a capsule-based neural network for intent detection. 
However, existing works do not study word-level supervisions for the slot filling task. In this work, we explicitly model the hierarchical relationship between words and slots on the word-level, as well as intents on the utterance-level via dynamic routing-by-agreement.

\noindent\textbf{Slot Filling}
Slot filling annotates the utterance with finer granularity: it associates certain parts of the utterance, usually named entities, with pre-defined slot tags. Currently, the slot filling is usually treated as a sequential labeling task. A recurrent neural network such as Gated Recurrent Unit (GRU) or Long Short-term Memory Network (LSTM) is used to learn context-aware word representations, and Conditional Random Fields (CRF) are used to annotate each word based on its slot type. Recently, \cite{shen2017disan,tan2017deep} introduce the self-attention mechanism for CRF-free sequential labeling.

\noindent\textbf{Joint Modeling via Sequence Labeling}
To overcome the error propagation in the word-level slot filling task and the utterance-level intent detection task in a pipeline, joint models are proposed to solve two tasks simultaneously in a unified framework.
\cite{xu2013convolutional} propose a Convolution Neural Network (CNN) based sequential labeling model for slot filling. The hidden states corresponding to each word are summed up in a classification module to predict the utterance intent. A Conditional Random Field module ensures the best slot tag sequence of the utterance from all possible tag sequences.
\cite{hakkani2016multi} adopt a Recurrent Neural Network (RNN) for slot filling and the last hidden state of the RNN is used to predict the utterance intent. 
\cite{liu2016attention} further introduce an RNN based encoder-decoder model for joint slot filling and intent detection. An attention weighted sum of all encoded hidden states is used to predict the utterance intent.
Some specific mechanisms are designed for RNNs to explicitly encode the slot from the utterance. For example, \cite{goo2018slot} utilize a slot-gated mechanism as a special gate function in Long Short-term Memory Network (LSTM) to improve slot filling by the learned intent context vector.
However, as the sequence becomes longer, it is risky to simply rely on the gate function to sequentially summarize and compress all slots and context information in a single vector \cite{cheng2016long}.

In this paper, we harness the capsule neural network to learn a hierarchy of feature detectors and explicitly model the hierarchical relationships among word-level slots and utterance-level intent. Also, instead of doing sequence labeling for slot filling, we use a dynamic routing-by-agreement schema between capsule layers to route each word in the utterance to its most appropriate slot type. And we further route slot representations, which are learned dynamically from words, to the most appropriate intent capsule for intent detection.

\chapter{\TaskNameThree}\label{chapter:c4}
This chapter was previously published as ``On the Generative Discovery of Structured Medical Knowledge'', in KDD'18~\cite{zhang2018generative}. DOI: \url{https://doi.org/10.1145/3219819.3220010}.
\newcommand{\KDDModelName}{CRVAE}
\newcommand{\KDDModelNameFull}{Conditional Relationship Variational Autoencoder}
\newcommand{\KDDTaskName}{\TaskNameThree}
\newcommand{\KDDTaskNameFull}{\TaskNameThree}
\newcommand{\textzh}[1]{\begin{CJK*}{UTF8}{gbsn}{#1}\end{CJK*}}
\section{Introduction}
Knowledge Graphs such as WordNet \cite{miller1995wordnet}, Yago \cite{fabian2007yago} and Freebase \cite{bollacker2008freebase} have been playing an essential role in many applications, such as knowledge inference, question answering, relation extraction, and so on. 
A large-scale of structured knowledge is embodied in Knowledge Graphs, in the form of triplets (head entity, tail entity and the relationship, denoted as $h~{\xrightarrow{r}}~t$). 
For example, the $Disease~{\xrightarrow{Cause}}~Symptom$ relationship indicates a ``Cause'' relationship from a disease entity ({\eg} \texttt{synovitis}) to a symptom entity (\eg \texttt{joint pain}) which is caused by this disease. 
Various linguistic expressions are usually observed among different triplets. For example, \texttt{nose plugged}, \texttt{blocked nose} and \texttt{sinus congestion} are symptom entities that share the same meaning but expressed very differently.
The expression diversity is also widely observed for triplets of the same relation: a relationship may also be instantiated by entity pairs in varying granularities or different relationship strength. For instance, Disease $\xrightarrow{Cause}$ Symptom relationship may include coarse-grained entity pairs like $<$\texttt{rhinitis}, \texttt{nose plugged}$>$, while $<$\texttt{acute rhinitis}, \texttt{nose plugged}$>$, $<$\texttt{chronic rhinitis}, \texttt{nose plugged}$>$ are considered as fine-grained entity pairs. As for the relationship strength, $<$\texttt{cold}, \texttt{fatigue}$>$ has greater relationship strength than $<$\texttt{cold}, \texttt{ear infections}$>$ as cold rarely cause serious complications such as ear infections. It is straightforward for human beings yet still challenging for a machine to understand the commonalities between different triplets.

Since most knowledge graphs were built either collaboratively or (partly) automatically \cite{ji2015knowledge}, they are far from complete \cite{socher2013reasoning}. The knowledge graph completion task aims at predicting relationships between entities based on existing triplets in a knowledge graph.
Many works have focused on extending existing knowledge graphs using well-trained classifiers to predict whether or not there is a relationship between two existing or new entities \cite{socher2013reasoning,bordes2013translating,komninos2017feature,trouillon2017knowledge,he2018knowledge}. 
Existing models such as for relation extraction \cite{agichtein2000snowball,baeza2007extracting,jiang2017metapad,liu2017heterogeneous,mintz2009distant,sahay2008discovering,wang2015constrained} or knowledge graph completion \cite{socher2013reasoning,komninos2017feature,trouillon2017knowledge,he2018knowledge,gardner2015efficient,lin2016neural,wang2015knowledge,zeng2014relation} adopt a discriminative setting. 
Although achieving decent performance in identifying the correctness of candidate triplets, their performances rely on well-prepared annotated triplets as the training data, as well as high-quality candidate triplets for testing.
Relation extraction methods aim to examine if a semantic relationship exists between two entities in the given context. And they also require a substantial collection of contexts over a full spectrum of relationships we would like to work on. However, they can be vulnerable to the ``garbage-in, garbage-out'' situation: the meaningful relational triplets for a specific relationship cannot be identified when no high-quality relational triplets having that relationship are among the candidate relational triplets. The choice of candidates may involve additional human annotation, which is tedious and labor-intensive. 
In both tasks mentioned above (Knowledge Graph Completion and Relation Extraction), the lacking preparation of external resource or additional human annotation is fatal to the successful discovery of structured knowledge \cite{ma2019mcvae}. Therefore, it is crucial for us to discover structured knowledge without substantial data requirement.

To reduce human annotation efforts for effective structured knowledge discovery, in this chapter we propose a novel research problem called \TaskNameThree, which aims at understanding each relationship between entities solely from the existing triplets via their diverse expressions. With the help of rich semantic information embodied in entity representations learned from a massive text corpus, we aim to discover meaningful and novel triplets of a specific relationship in a generative fashion, without sophisticated feature engineering and substantial data requirement such as large-scale text corpora as contexts, or further data preparation.

We introduce a generative perspective to increase the scale of high-quality structured knowledge harnessing the massiveness of the unannotated text corpus.
The proposed model explores the generative modeling capacity for entity pairs and their relationships while incorporating deep learning for hands-free feature engineering. It is able to generate meaningful triplets that are not yet observed, which expand the scale of existing structured knowledge.

Specifically, the model takes the triplets as the input. It encodes each triplet r into a latent space conditioned on the relationship type. Based on pre-trained entity representations from a massive text corpus, the encoding process further addresses relationship-enhanced entity representations, entity interactions, and expressive latent variables. The latent variables are decoded to reconstruct both the head and tail entity. Once trained, the generator samples directly from the learned latent variables and decodes them into novel triplets that expand the scale of structured knowledge with minimized additional human annotations.
The performance of the proposed method is evaluated on real-world structured knowledge data in the medical domain both quantitatively and qualitatively.
\section{Preliminaries}
In this section, we briefly review preliminaries that relate to the proposed model.

\noindent\textbf{Autoencoder (AE)}
The traditional autoencoder \cite{bengio2009learning} is a multi-layer non-recurrent neural network architecture which has been widely used for unsupervised representation learning. When given an input data $x$, the autoencoder starts with an encoder net where the input is mapped into a low-dimensional latent variable $z = encoder\_{net}(x)$ through one or more layers of non-linear transformations, followed by a decoder net where the resulting latent variable $z$ is mapped to an output data $x'= decoder\_{net}(z)$ which has the same number of units as the input data $x$, via one or more non-linear hidden layers. The objective of the AE is to minimize the data reconstruction loss:
\begin{equation}
\mathcal{L}_{AE}(x) = \left\| {x - x'} \right\|^2 = \left\| {x - decoder\_net}(encoder\_net(x)) \right\|^2,
\end{equation}
and the resulting latent variable $z$ is the low-dimensional latent feature learned from the data $x$ in a totally unsupervised fashion.

\noindent\textbf{Variational Autoencoder (VAE)}
The concept of automatic encoding and decoding makes AE suitable for generative models.
Unlike the traditional autoencoder \cite{bengio2009learning} where the hidden variable $z$ has unspecified distributions, the variational autoencoder (VAE) \cite{kingma2014stochastic} roots in Bayesian inference and inherits the architecture of AE to encode the Bayes automatically for an expressive generation. VAE assumes that the input data $x$ can be encoded into a set of latent variables $z$ with certain distributions, such as multivariate Gaussian distributions. The resulting Gaussian latent variables $z$ are generated by the generative distribution $P_\theta(z)$ and $x'$ is generated with a Bayesian model by a conditional distribution on $z$: $P_\theta(x'|z)$.
VAE infers the latent distribution $P(z)$ using $P_\theta(z|x)$. $P_\theta(z|x)$ can be considered as some mapping from $x$ to $z$, which is inferred by variational inference as one of the popular Bayesian inference methods. In VAE, $P_\theta(z|x)$ is usually inferred using a simpler distribution $Q_\phi(z|x)$ such as a Gaussian distribution.
The objective of VAE is to optimize its variational lower bound:
\begin{equation}
\mathcal{L}_{VAE}(x,y;\theta ,\phi ) =  - KL\left[ {{Q_\phi }\left( {z|x} \right)||{P_\theta }\left(z|x \right)} \right] +  {\log \left( {{P_\theta }\left( {x} \right)} \right)},
\end{equation}
where the first term uses the KL-divergence to minimize the difference between the simple distribution $Q_\phi(z|x)$ and its true distribution $P_\theta(z|x)$, while the second term maximizes the $log\left( P_\theta(x) \right)$.

\noindent\textbf{Conditional Variational Autoencoder (CVAE)}
Although the VAE can generate data that belongs to different types, the latent variable $z$ is only modeled by $x$ in $P_\theta(z|x)$ without knowing the type of it. Thus it cannot generate an output $x'$ that belongs to a particular type $y$. The conditional variational autoencoder (CVAE) \cite{sohn2015learning} is an extension to VAE that generates $x'$ with conditions.
CVAE models both the data $x$ and latent variables $z$. However, both $x$ and $z$ are conditioned on a class label $y$:
\begin{equation}
\mathcal{L}_{CVAE}(x,y;\theta ,\phi ) =  - KL\left[ {{Q_\phi }\left( {z|x,y} \right)||{P_\theta }\left(z|x \right)} \right] +  {\log \left( {{P_\theta }\left( {x|y} \right)} \right)}.
\end{equation}
In this way, the real latent variable is distributed under $P_\theta(z|y)$ instead of $P_\theta(z)$. With such appealing formulation, we can have a separate $P_\theta(z|y)$ for each class $y$.
\section{Proposed Approach}
\begin{figure}[bt!]
\centering
\epsfig{file=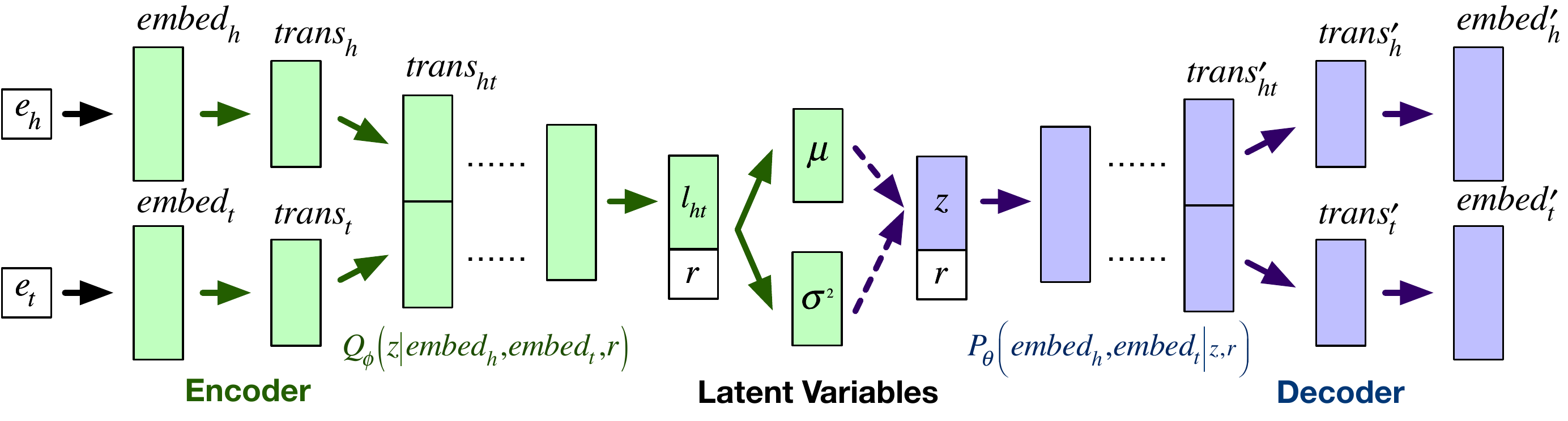, width=6in}
\caption{An overview of the proposed model {\KDDModelName} during training. }\label{fig::model_overview_vae}
\end{figure}

In this section, we introduce the {\KDDModelNameFull} (\KDDModelName) model for the {\KDDTaskName} problem. The proposed model consists of three modules: encoder, decoder, and generator. The encoder module takes entity pairs and their relationship indicator as the input, trained to enhance entity representations and encode the diversely expressed entity pairs for each relationship to a latent space as $Q_\phi$. The decoder is jointly trained to reconstruct the entity pairs as $P_\theta$. The generator model shares the same structure with the decoder. However, instead of reconstructing the relational entity pair given in the input, it directly samples from the learned latent variable distribution to generate meaningful relational entity pairs for a particular relationship. 
\ref{fig::model_overview_vae} gives an overview of the proposed model, where the encoder module is show in green color and the decoder module is show in blue. Model inputs are in white color.

The model takes a tuple $<$$e_h, e_t$$>$ and a relationship indicator $r$ as the input, where $e_h$ and $e_t$ are head and tail entity of a relationship $r$. For example, $e_h=$``\textit{synovitis}'' and $e_t$=``\textit{joint pain}'', while the corresponding $r$ is an indicator for \textit{Disease}~$\xrightarrow{Cause}~$\textit{Symptom}.

To effectively represent entities, pre-trained word embeddings that embody rich semantic information can be obtained as initial entity representations for $e_h$ and $e_t$. For simplicity, we adopt 200-dimensional word embeddings pre-trained using Skip-gram \cite{mikolov2013efficient}. 
After a table lookup on the pre-trained word vector matrix $W_{embed} \in \mathbb{R}^{V \times D_E}$ where $V$ is the vocabulary size (usually tens of thousands) and $D_E$ is the dimension of the initial entity representation (usually tens or hundreds), $embed_h \in \mathbb{R}^{1 \times D_E}$ and $embed_t \in \mathbb{R}^{1 \times D_E}$ are derived as the initial embedding of entities.

\subsection{Encoder}
With the initial entity representation ${embed}_h$ and ${embed}_t$ and their relationship indicator $r$, the encoder first translates and then maps entity pairs to a latent space as $Q_\phi(z|{embed}_h,{embed}_t,r)$.
\noindent\textbf{Translating for Relationship-enhancing}
The initial embedding obtained from word embedding reflects semantic and categorical information. However, it is not specifically designed to model the relationship between entities.

To get entity representations that address relationship information, the encoder learns to translate each entity from its initial embedding space to a relationship-enhanced embedding space that distills relational commonalities. For example, a non-linear transformation can be used: $translate(x) = f(x \cdot W_{trans}+b_{trans})$ where $f$ can be an non-linear activation function such as the Exponential Linear Unit (ELU) \cite{clevert2015fast}. $W_{trans} \in \mathbb{R}^{D_E \times D_R}$ is the weight variable and $b_{trans} \in \mathbb{R}^{1 \times D_R}$ is the bias where $D_R$ is the dimension for relationship-enhanced embeddings.
\begin{equation}
trans_h = translate(embed\_h),\quad trans_t = translate(embed\_t)    
\end{equation}
are obtained as relationship-enhanced embeddings for $e_h$ and $e_t$.

\noindent\textbf{Mapping to Latent Variables}
The relationship-enhanced entity representation $trans_h$ and $trans_t$ are concatenated
\begin{equation}
trans_{ht} = \left[trans_h, trans_t\right]
\end{equation} and mapped to the latent space by multiple fully connected layers.
For example, we can obtain a variable $l_{ht}$ that addresses the relationship information, as well as entity interactions from two medical entities, by applying three consecutive non-linear fully connected layers on $trans_{ht}$.

As a variational inference model, we assume a simple Gaussian distribution of $Q_\phi(z|{embed}_h,{embed}_t,r)$ for the entity pairs $<$$e_h, e_t$$>$ with a relationship $r$. Therefore, for each entity pair $<$$e_h, e_t$$>$ and a relationship indicator $r$, a mean vector $\mu$ and a variance vector $\sigma^2$ can be learned as latent variables to model $Q_\phi(z|{embed}_h,{embed}_t,r)$:
\begin{equation}
{\mu} = \left[l_{ht}, r\right] \cdot W_\mu + b_\mu,\quad{\sigma^2} = \left[l_{ht}, r\right] \cdot W_\sigma + b_\sigma,
\end{equation}
where a one-hot indicator $r \in \mathbb{R}^{1 \times |R|}$ is used for the relationship $r$ and $|R|$ is the number of all relationships. $W_\mu, W_\sigma \in \mathbb{R}^{\left(D_{l_{ht}} + |R|\right) \times D_L}$ are weight terms and $b_\mu, b_\sigma \in \mathbb{R}^{1 \times D_L}$ are bias terms. $D_L$ is the dimension for latent variables and $D_{l_{ht}}$ is the dimension for $l_{ht}$. To stabilize the training, we model the variation vector $\sigma^2$ by its log form $\log \sigma^2$ (to be explained in \ref{eq::log_sigma}).

\subsection{Decoder}
Once we obtain latent variables $\mu$, $\sigma^2$ for an input tuple $<$$e_h, e_t$$>$ which has the relationship $r$, the decoder uses latent variables and the relationship indicator $r$ to reconstruct the relational medical entity pair.
The decoder implements the $P_\theta({embed}_h, {embed}_t|z, r)$.

Given $\mu$, $\sigma^2$, it is intuitive to sample the latent value $z$ from the distribution $N(\mu, \sigma^2)$ directly. 
However, such operator is not differentiable thus optimization methods failed to calculate its gradient. To solve this problem, a reparameterization trick is introduced in \cite{kingma2014stochastic} to divert the non-differentiable part out of the network. 
Instead of directly sampling from $N(\mu, \sigma^2)$, we sample from a standard normal distribution $\epsilon \sim N(0,{\textrm I})$ and convert it back to $z$ by $z = \mu + \sigma \epsilon$.
In this way, sampling from $\epsilon$ does not depend on the network.

Similarly as the use of multiple non-linear fully connected layers for the mapping in the encoder, multiple non-linear fully connected layers are used for an inverse mapping in the decoder. After the inverse mapping we obtain $trans'_{ht} \in \mathbb{R}^{1 \times 2 D_R}$.
The first $D_R$ dimensions of $trans'_{ht}$ are considered as a decoded relationship-enhanced embedding for $e_h$, while the last $D_R$ dimensions are for $e_t$:
\begin{equation}
trans'_h = {trans'_{ht}}\left[:D_R\right],\quad trans'_t = {trans'_{ht}}\left[D_R:\right],
\end{equation}
where ${trans'_h}, {trans'_t} \in \mathbb{R}^{1 \times D_R}$.
${trans'}_h$ and ${trans'}_t$ are further inversely translated back to the initial embedding space $\mathbb{R}^{D_E}$:
\begin{equation}
{embed'_h} = f({trans}'_h \cdot W_{trans\_inv} + b_{trans\_inv}),\quad{embed'_t} = f({trans}'_t \cdot W_{trans\_inv} + b_{trans\_inv}),
\end{equation}
where ${embed'_h}, {embed'_t} \in \mathbb{R}^{1 \times D_E}$ are considered as reconstructed representations for $embed_h$ and $embed_t$.

\subsection{Training}\label{sec::loss_function}
Inspired by the loss function of CVAE, the loss function of \KDDModelName~ is formulated to minimize the variational lower bound:
\begin{equation}\label{eq::base}
\begin{aligned}
&\mathcal{L}_{\KDDModelName}(embed_h,embed_t,r;\theta ,\phi ) =  \\&- KL\left[ {{Q_\phi }\left( {z|embed_h,embed_t,r} \right)||{P_\theta }\left(z|embed_h,embed_t,r \right)} \right] + {\log \left( {{P_\theta }\left( {embed_h,embed_t|r} \right)} \right)}.
\end{aligned}
\end{equation}

The first term minimizes the KL divergence loss between the unknown true distribution ${P_\theta }\left(z|embed_h,embed_t,r \right)$ and a simple distribution ${Q_\phi }\left( {z|embed_h,embed_t,r} \right)$. The second term models the entity pairs by ${\log \left( {{P_\theta }\left( {embed_h,embed_t|r} \right)} \right)}$.
The above equation can be reformulated as:
\begin{equation}\label{eq::first_term}
\begin{aligned}
&\mathcal{L}_{\KDDModelName}(embed_h,embed_t,r;\theta ,\phi ) =  \\
&- KL\left[ {{Q_\phi }\left( {z|embed_h, embed_t, r} \right)||{P_\theta }\left(z | r\right)} \right] + \mathbb{E} \left[ {\log \left( {{P_\theta }\left( {embed_h, embed_t|z, r} \right)} \right)} \right],
\end{aligned}
\end{equation}
where ${P_\theta }\left(z | r\right)$ describes the true latent distribution $z$ given a certain relationship $r$ and 
\begin{equation}
\mathbb{E} \left[ {\log \left( {{P_\theta }\left( {embed_h, embed_t|z, r} \right)} \right)} \right]    
\end{equation}
estimates the maximum likelihood. Since we want to sample from $P_\theta(z|r)$ in the generator, the first term aims to let $Q_\phi(z|embed_h, embed_t, r)$ be as close as possible to $P_\theta(z|r)$ which has a simple distribution $N(0,{\textrm I})$ so that it is easy to sample from. Furthermore, if $P_\theta(z|r) \sim N(0,{\textrm I})$ and $Q(z|embed_h, embed_t, r) \sim N(\mu, \sigma^2)$, then a closed-form solution for the first term in \ref{eq::base} is derived as:
\begin{equation}\label{eq::log_sigma}
\begin{aligned}
- KL\left[ {{Q_\phi }\left( {z|embed_h, embed_t, r} \right)||{P_\theta }\left(z | r\right)} \right] = - KL\left[ { N(\mu, \sigma)||N(0,{\textrm I)}} \right]  \\
= - \frac{1}{2}( tr(\sigma^2  ) + \mu^T \mu  - D_L -\log \det(\sigma ^2)) = - \frac{1}{2} \sum\limits_{l}^{D_L} ({\sigma^2_l+\mu^2_l-1-\log \sigma^2_l} ), \\
\end{aligned}
\end{equation}
where $l$ in the subscript indicates the $l$-th dimension of the vector.
Since it is more stable to have exponential term than a log term, we model $\log\left( \sigma^2 \right)$ as $\sigma^2$ which results in the final closed-form of \ref{eq::log_sigma}:
\begin{equation}
- \frac{1}{2} \sum\limits_{l}^{D_L} {\left( \exp\left(\sigma^2  \right)_l + \mu^2_l -1 - \sigma^2_l \right)}.
\end{equation}

The second term in \ref{eq::base} penalizes the maximum likelihood, where is the conditional probability $P_\theta(embed_h, embed_t|z, r)$ of a certain entity pair $<$$e_h, e_t$$>$ given the latent variable $z$ and the relationship indicator $r$. The mean squared error (MSE) is adopted to calculate the difference between $<$$embed_h, embed_t$$>$ and $<$${embed'_h}, {embed'_t}$$>$:
\begin{equation}
\begin{aligned}
&\mathbb{E} \left[ {\log \left( {{P_\theta }\left( {embed_h,embed_t|z, r} \right)} \right)} \right] =\\
&\frac{1}{2D_E} \left( {{|| {embed_h} - {embed'_h} ||}^2_2 
+ {{|| {embed_t} - {embed'_t} ||}^2_2} }\right), 
\end{aligned}
\end{equation}
where ${\left\|  \cdot  \right\|_2}$ is the vector $\ell_2$ norm.

To minimize the $\mathcal{L}_{\KDDModelName}$, existing gradient-based optimizers such as Adadelta \cite{zeiler2012adadelta} can be used. Furthermore, a warm-up technique introduced in \cite{sonderby2016train} can let the training start with deterministic and gradually switch to variational, by multiplying $\beta$ to the first term. The final loss function used for training is formulated as:
\begin{equation}
\begin{aligned}
&\mathcal{L}_{\KDDModelName} = - \frac{\beta}{2} \sum\limits_{l}^{D_L} \left({\exp\left(\sigma^2  \right)_l + \mu^2_l -1 - \log \sigma^2_l} \right) \\
&+ \frac{1}{2D_E} \left( {{|| {embed_h} - {embed'_h} ||}^2_2 
+ {{|| {embed_t} - {embed'_t} ||}^2_2} }\right),
\end{aligned}
\end{equation}
where $\beta$ is initialized as 0 and increase by 0.1 at the end of each training epoch, until it reaches 1.0 as its maximum.

\subsection{Generator}
When we would like to generate entity pairs of a specific relationship, a density-based sampling method is introduced for the generator to sample $\hat z$ from the distribution of latent variables conditioned on that relationship $r$. 

\begin{figure}[hbt!]
\centering
\epsfig{file=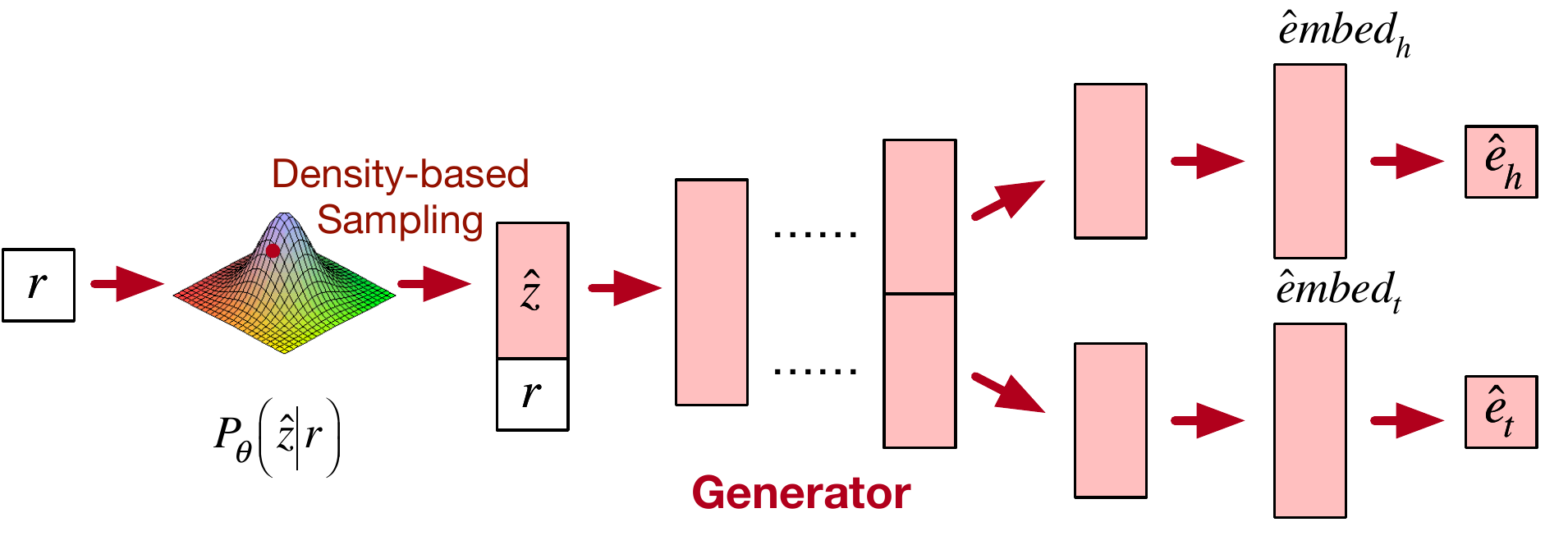, width=4.5in}
\caption{An overview of the proposed model {\KDDModelName} during generation.}\label{fig::generator}
\end{figure}

Instead of using the latent variable $z$ provided by certain $\mu$ and $\log\sigma^2$ in the encoding process from a certain $e_h, e_t$ and $r$, the generator tries to sample $\hat z$ directly from $P_\theta({\hat z}|r)$ to get the latent space value $\hat z$ for a particular relationship $r$. Once $\hat z$ is obtained, the decoder structure is used to decode the entity pair.
\ref{fig::generator} illustrates the generative process.
The denser region in the latent space $P_\theta({\hat z}| r)$ indicates that more densely entity pairs are located in the manifold. Therefore, a sampling method that considers the density distribution of $P_\theta({\hat z}|r)$ samples more often from that region to preserve the true latent space distribution. Specifically, for each relationship $r$, the density-based sampling samples ${\hat z}$ directly from $P_\theta({\hat z}|r) \sim N(0,{\textrm I})$, when trained properly.
The resulting vectors ${\hat embed}_h$ and ${\hat embed}_t$ are mapped back to entity names in natural language, namely ${\hat e_t}$ and ${\hat e_h}$, by finding the nearest neighbor in their initial embedding space $\mathbb{R}^{1 \times D_E}$ using $W_{embed}$. The $\ell$-2 distance measure is used for the nearest neighbor search. 

Note that the vocabulary of pre-trained word embedding is way more comprehensive than entities from labeled triplets in training. Using the pre-trained word embedding gives our model the ability to introduce unseen entities that are in the vocabulary, but not necessarily in the training data.
\section{Evaluation}
\subsection{Dataset}
The dataset consists of 46,018 real-world triplets in Chinese, and it covers six different types of medical relationships, where 70\% data are used for training and 30\% validation data are used for hyperparameter tuning. Since the proposed model discovers entity pairs by directly sampling from the latent space, not by verifying pre-determined test cases, we evaluate the generated entity pairs directly. \ref{tab::data_stats_grvae} shows the statistics and representative samples for each medical relationship. We use 200-dimensional word embeddings learned from a Chinese medical corpus on the healthcare forum as the initial entity representation. The vocabulary covers 126,270 words.

\begin{table*}[hbt]
\centering
\resizebox{6in}{!}{%
\begin{tabular}{ccl}
\toprule
\textbf{RELATIONSHIP} & \textbf{COUNT} & \textbf{ENTITY PAIRS} \\
\midrule
\multirow{3}{*}{Disease $\xrightarrow{Cause}$ Body Part} & \multirow{3}{*}{2320} &  $<$\texttt{tricuspid insufficiency} (\textzh{三尖瓣闭锁}),  \texttt{tricuspid valve} (\textzh{三尖瓣})$>$\\
                 &                   & $<$\texttt{vaginal cancer} (\textzh{阴道癌}), \texttt{reproductive system} (\textzh{生殖})$>$\\
                 &                   & $<$\texttt{hydrocephaly} (\textzh{脑积水}), \texttt{head} (\textzh{头部})$>$\\\midrule

\multirow{3}{*}{Disease $\xrightarrow{Related To}$ Disease} & \multirow{3}{*}{4614} &  $<$\texttt{infant hydrocephalus} (\textzh{婴儿脑积水}), \texttt{congenital hydrocephalus} (\textzh{先天性脑积水})$>$\\
                 &                   & $<$\texttt{urethritis} (\textzh{尿道炎}), \texttt{cystitis} (\textzh{膀胱炎})$>$  \\  & & $<$\texttt{retention of food in the stomach} (\textzh{食滞胃脘}), \texttt{infantile indigestion} (\textzh{小儿消化不良})$>$\\\midrule
                 
\multirow{3}{*}{Disease $\xrightarrow{Need}$ Examine} & \multirow{3}{*}{4185} &  $<$\texttt{salicylates poisoning} (\textzh{水杨酸类中毒}),  \texttt{routine urianlysis} (\textzh{尿常规})$>$\\
                 &                   & $<$\texttt{tetralogy triad} (\textzh{法洛三联症}), \texttt{electrocardiogram, ECG} (\textzh{心电图})$>$  \\  & & $<$\texttt{epididymitis} (\textzh{附睾炎})  , \texttt{cremasteric reflex} (\textzh{提睾反射})$>$\\\midrule
                                  
\multirow{3}{*}{Symptom $\xrightarrow{Belong To}$ Department} & \multirow{3}{*}{8595} &  $<$\texttt{anchylosis, stiffness of a joint} (\textzh{关节强直 }), \texttt{orthopedics} (\textzh{骨科})$>$\\
                 &                   & $<$\texttt{female lower abdominal pain} (\textzh{女性小腹疼痛}), \texttt{gynecology} (\textzh{妇科})$>$  \\  & & $<$\texttt{absent infant sucking reflex} (\textzh{吸吮反射消失}), \texttt{neonatology} (\textzh{新生儿科})$>$\\\midrule
                 
\multirow{3}{*}{Disease $\xrightarrow{Cause}$ Symptom} & \multirow{3}{*}{16642} &  $<$\texttt{peritonitis} (\textzh{腹膜炎}), \texttt{abdominal venous engorgement} (\textzh{腹部静脉怒张})$>$\\
                 &                   & $<$\texttt{urethritis} (\textzh{尿道炎}), \texttt{urethra itching} (\textzh{尿道痒感})$>$  \\  & &$<$\texttt{radial nerve palsy} (\textzh{桡神经麻痹}), \texttt{upper extremity weakness} (\textzh{上肢无力})$>$\\\midrule
                 
\multirow{3}{*}{Symptom $\xrightarrow{Related To}$ Symptom} & \multirow{3}{*}{9662} &  $<$\texttt{redness and swelling around the umbilicus} (\textzh{脐周红肿}), \texttt{periumbilical swelling} (\textzh{脐周肿胀})$>$\\
 & & $<$\texttt{muscular contusion} (\textzh{肌肉挫伤}), \texttt{disinsertion} (\textzh{肌腱断裂})$>$\\            
 & & $<$\texttt{fingers benumbed with cold} (\textzh{手指冻肿}), \texttt{skin frostbite} (\textzh{皮肤冻伤})$>$\\            
\bottomrule                      
\end{tabular}%
}
\caption{Sample medical relationships and entity pairs.}\label{tab::data_stats_grvae}
\end{table*}
\subsection{Experiment Settings}
\noindent\textbf{Evaluation Metric}
Three evaluation metrics are introduced to quantitatively measure the generated relational medical entity pairs: quality, support, and novelty.

\textbf{Quality}
Since it is hard for the machine to evaluate whether a entity pair is meaningful or not, human annotation is involved in assessing the quality of the generated entity pairs. We deploy a human annotation task on Amazon Mechanical Turk. Annotators need to pass at least four in five sample cases to qualify the annotation. Majority voting of three annotators is adopted. The quality is measured by:
\begin{equation}
quality = \frac{\text{\# of entity pairs that are meaningful}}{\text{\#  of all the generated entity pairs}}.
\end{equation}

\textbf{Support}
Besides human annotations, a support score quantitatively measures the belongingness of an entity pair generated by a specific relationship to existing entity pairs with that relationship.
For each generated entity pair $<$${\hat e_h},{\hat e_t}$$>$, the support score measures its similarities to known entity pairs of each relationship $r_c$:
\begin{equation}
support_{<{\hat e_h},{\hat e_t}, r_c>} = \frac{1}{1 + distance({\hat embed}_h,{\hat embed}_t, r_c)},
\end{equation}
where $distance({\hat embed}_h,{\hat embed}_t, r_c)$ calculates the distance between the vector ${\hat embed}_h-{\hat embed}_t$ and $NN_{r_c}\left({\hat embed}_h - {\hat embed}_t\right)$ using distance measure such as cosine distance. The $NN_{r_c}$ implements the nearest neighbor search over the $embed_h-embed_t$ space among all the entity pairs having the relationship $r_c$.
For each generated entity pair, the support scores of all relationships are normalized:
\begin{equation}
norm\_support_{<{\hat e_h},{\hat e_t}, r_c>} = \frac{support_{<{\hat e_h},{\hat e_t}, r_c>}}{\sum\limits_{r_i}^{|R|}{support_{<{\hat e_h},{\hat e_t}, r_i>}}}.
\end{equation}
The generated entity pair $<$${\hat e}_h, {\hat e}_t$$>$ finds support from its estimated relationship which has the highest score, while the relationship $r$ given during the generating process is considered as the ground truth for $<$${\hat e}_h, {\hat e}_t$$>$. The final support value is based on the accuracy of the estimated relationship and the ground truth relationship.

\textbf{Novelty}
The ability to generate novel entity pairs is one of our key contributions. Due to different scope of knowledge among individuals, human annotators are not able to precisely evaluate the novelty. 
We measure the novelty of the generation process by:
\begin{equation}
novelty= \frac{\text{\# of entity pairs that do not exist in the dataset}}{\text{\# of all the generated entity pairs}}.
\end{equation}

\noindent\textbf{Baselines}
Considering that no known methods are currently available for the {\KDDTaskName} problem, and we consider it unfair to compare with discriminative methods which have external resources or further data requirements, the performance on the following models are compared:
\begin{itemize}
\item \textbf{\KDDModelName-MONO}: The proposed model that works with all entity pairs having the same relationship in both training and generation. For each relationship, we train a separate {\KDDModelName} with entity pairs having that relationship.
\item \textbf{RVAE}: The unconditional version of the model {\KDDModelName} where the relationship indicator $r$ is not provided during model training and generation.
\item \textbf{\KDDModelName-RAND}: The proposed model {\KDDModelName} with a random sampling based generator. Rather than using the density-based sampling strategy, the generator of \KDDModelName-RAND samples randomly from the latent space.
\item \textbf{\KDDModelName}: The proposed method where entity pairs with all types of relationships are used together to train the model. The training is conditioned on relationships, and density-based sampling is used.
\item \textbf{\KDDModelName-WA}: The proposed method with the warm-up strategy introduced in Section \ref{sec::loss_function}.
\end{itemize}
\begin{table}[h]
\centering
\resizebox{6in}{!}{%
\begin{tabular}{lllll}
\toprule
\textbf{MODEL}      & \textbf{QUALITY} & \textbf{SUPPORT} & \textbf{NOVELTY} & \textbf{LOSS (TRAIN / VALID)}\\ \midrule
\KDDModelName-MONO & 0.6698  & 0.9550  & 0.5118  & 47.3002 / 116.6739\\
\KDDModelName-RAND & 0.2550  & 0.3764  & 0.9952  & 43.0954 / 83.6589 \\
\KDDModelName      & 0.7308  & 0.9048  & 0.5682  & 43.0954 / 83.6589 \\
\KDDModelName-WA   & 0.7717  & 0.9291  & 0.6193  & 33.4399 / 57.9470 \\ \bottomrule
\end{tabular}
}
\caption{Performance comparison results.}
\label{tab::general_comparison}
\end{table}

\subsection{Experiment Results}
We generate 1000 entity pairs for each medical relationship for evaluation. \ref{tab::general_comparison} summarizes the performance of the proposed method when comparing with other alternatives.
In summary, \KDDModelName-MONO demonstrates the power of generative model that learns commonalities purely from the diversely expressed entity pairs without substantial data requirements. By comparing \KDDModelName-RAND and {\KDDModelName} we show the effectiveness of the density-based sampling in generating high-quality entity pairs. The warm up technique adopted in \KDDModelName-WA is able to give {\KDDModelName} a further performance boost. 
As a qualitative measure, we also provide entity pairs generated by the proposed model in \ref{tab::exp_results_cases}, from which we can see the meaningful and novel structured knowledge discovered in a generative fashion. 
\begin{table}[htb!]
\centering
\resizebox{5in}{!}{%
\begin{tabular}{l}
\toprule
Disease $\xrightarrow{Cause}$ Body Part  \\
$<$\texttt{dysentery} (\textzh{痢疾}), \texttt{intestine} (\textzh{肠})$>$ \\
$<$\texttt{brain tumor} (\textzh{脑瘤}), \texttt{head} (\textzh{头部})$>$\\ 
$<$\texttt{leukopenia} (\textzh{白细胞减少症}), \texttt{vascular system} (\textzh{血液})$>$\\ \midrule
Disease $\xrightarrow{Related To}$ Disease \\
$<$\texttt{foreign body in esophagus} (\textzh{食管异物}), \texttt{bowel obstruction} (\textzh{肠梗阻})$>$ \\
$<$\texttt{brain contusion} (\textzh{脑挫裂伤}), \texttt{amnesia} (\textzh{记忆障碍})$>$\\ 
$<$\texttt{respiratory acidosis} (\textzh{呼吸性酸中毒}), \texttt{pulmonary edema} (\textzh{肺水肿})$>$\\ \midrule
Disease $\xrightarrow{Need}$ Examine  \\
$<$\texttt{uremia} (\textzh{尿毒症}), \texttt{routine urianlysis} (\textzh{尿常规})$>$\\
$<$\texttt{bacterial meningitis} (\textzh{细菌性脑膜炎}), \texttt{cranial CT} (\textzh{头颅CT})$>$\\ 
$<$\texttt{bowel obstruction} (\textzh{肠梗阻}), \texttt{abdominal x-ray} (\textzh{腹部平片})$>$\\ \midrule
Symptom $\xrightarrow{Belong To}$ Department \\
$<$\texttt{retained placenta} (\textzh{胎盘滞留}), \texttt{obstetrics} (\textzh{产科})$>$\\
$<$\texttt{fluid retention} (\textzh{水潴留}), \texttt{nephrology} (\textzh{肾内科})$>$\\ 
$<$\texttt{stuffy nose} (\textzh{鼻塞}), \texttt{otolaryngology} (\textzh{耳鼻咽喉科})$>$\\ \midrule
Disease $\xrightarrow{Cause}$ Symptom  \\
$<$\texttt{otogenic brain abscess} (\textzh{耳源性脑脓肿}), \texttt{earache} (\textzh{耳痛})$>$\\
$<$\texttt{neuritis} (\textzh{神经炎}), \texttt{numbness in the hands} (\textzh{手麻})$>$\\
$<$\texttt{open head injury} (\textzh{开放性颅脑损伤}), \texttt{loss of consciousness} (\textzh{意识模糊})$>$\\ \midrule
Symptom $\xrightarrow{Related To}$ Symptom  \\ 
$<$\texttt{fatigue} (\textzh{乏力}), \texttt{feel wobbly and rough} (\textzh{四肢无力})$>$\\
$<$\texttt{joint pain} (\textzh{关节痛}), \texttt{limited joint mobility} (\textzh{关节活动受限})$>$\\
$<$\texttt{blurred vision} (\textzh{雾视}), \texttt{eye discomfort} (\textzh{眼睛不舒服})$>$\\ \bottomrule
\end{tabular}
}
\caption{Novel and meaningful entity pairs generated by the proposed method.}\label{tab::exp_results_cases}
\end{table}

\noindent\textbf{Generative Modeling Capacity}
Unlike discriminative models which utilize the discrepancies among instances of different classes to discriminate one class from another, the generative nature of the proposed method makes it generate entity pairs only when it fully understands the diverse expressions within each relationship. To validate such appealing property, we introduce the baseline \KDDModelName-MONO which works with all entity pairs having the same relationship in both training and generation. 

\ref{tab::exp_mono} compares the fine-grained quality, support and novelty of the generated entity pairs of \KDDModelName-MONO and {\KDDModelName} on each relationship.
The \KDDModelName-MONO achieves a reasonable performance on each relationship, which shows that the generative modeling has the ability to learn directly from the existing entity pairs without additional data requirement. Furthermore, when all types of entity pairs are trained altogether in \KDDModelName, we observe a consistent improvement in not only quality but also novelty.

\begin{table}[bht!]
\centering
\resizebox{6in}{!}{%
\begin{tabular}{lllll}
\toprule
\textbf{\KDDModelName-MONO}       & \textbf{QUALITY} & \textbf{SUPPORT} & \textbf{NOVELTY} & \textbf{LOSS (TRAIN/VALID)}\\ \midrule
Disease $\xrightarrow{Cause}$ Body Part    &  0.6830   & 1.0000        & 0.4880      &  54.9830 / 126.7426\\
Disease $\xrightarrow{Related To}$ Disease     &  0.6890          & 0.8700        & 0.4830      & 51.5131 / 155.0721 \\
Disease $\xrightarrow{Need}$ Examine     &  0.7080       & 1.0000        & 0.5210   &   54.7635 / 136.4802   \\
Symptom $\xrightarrow{Belong To}$ Department     &   0.6870     & 1.0000        & 0.4660  &  39.0959 / 72.5872    \\
Disease $\xrightarrow{Cause}$ Symptom     &   0.5870  & 0.9400        & 0.5730  &      37.3276 / 83.8797 \\ 
Symptom $\xrightarrow{Related To}$ Symptom   &  0.6650       & 0.9200        & 0.5400  &     46.1180 / 125.2818 \\  
\midrule
\textbf{\KDDModelName}  &            &         &    \\ \midrule
Disease $\xrightarrow{Cause}$ Body Part    &   0.7560    & 0.9990        & 0.7240      & \multirow{6}{*}{43.0954 / 83.6589} \\
Disease $\xrightarrow{Related To}$ Disease     &  0.6910   & 0.7440        & 0.8670        \\
Disease $\xrightarrow{Need}$ Examine     &   0.7570   & 0.9810        & 0.8710        \\
Symptom $\xrightarrow{Belong To}$ Department     &   0.7680   & 0.9950        & 0.6130        \\
Disease $\xrightarrow{Cause}$ Symptom     &  0.7020   & 0.8820        & 0.9270        \\ 
Symptom $\xrightarrow{Related To}$ Symptom   &  0.7110  & 0.8280        & 0.8880        \\  
\bottomrule
\end{tabular}
}
\caption{Performance comparison between \KDDModelName-MONO and \KDDModelName.}\label{tab::exp_mono}
\end{table}

\noindent\textbf{Effectiveness of Density-based Sampling}
To validate the effectiveness of the density-based sampling for the generator, we compare the proposed method with \KDDModelName-RAND where a random sampling strategy is adopted.
From \ref{tab::general_comparison} we can see that when the distribution of the latent space is not considered, the random sampling strategy in \KDDModelName-RAND tends to generate more entity pairs that are not seen in the existing dataset. However, the generated entity pairs are of low quality and support.

{\KDDModelName} adopts a density-based sampling.
The dense region in the latent space indicates that more entity pairs are located. Therefore, in \KDDModelName, the quality and support of the generated entity pairs benefit from sampling more often at denser regions in the latent space, resulting in less novel but higher quality entity pairs. 

\noindent\textbf{Ability to Infer Conditionally}
To effectively discover structured medical knowledge, one of our key contributions is to generate relational medical entity pairs for a specific relationship. That is, the ability to infer new entity pairs for a particular relationship without additional data preparation. Besides seamlessly incorporating this idea in the model design, we also show such conditional inference ability by visualization. 

\begin{figure}[hbt]
\centering
\epsfig{file=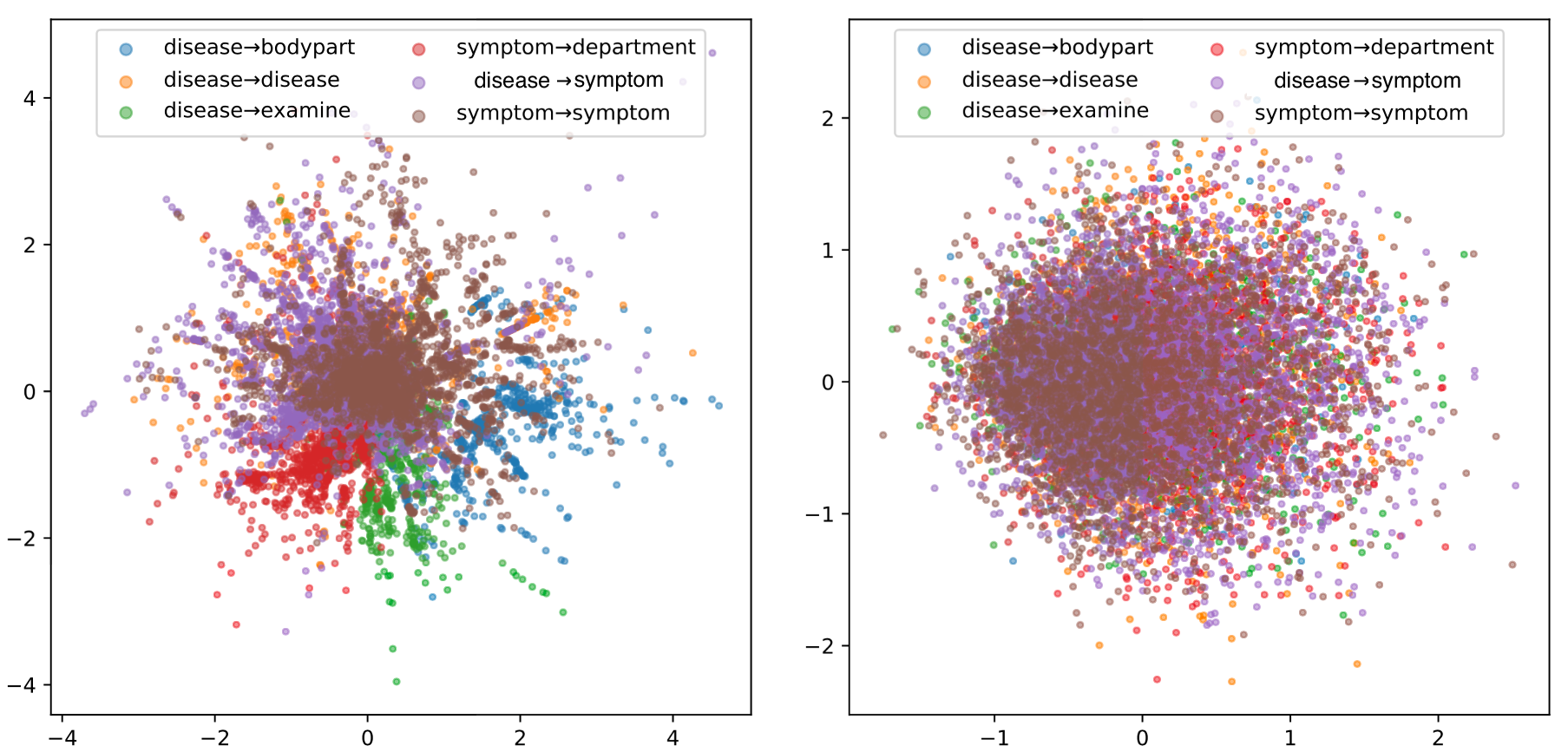, width=5in}
\caption{Visualizing the latent variable $\mu$ of RVAE (left) and {\KDDModelName} (right).}\label{fig::vis_latent}
\end{figure}

\ref{fig::vis_latent} shows the $\mu$ of validation samples after being mapped into a two-dimensional space using Primary Component Analysis for dimension reduction. 
The samples are colored based on their ground truth relationship indicators. The left figure indicates that when the relationship indicator $r$ is not given during the training/validation, RVAE is still able to map different relationships into various regions in the latent space, while a single distribution models all types of relationships. Such property is appealing for an unsupervised model, but since the relationship indicator $r$ is not given during training, RVAE fails to generate entity pairs having a particular relationship, unless we manually assign a boundary for each relationship in the latent space. The right figure shows that when the relationship indicator $r$ is incorporated during the training, \KDDModelName~ learns to let each relationship have a unified latent representation $P_\theta({\hat z}| r)$. A separate but nearly identical distribution is used to model each relationship. Such property may enable the generator of our model to sample the expression variations from a relationship-independent latent space, while the relationship indicator $r$ provides the categorical information regarding what type of relationship should the expression variation applies on.

\noindent\textbf{Relationship-enhancing Entity Adjustment}
\begin{table}[tb]
\centering
\resizebox{\linewidth}{!}{%
\begin{tabular}{ll}
\toprule
$\bullet$ \texttt{genital tract malformation} (\textzh{生殖道畸形}) &\\
NN in the relationship-enhanced space $\mathbb{R}^{1\times D_R}$ & {NN in the initial embedding space $\mathbb{R}^{1\times D_E}$}\\  \cmidrule(lr){1-1} \cmidrule(lr){2-2}
\texttt{genital tract} (\textzh{生殖道})&  \texttt{reproductive system} (\textzh{生殖系统})\\
\texttt{reproductive system} (\textzh{生殖系统})&    \texttt{reproductive tract tumors} (\textzh{生殖道肿瘤})    \\
\textbf{\texttt{heart malformations}} (\textzh{心脏畸形})&  \texttt{urinary system malformations} (\textzh{泌尿系畸形})\\
\textbf{\texttt{chromosome abnormalities}} (\textzh{染色体异常})&   \textbf{\texttt{infertility}} (\textzh{不孕})\\
\texttt{reproductive tract tumors} (\textzh{生殖道肿瘤})&       \texttt{vaginal atresia} (\textzh{阴道闭锁})\\
\texttt{generative organs} (\textzh{生殖器官})&        \texttt{genital tract} (\textzh{生殖道})\\
\textbf{\texttt{urinary system malformations}} (\textzh{泌尿系畸形})   &    \texttt{generative organs} (\textzh{生殖器官})\\
\textbf{\texttt{gastrointestinal malformations}} (\textzh{消化道畸形}) &    \textbf{\texttt{acyesis}} (\textzh{不孕症})\\
\toprule
$\bullet$ \texttt{muscle strain} (\textzh{肌肉拉伤}) & \\
NN in the relationship-enhanced space $\mathbb{R}^{1\times D_R}$& {NN in the initial embedding space $\mathbb{R}^{1\times D_E}$}\\  \cmidrule(lr){1-1} \cmidrule(lr){2-2}
\texttt{strain} (\textzh{拉伤}) & \textzh{拉伤} (\texttt{strain})\\
\textbf{\texttt{ligament strain}} (\textzh{韧带拉伤}) & \texttt{muscle tear} (\textzh{肌肉撕裂})\\
\texttt{sprain} (\textzh{扭伤})& \textbf{\texttt{pull-up}} (\textzh{引体向上})\\
\textbf{\texttt{foot pain}} (\textzh{足痛})& \texttt{sprain} (\textzh{扭伤})\\
\texttt{muscle tear} (\textzh{肌肉撕裂})& \texttt{muscle fatigue} (\textzh{肌肉疲劳})\\
\textbf{\texttt{plantar fasciitis}} (\textzh{足底筋膜炎})& \texttt{tenosynovitis} (\textzh{腱鞘炎})\\
\textbf{\texttt{joint sprain}} (\textzh{关节扭伤})& \texttt{tendonitis} (\textzh{肌腱炎})\\
\texttt{repetitive strain injury, RSI} (\textzh{劳损})& \textbf{\texttt{amount of exercise}} (\textzh{运动量})\\
\bottomrule
\end{tabular}
}
\caption{The effectiveness of relationship-enhancing adjustment.}\label{tab::case_embedding}
\end{table}
To show the effectiveness of relationship-enhancement, \ref{tab::case_embedding} shows the nearest neighbors of a disease entity \texttt{genital tract}\\\texttt{ malformation} (\textzh{生殖道畸形}) and a symptom entity \texttt{muscle strain} (\textzh{肌肉拉伤}) in their original embedding space, as well as in the space after relationship-enhancing.

From these cases we can see that the original entity representations trained with skip-gram \cite{mikolov2013efficient} tend to put entities in proximity when they appear in similar contexts. In the first case, the entity \texttt{genital tract malformation} (\textzh{生殖道畸形}) is in close proximity to \texttt{infertility} (\textzh{不孕}) and \texttt{acyesis} (\textzh{不孕症}). In the second case, entities that have similar context like \texttt{pull-up} (\textzh{引体向上}) and \texttt{amount of exercise} (\textzh{运动量}) are found near by the entity \texttt{muscle strain} (\textzh{肌肉拉伤}).

The translation layer adjusts the original entity representation so that they are more suitable for \TaskNameThree. The nearest neighbors in the adjusted space are not necessarily entities that co-occur in the same context, but more relation-wise similar with the given entity. For example, \texttt{heart malformations} (\textzh{心脏畸形}) and \texttt{chromosome abnormalities} (\textzh{染色体异常}) may not be semantically similar with the given word \texttt{genital tract malformation} (\textzh{生殖道畸形}), but they may serve similar functionalities in a Disease $\xrightarrow{Cause}$ Symptom relationship.

\subsection{Hyperparameter Analysis}
We train the proposed model with a wide range of hyperparameter configurations, which are listed in \ref{tab::para_config}. 
We vary the batch size from 64 to 256. The dimension $D_R$ for translating the initial entity embeddings is set from 64 to 2048. We try two to seven hidden layers from $trans_{ht}$ to $l_{ht}$ and from $[z,r]$ to $trans'_{ht}$, with different non-linear activation functions. For each hidden layer, the hidden unit number $D_H$ is set from 2 to 1024. The latent dimension $D_L$ is set from 2 to 200.
\begin{table}[hbt]
\centering
\resizebox{6in}{!}{%
\begin{tabular}{ll}
\toprule
Parameter  & Value                                                     \\ \midrule
Batch Size & 64, 128, 256                                              \\
$D_R$      & 64, 128, 256, 512, 640, 768, 1024, 1280, 1536, 1792, 2048 \\
$D_H$      & 2, 4, 8, 16, 32, 64, 128, 256, 512, 640, 768, 1024        \\ 
$D_L$      & 2, 3, 4, 5, 10, 20, 50, 100, 200                          \\
Activation & ELU \cite{clevert2015fast}, ReLU \cite{nair2010rectified}, Sigmoid, Tanh \\
Optimizer  & Adadelta \cite{zeiler2012adadelta}, Adagrad \cite{duchi2011adaptive}, Adam \cite{kingma2014adam}, RMSProp \cite{tieleman2012lecture} \\ \bottomrule 
\end{tabular}
}
\caption{Hyperparameter configurations.}\label{tab::para_config}
\end{table}

The top-5 hyperparameter settings with low validation losses are shown in \ref{tab::parameters}.
Among the combinations of hyperparameter configurations, we find that for fully connected hidden layers from $trans_{ht}$ to $l_{ht}$, a sequence of six consecutive layers: 1792-640-640-512-256-64 works the best for the encoder with ELU as the activation function. For $[z,r]$ to $trans'_{ht}$ in the decoder, such layer setting is organized in a reverse order. A batch size of 64 and the Adadelta optimizer work the best for our task. $D_R=640$ is used. The latent dimension $D_L=200$ is adopted for $\mu$ and $\sigma^2$. 
We use Xavier initialization \cite{glorot2010understanding} for weight variables and zeros for biases. Such configuration achieves a training loss of 43.0954 and a validation loss of 83.6589.

\begin{table}[h]
\sisetup{round-mode=places,round-precision=4}
\centering
\resizebox{4.5in}{!}{%
\begin{tabular}{cclcccc}
\toprule
\small{Batch} & $D_R$ & $\{D_H\}$ & $D_L$ & \small{Act.} & \small{Optimizer} & Loss(Training /Valid) \\ 
\cmidrule(lr){1-6} \cmidrule(lr){7-7}
64         &   640           &  1792-640-640-512-256-64        &         200   &  ELU      &       Adadelta  &    43.0954 / 83.6589 \\
64         &   640            &  1792-256-640-512-256-128  &   200           &   ELU        &  Adadelta      &    51.0695 / 86.9153 \\
64         &  640            &   1792-256-640-512-256-64  &  200            &   ELU        & Adadelta      &   50.4392 / 88.6438 \\
128 &640 &1792-640-768-512-64-128 &50 &ELU &Adadelta &50.5997 / 89.0125\\
256 &640 & 512-768-640-256-512 &50 &ELU &Adam &62.1955 / 89.2014\\
\bottomrule
\end{tabular}
}
\caption{Hyperparameter analysis.}
\label{tab::parameters}
\end{table}
\section{Related Works}
\noindent\textbf{Deep Generative Models}: 
Recent years have witnessed an increasing interest in deep generative models that generate observable data based on hidden parameters.
Various deep generative models have been developed, such as Generative Adversarial Networks (GANs) \cite{radford2015unsupervised} and Variational Autoencoders (VAEs) \cite{kingma2013auto}. 
Unlike Generative Adversarial Networks (GANs) \cite{radford2015unsupervised} which generate data based on arbitrary noises, the Variational Autoencoders (VAEs) \cite{kingma2013auto} setting we adopted is more expressive since it tries to model the underlying probability distribution of the data by latent variables so that we can sample from that distribution to generate new data accordingly. 
An increasing number of models and applications are proposed which consider data in different modalities, such as generating images \cite{pu2016variational,gregor2015draw} or natural language \cite{bowman2016generating,marcheggiani2016discrete,xu2017variational}. \cite{yao2011structured} works on generative relation discovery with a probabilistic graphic model that requires hand-crafted relation-level features.
As far as we know, the {\TaskNameThree} problem we studied in this work, which is suitable for deep generative modeling, has not been studied in a generative perspective with restricted data requirement.

\noindent\textbf{Knowledge Graph Completion}:
Existing knowledge graph completion methods \cite{bordes2011learning,wang2014knowledge,sun2012will,gardner2015efficient,wang2015knowledge,lin2016neural} are discriminative models. During training, those methods are trained to distinguish entity pairs of one relationship from another \cite{zeng2014relation,lin2016neural}, or to identify meaningful entity pairs from randomly sampled negative entity pairs with no relationships \cite{bordes2013translating,socher2013reasoning}. During testing, some candidate entity pairs are prepared ahead of time and given to the model. The model examines what kind of, and how likely there is a relationship for each candidate entity pair. Other works such as \cite{zhang2019missing} aligns entities from multiple existing knowledge graphs for synergistic completion.
The proposed model can be seen as augmenting an existing knowledge graph in a generative way. Although both knowledge graph completion task and our task provide additional entity pairs as their results, they share different objectives, and adopt entirely different approaches.
The knowledge base completion models rely on the discrepancies among entity pairs of different relationships to distinguish one from another. Otherwise, random negative samples are used for discriminative training. Our model does not rely on discrepancies among relationships: it exploits the commonalities from diverse expressions within each relationship for a rational generation.
Knowledge graph completion methods are also vulnerable to low-quality candidate entity pairs during testing: the truly meaningful entity pairs cannot be even obtained when they are not a part of the candidate entity pairs for discriminative models to examine. The choice of candidates involves additional human annotation to improve efficiency; otherwise, any dyadic combinations of medical entities need to be fed to and tested by the model. While the generative nature of our model makes it only generate rational entity pairs by learning from the existing rational ones: no additional data needs to be prepared for generative discovery.

\noindent\textbf{Relationship Extraction}:
There is another related research area that studies relation extraction \cite{baeza2007extracting,agichtein2000snowball,sahay2008discovering,mintz2009distant,wang2015constrained,jiang2017metapad,liu2017heterogeneous}, which usually amounts to examining whether or not a relation exists between two given entities in a context \cite{culotta2006integrating}.
Most relationship extraction methods require large amounts of high-quality external information, such as a large text corpus \cite{baeza2007extracting,agichtein2000snowball,sahay2008discovering,li2016extracting} and knowledge graphs \cite{chang2014typed,syed2010automatic,verga2016generalizing}.
However, in specific domains such as the medical domain, it is tedious and label-intensive to obtain a sufficient amount of free-text corpora which contains the co-occurrence of all kinds of entity pairs. Thus, we propose an effective generative method that learns from the existing entity pairs directly. Pre-trained word vectors are used in our model to provide initial entity representations, which do not introduce further labeling cost.

\chapter{\TaskNameFour}\label{chapter:c5}
Part of this chapter was published as ``SynonymNet: Multi-context Bilateral Matching for Entity Synonyms'', on ArXiv~\cite{zhang2018synonymnet}: \url{https://arxiv.org/abs/1901.00056}.
\newcommand{\ACLModelName}{\textsc{SynonymNet}}
\section{Introduction}
Discovering synonymous entities from a massive corpus is an indispensable task for automated knowledge discovery. For each entity, its synonyms refer to the entities that can be used interchangeably under certain contexts. For example, \texttt{Clogged Nose} and \texttt{Nasal Congestion} are synonyms relative to the context in which they are mentioned.
Given two entities, the synonym discovery task determines how likely these two entities are synonym with each other. 
The main goal of synonym discovery is to learn a metric that distinguishes synonym entities from non-synonym ones.

The synonym discovery task is challenging to deal with, a part of which due to the various entity
expressions. For example, \texttt{U.S.A}/ \texttt{United States of America}/ \texttt{United States}/ \texttt{U.S.} refer to the same entity but are expressed quite differently.
Recent works on synonym discovery focus on learning the similarity from entities and their character-level features \cite{neculoiu2016learning,mueller2016siamese}. These methods work well for synonyms that share a lot of character-level features like \texttt{airplane}/ \texttt{aeroplane} or an entity and its abbreviation like \texttt{Acquired Immune Deficiency Syndrome}/ \texttt{AIDS}. However, a much larger number of synonym entities in the real world do not share a lot of character-level features, such as \texttt{JD}/ \texttt{law degree}, or \texttt{clogged nose}/ \texttt{nasal congestion}. With only character-level features being used, these models hardly obtain the ability to discriminate entities that share similar semantics but are not alike verbatim.

Context information is helpful in indicating entity synonymity, as the meaning of an entity can be better reflected by the contexts in which it appears. 
Modeling the context for entity synonym usually suffers from following challenges:
1) \noindent\textbf{Semantic Structure}. Context, as a snippet of natural language sentence, is essentially semantically structured. Some existing models encode the semantic structures in the contexts implicitly during the entity representation learning \cite{mikolov2013distributed,pennington2014glove,peters2018deep}. The context-aware entity representations embody meaningful semantics: entities with similar contexts are likely to live in proximity in the embedding space. 
Some other works extract and model contexts in an explicit manner with structured annotations. Structured annotations such as dependency parsing \cite{qu2017automatic}, user click information \cite{wei2009context}, or signed heterogeneous graphs \cite{ren2015synonym} are introduced to guide synonym discovery.
2) \noindent\textbf{Diverse Contexts}.
An entity can be mentioned under a wide range of circumstances. Previous works on context-based synonym discovery either focus on entity information only \cite{neculoiu2016learning,mueller2016siamese}, or use a single piece of context for each entity \cite{liao2017deep,qu2017automatic} to learn a similarity function for entity matching. While in practice, similar context is only a sufficient but not necessary condition for context matching. Notably, in some domains such as medical, the context expression preference varies a lot from individuals. For example, \texttt{sinus congestion} is usually referred by medical professionals in the medical literature, while patients  often use \texttt{stuffy nose} on social media. It is not practical to assume that each piece of context is equally informative to represent the meaning of an entity:
a context may contribute differently when matched with different contexts of other entities.
Thus it is imperative to focus on multiple pieces of contexts with a dynamic matching schema for accuracy and robustness.

In light of these challenges, we propose a framework to discover synonym entities from a massive corpus without additional structured annotation. Candidate entities are obtained from a massive text corpus unsupervisely.
A novel neural network model {\ACLModelName} is proposed to detect entity synonyms based on two given entities via a bilateral matching among multiple pieces of contexts in which each entity appears. A leaky unit is designed to explicitly alleviate the noises from uninformative context during the matching process. 

The contribution of this work is summarized as follows:
\vspace{-0.1in}
\begin{itemize}
\item We propose {\ACLModelName}, a context-aware bilateral matching model to detect entity synonyms. {\ACLModelName} utilizes multiple pieces of contexts in which each entity appears, and a bilateral matching schema with leaky units to determine entity synonymity.
\item We introduce a synonym discovery framework that adopts {\ACLModelName} to
obtain synonym entities from a free-text corpus without additional structured annotation.
\item Experiments on generic and domain-specific real-world datasets in English and Chinese demonstrate the effectiveness of the proposed model for synonym discovery.
\end{itemize}
\section{Proposed Approach}
\begin{figure*}[bth!]
\centering
\includegraphics[width=\linewidth]{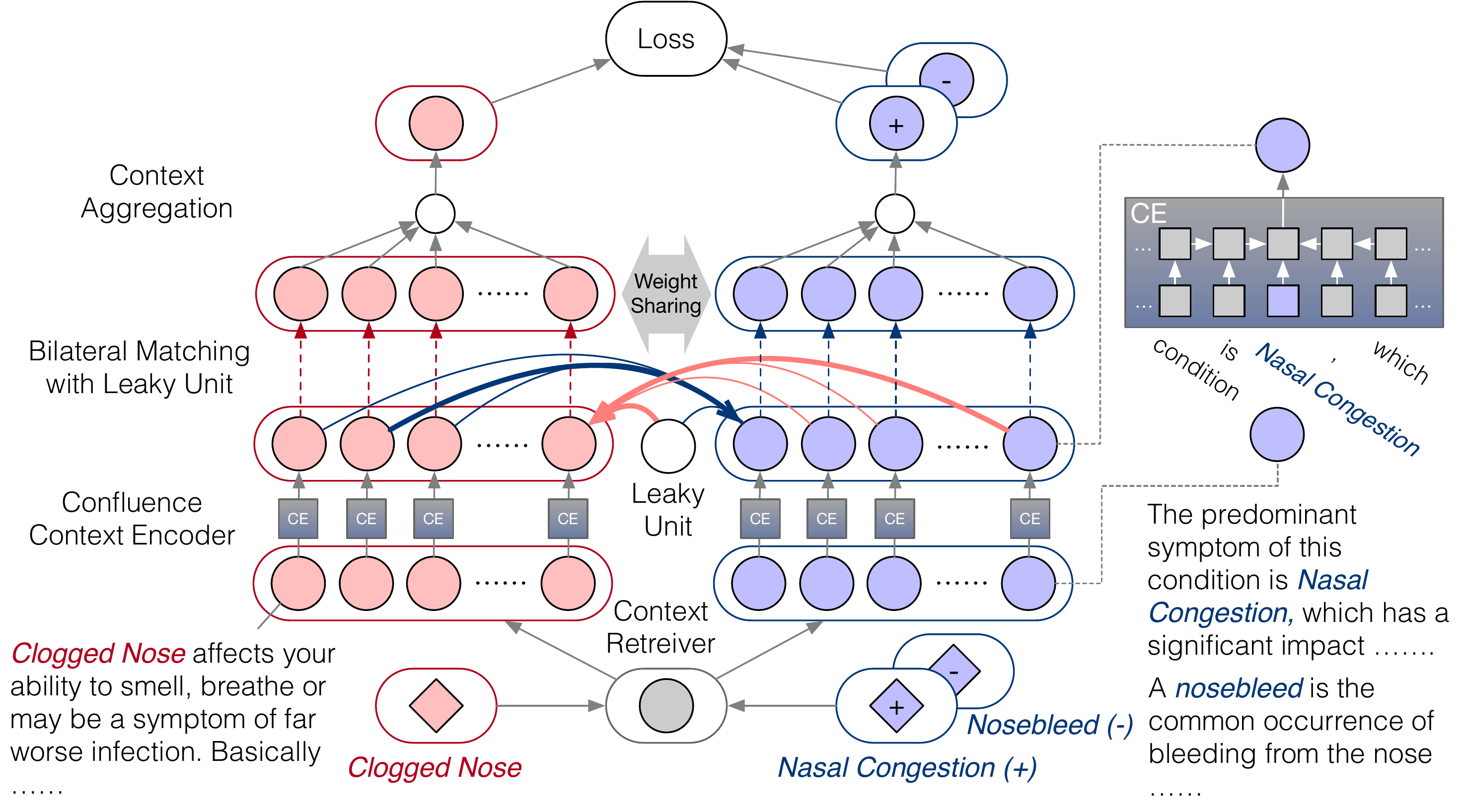}
\caption{An overview of the proposed model {\ACLModelName}.}\label{fig::model_overview}
\end{figure*}

We introduce {\ACLModelName}, our proposed model that detects whether or not two entities are synonyms to each other based on a bilateral matching between multiple pieces of contexts in which entities appear.
\ref{fig::model_overview} gives an overview of the proposed model. The diamonds are entities. Each circle is associated with a piece of context in which an entity appears. {\ACLModelName} learns to minimize the loss calculated using multiple pieces of contexts via bilateral matching with leaky units.

\subsection{Context Retriever}
For each entity $e$, the context retriever randomly fetches $P$ pieces of contexts from the corpus $D$ in which the entity appears. We denote the retrieved contexts for $e$ as a set 
$C=\{c_1,c_2,...,c_P\}$, where $P$ is the number of context pieces.
Each piece of context $c_p \in C$ contains a sequence of words
$c_p = (w^{(1)}_{p}, w^{(2)}_{p}, ..., w^{(T)}_{p}),$
where $T$ is the length of the context, which varies from one instance to another. ${w}^{(t)}_{p}$ is the $t$-th word in the $p$-th context retrieved for an entity $e$.

\subsection{Confluence Context Encoder}\vspace{-.1in}
For the $p$-th context $c_p$, an encoder tries to learn a continuous vector that represents the context.
For example, a recurrent neural network (RNN) such as a bidirectional LSTM (Bi-LSTM) \cite{hochreiter1997long} can be applied to sequentially encode the context into hidden states:
\begin{equation}
{\mathbf{\overset{\lower0.5em\hbox{$\smash{\scriptscriptstyle\rightarrow}$}}{h^{(t)}_p}}} = \textrm{LSTM}_{fw} (\mathbf{w}^{(t)}_{p}, 
{\mathbf{\overset{\lower0.5em\hbox{$\smash{\scriptscriptstyle\rightarrow}$}}{h^{(t-1)}_p}}}),
\end{equation}
\begin{equation}
{\mathbf{\overset{\lower0.5em\hbox{$\smash{\scriptscriptstyle\leftarrow}$}}{h^{(t)}_p} }} = \textrm{LSTM}_{bw} (\mathbf{w}^{(t)}_{p}, 
{\mathbf{\overset{\lower0.5em\hbox{$\smash{\scriptscriptstyle\leftarrow}$}}{h^{(t+1)}_p} }}),
\end{equation}
where $\mathbf{w}^{(t)}_{p}$ is the word embedding vector used for the word $w^{(t)}_{p}$.
We could concatenate the last hidden state ${\mathbf{\overset{\lower0.5em\hbox{$\smash{\scriptscriptstyle\rightarrow}$}}{h^{(T)}_p} }}$ in the forward $\textrm{LSTM}_{fw}$ with the first hidden state ${\mathbf{\overset{\lower0.5em\hbox{$\smash{\scriptscriptstyle\leftarrow}$}}{h^{(1)}_p} }}$ from the backward $\textrm{LSTM}_{bw}$ to obtain the context vector $\mathbf{h}_p$ for $c_p$:
${{\mathbf{h}}_p} = [{\mathbf{\overset{\lower0.5em\hbox{$\smash{\scriptscriptstyle\rightarrow}$}}{h^{(T)}_p} }}, {\mathbf{\overset{\lower0.5em\hbox{$\smash{\scriptscriptstyle\leftarrow}$}}{h^{(1)}_p} }}].$
However, such approach does not explicitly consider the location where the entity is mentioned in the context. As the context becomes longer, it is getting risky to simply rely on the gate functions of LSTM to properly encode the context.

We introduce an encoder architecture that models contexts for synonym discovery, namely the confluence context encoder. The confluence context encoder learns to encode the local information around the entity from the raw context, without utilizing additional structured annotations.
It focuses on both forward and backward directions. However, the encoding process for each direction ceases immediately after it goes beyond the entity word in the context:
${{\mathbf{h}}_p} = [{\mathbf{\overset{\lower0.5em\hbox{$\smash{\scriptscriptstyle\rightarrow}$}}{\mathbf{h}^{(t_e)}_p} }}, {\mathbf{\overset{\lower0.5em\hbox{$\smash{\scriptscriptstyle\leftarrow}$}}{\mathbf{h}^{(t_e)}_p} }}],$
where $t_e$ is the index of the entity word $e$ in the context and $\mathbf{h}_p\in\mathbb{R}^{1\times d_{CE}}$. By doing this, the confluence context encoder summarizes the context while explicitly considers the entity's location in the context, where no additional computation cost is introduced.

Comparing with existing works for context modeling \cite{cambria2018senticnet} where the left context and right context are modeled separately, but with the entity word being discarded, the confluence context encoder preserves entity mention information as well as the inter-dependencies between the left and right contexts.

\subsection{Bilateral Matching with Leaky Unit}
Considering the base case, where we want to identify whether or not two entities, say $e$ and $k$, are synonyms with each other, we propose to find the consensus information from multiple pieces of contexts via a bilateral matching schema.
Recall that for entity $e$, $P$ pieces of contexts $H = \{\mathbf{h}_1, \mathbf{h}_2, ..., \mathbf{h}_{P}\}$ are randomly fetched and encoded. And for entity $k$, we denote $Q$ pieces of contexts being fetched and encoded as $G = \{\mathbf{g}_1, \mathbf{g}_2, ..., \mathbf{g}_Q\}$. 
Instead of focusing on a single piece of context to determine entity synonymity, we adopt a bilateral matching between multiple pieces of encoded contexts for both accuracy and robustness.

\noindent\textit{H$\rightarrow$G} matching phrase:
For each $\mathbf{h_{p}}$ in $H$ and $\mathbf{g_{q}}$ in $G$, the matching score $m_{p \to q}$ is calculated as: 
\begin{equation}
{m_{p \to q}} = \frac{{\exp ({{\mathbf{h}}_p}{{\mathbf{W}}_{{\text{BM}}}}{\mathbf{g}}_q^{\text{T}})}}{{\sum\limits_{p' \in P} {\exp ({{\mathbf{h}}_{p'}}{{\mathbf{W}}_{{\text{BM}}}}{\mathbf{g}}_q^{\text{T}})}}},
\end{equation}
where $\mathbf{W}_{\text{BM}} \in \mathbb{R}^{d_{CE} \times d_{CE}}$ is a bi-linear weight matrix. 

Similarly, the \textit{H$\leftarrow$G} matching phrase considers how much each context $\mathbf{g}_q \in G$ could be useful to $\mathbf{h}_p \in H$:
\begin{equation}
{m_{p \leftarrow q}} = \frac{{\exp ({{\mathbf{g}}_q}{{\mathbf{W}}_{{\text{BM}}}}{\mathbf{h}}_p^{\text{T}})}}{{\sum\limits_{q' \in Q} {\exp ({{\mathbf{g}}_{q'}}{{\mathbf{W}}_{{\text{BM}}}}{\mathbf{h}}_p^{\text{T}})} }}.
\end{equation}
Note that $P\times Q$ matching needs to be conducted in total for each entity pair. We write the equations for each $\mathbf{h}_p\in H$ and ${\mathbf{g}}_q\in G$ for clarity. Regarding the implementation, the bilateral matching can be easily written and effectively computed in a matrix form, where a matrix multiplication is used $\mathbf{H}{{\mathbf{W}}_{{\text{BM}}}}\mathbf{G}^T \in \mathbb{R}^{P{\times}Q}$ where $\mathbf{H}\in \mathbb{R}^{P{\times}D_{CE}}$ and $\mathbf{G}\in\mathbb{R}^{Q{\times}D_{CE}}$. The matching score matrix $\mathbf{M}$ can be obtained by taking softmax on the $\mathbf{H}{{\mathbf{W}}_{{\text{BM}}}}\mathbf{G}^T$ matrix over certain axis (over 0-axis for $\mathbf{M}_{p \to q}$, 1-axis for $\mathbf{M}_{p \leftarrow q}$).

Not all contexts are informative during the matching for two given entities. For example, some contexts may contain
intricate contextual information even if they mention the entity explicitly. In this work, we introduce a leaky unit during the bilateral matching, so that uninformative contexts can be routed via the leaky unit rather than forced to be matched with any informative contexts. 
The leaky unit is a domain-dependent vector $\mathbf{l} \in \mathbb{R}^{1\times d_{CE}}$ learned with the model. For simplicity, we keep $\mathbf{l}$ as a zero vector.
If we use the \textit{H$\rightarrow$G} matching phrase as an example, the matching score from the leaky unit $\mathbf{l}$ to the $q$-th encoded context in $\mathbf{g}_q$ is:
\begin{equation}
{m_{l \to q}} = \frac{{\exp ({\mathbf{l}}{{\mathbf{W}}_{{\text{BM}}}}{\mathbf{g}}_q^{\text{T}})}}{{\exp ({\mathbf{l}}{{\mathbf{W}}_{{\text{BM}}}}{\mathbf{g}}_q^{\text{T}}) + \sum\limits_{p' \in P} {\exp ({{\mathbf{h}}_{p'}}{{\mathbf{W}}_{{\text{BM}}}}{\mathbf{g}}_q^{\text{T}})} }}.
\end{equation}
Then, if there is any uninformative context in $H$, say the ${\tilde p}$-th encoded context, $\mathbf{h}_{\tilde p}$ will contribute less when matched with $\mathbf{g}_q$ due to the leaky effect: when $\mathbf{h}_{\tilde p}$ is less informative than the leaky unit $\mathbf{l}$.
\begin{equation}
{m_{{\tilde p} \to q}} = \frac{{\exp ({{\mathbf{h}}_{\tilde p}}{{\mathbf{W}}_{{\text{BM}}}}{\mathbf{g}}_q^{\text{T}})}}{{\exp ({\mathbf{l}}{{\mathbf{W}}_{{\text{BM}}}}{\mathbf{g}}_q^{\text{T}}) + \sum\limits_{p' \in P} {\exp ({{\mathbf{h}}_{p'}}{{\mathbf{W}}_{{\text{BM}}}}{\mathbf{g}}_q^{\text{T}})}}}.
\end{equation}

\subsection{Context Aggregation}
The informativeness of a context for an entity should not be a fixed value: it heavily depends on the other entity and the other entity's contexts that we are comparing with.
The bilateral matching scores indicate the matching among multiple pieces of encoded contexts for two entities. For each piece of encoded context, say $\mathbf{g}_q$ for the entity $k$, we use the highest matched score with its counterpart as the relative informativeness score of $\mathbf{g}_q$ to $k$, denote as ${a_q} = \max ({m_{p \to q}}|p \in P).$
Then, we aggregate multiple pieces of encoded contexts for each entity to a global context based on the relative informativeness scores:
\begin{equation}
\begin{gathered}
 \text{for entity~$e$:}~~~\mathbf{\bar h}  = \sum\nolimits_{p \in P} {{a_p}{{\mathbf{h}}_p}},\\
 \text{for entity~$k$:}~~~\mathbf{\bar g}  = \sum\nolimits_{q \in Q} {{a_q}{{\mathbf{g}}_q}}. 
\end{gathered}
\end{equation}
Note that due to the leaky effect, less informative contexts are not forced to be heavily involved during the aggregation: the leaky unit may be more competitive than contexts that are less informative, thus assigned with larger matching scores. However, as the leaky unit is not used for aggregation, scores on informative contexts become more salient during context aggregation.

\subsection{Training Objectives}
We introduce two architectures for training the {\ACLModelName}: a siamese architecture and a triplet architecture.
\\\noindent\textbf{Siamese Architecture}
The Siamese architecture takes two entities $e$ and $k$, along with their contexts $H$ and $G$ as the input. 
The following loss function $L_\text{Siamese}$ is used in training for the Siamese architecture:
\begin{equation}
L_\text{Siamese} = y L_{+}(e, k) + (1-y) L_{-}(e, k),
\end{equation}
where it contains losses for two cases: $L_{+}(e,k)$ when $e$ and $k$ are synonyms to each other ($y=1$), and $L_{-}(e,k)$ when $e$ and $k$ are not ($y=0$). Specifically, inspired by \cite{neculoiu2016learning}, we have
\begin{equation}
\begin{gathered}
  {L_{+}}(e,k) = \frac{1}{4}{(1 - s(\mathbf{\bar h} ,\mathbf{\bar g} ))^2}, \hfill \\
  {L_{-}}(e,k) = {max(s(\mathbf{\bar h} ,\mathbf{\bar g} ) - m,0)^2}, \hfill \\ 
\end{gathered}
\label{eq::siamese}
\end{equation}
where $s(\cdot)$ is a similarity function, e.g. cosine similarity, and $m$ is the margin value. $L_{+}(e,k)$ decreases monotonically as the similarity score becomes higher within the range of [-1,1]. $L_{+}(e,k)=0$ when $s(\mathbf{\bar h} ,\mathbf{\bar g})=1$. For $L_{-}(e,k)$, it remains zero when $s(\mathbf{\bar h} ,\mathbf{\bar g})$ is smaller than a margin $m$. Otherwise $L_{-}(e,k)$ increases as $s(\mathbf{\bar h} ,\mathbf{\bar g})$ becomes larger.

\noindent\textbf{Triplet Architecture}
The Siamese loss makes the model assign rational pairs with absolute high scores and irrational ones with low scores, while the rationality of entity synonymity could be quite relative to the context.
The triplet architecture learns a metric such that the global context $\mathbf{\bar h}$ of an entity $e$ is relatively closer to a global context $\mathbf{\bar g_{+}}$ of its synonym entity, say $k_{+}$, than it is to the global context $\mathbf{\bar g_{-}}$ of a negative example $\mathbf{\bar g_{-}}$ by some margin value $m$. The following loss function $L_{{\text{Triplet}}}$ is used in training for the Triplet architecture:
\begin{equation}
{L_{{\text{Triplet}}}} = \max (s({\mathbf{\bar h}},{{{\mathbf{\bar g}}}_ - }) - s({\mathbf{\bar h}},{{{\mathbf{\bar g}}}_ + }) + m,0).
\label{eq::triplet}
\end{equation}

\subsection{Inference}
\begin{figure}[th!]
\centering
\includegraphics[width=5in]{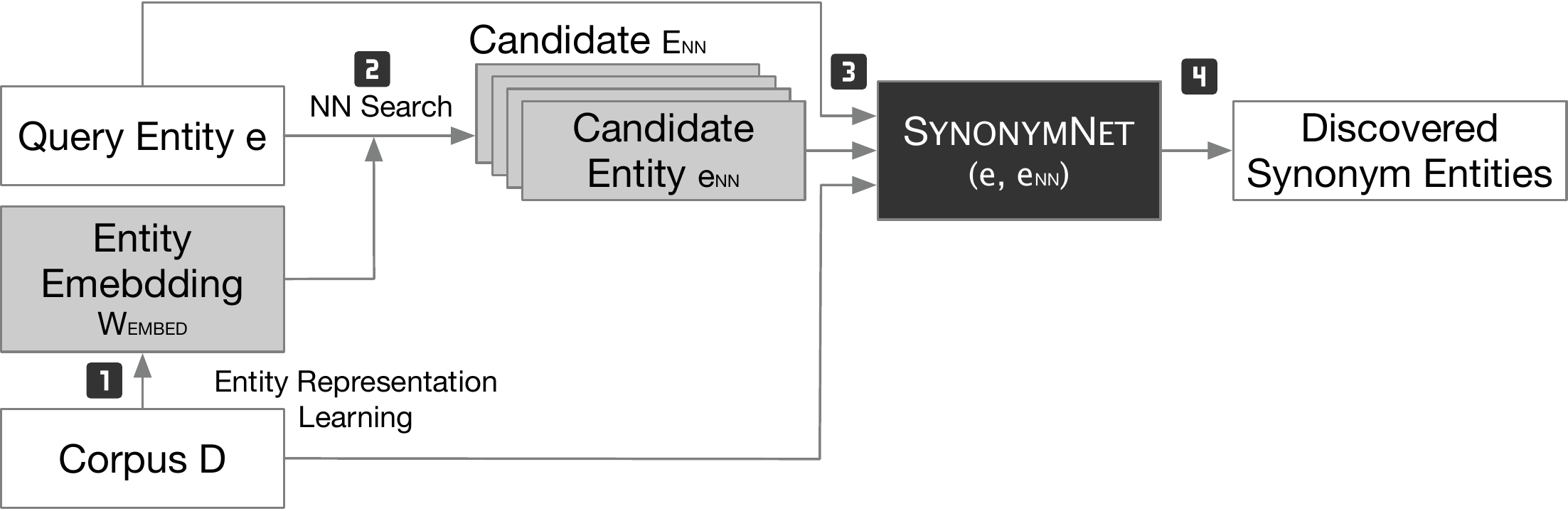}
\caption{Synonym discovery during the inference phase with {\ACLModelName}.}\label{fig::framework_overview}
\end{figure}

The objective of the inference phase is to discover synonym entities for a given query entity from the corpus effectively. 
We utilize context-aware word representations to obtain candidate entities that narrow down the search space. The {\ACLModelName} verifies entity synonymity by assigning a synonym score for two entities based on multiple pieces of contexts.
The overall framework is described in \ref{fig::framework_overview}, which contains four steps \noindent(1): Obtain entity representations $\mathbf{W}_{\text{EMBED}}$ from the corpus $D$. 
\noindent(2): For each query entity $e$, search in the entity embedding space and construct a candidate entity set $E_{NN}$.
\noindent(3): Retrieve contexts for the query entity $e$ and each candidate entity $e_{NN} \in E_{NN}$ from the corpus $D$, and feed the encoded contexts into {\ACLModelName}.
\noindent(4): Discover synonym entities of the given entity by the output of \ACLModelName.

When given a query entity $e$, it is tedious and very ineffective to verify its synonymity with all the other possible entities. 
In the first step, we train entity representation unsupervisely from the massive corpus $D$ using methods such as skip-gram \cite{mikolov2013distributed} or GloVe \cite{pennington2014glove}. An embedding matrix can be learned $\mathbf{W}_{\text{EMBED}} \in \mathbb{R} ^{v\times d_{\text{EMBED}}}$, where $v$ is the number of unique tokens in $D$.
Although these unsupervised methods utilize the context information to learn semantically meaningful representations for entities, they are not directly applicable to entity synonym discovery. 
However, they do serve as an effective way to obtain candidates as they tend to give entities with similar neighboring context words similar representations. 
For example, \texttt{nba championship}, \texttt{chicago black hawks} and \texttt{american league championship series} have similar representations because they tend to share some similar neighboring words. But they are not synonyms with each other.

In the second step, we construct a candidate entity list $E_{NN}$ by finding nearest neighbors of a query entity $e$ in the entity embedding space of $\mathbb{R}^{d_\text{EMBED}}$. Ranking entities by their proximities with the query entity on the entity embedding space significantly narrows down the search space for synonym discovery.

For each candidate entity $e_{NN} \in E_{NN}$ and the query entity $e$, we randomly fetch multiple pieces of contexts in which entities are mentioned, and feed them into the proposed {\ACLModelName} model.

{\ACLModelName} calculates a score $s(e, e_{NN})$ based on the bilateral matching with leaky units over multiple pieces of contexts. The candidate entity $e_{NN}$ is considered as a synonym to the query entity $e$ when it receives a higher score $s(e, e_{NN})$ than other non-synonym entities, or exceeds a specific threshold. 

Here we provide pseudo codes for the synonym discovery using {\ACLModelName}.
\begin{algorithm}
\begin{algorithmic}
\State{\textbf{Data:} Candidate entity $e$, Entity Word Embeddings $W_{\text{EMBED}} \in \mathbb{R}^{v\times d}$, Document $D$}
\State{\textbf{Result:} Entity Set $K$ where each $k \in K$ is a synonym entity of $e$}
\State{$E_{NN}$ = \text{NearestNeighbor}($e$, $W_{\text{EMBED}}$)}
\State{Order $E_{NN}$ by the distance to $e$;}
\For{$e_{NN}$ in $E_{NN}$}
    \State {Retrieve Contexts for $e_{NN}$ from Document $D$;}
    \State {Apply {\ACLModelName} on $e$ and $e_{NN}$;}
    \If{$s(e, e_{NN})>$threshold}
        \State{Add $e_{NN}$ as a synonym of $e$ to $K$;}
        \EndIf
    \EndFor
\caption{Effective Synonym Discovery via \ACLModelName.}\label{alg::framework}
\end{algorithmic}
\end{algorithm}
\section{Evaluation}
\subsection{Datasets}
Three datasets are prepared to show the effectiveness of the proposed model on synonym discovery. The Wiki dataset contains 6.8M documents from Wikipedia\footnote{https://www.wikipedia.org/} with generic synonym entities obtained from Freebase\footnote{https://developers.google.com/freebase}.
The PubMed is an English dataset where 0.82M research paper abstracts are collected from PubMed\footnote{https://www.ncbi.nlm.nih.gov/pubmed} and UMLS\footnote{https://www.nlm.nih.gov/research/umls/} contains existing entity synonym information in the medical domain. The {Wiki + FreeBase} and {PubMed + UMLS} are public available datasets used in previous synonym discovery tasks \cite{qu2017automatic}.
The MedBook is a Chinese dataset collected by authors where we collect 0.51M pieces of contexts from Chinese medical textbooks as well as online medical question answering forums. Synonym entities in the medical domain are obtained from MKG, a medical knowledge graph.
\ref{tab::data_stats} shows the dataset statistics.

\begin{table}[th!]
\centering
\begin{tabular}{llll}
\hline
\textbf{Dataset} & \textbf{Wiki + FreeBase} &\textbf{PubMed + UMLS} & \textbf{MedBooK + MKG} \\\hline
\#ENTITY         & 9274         & 6339          & 32,002           \\
$\quad$\#VALID          & 394          & 386     & 661       \\
$\quad$\#TEST           & 104          & 163           & 468       \\
\#SYNSET         & 4615         & 708           & 6600           \\
\#CONTEXT        & 6,839,331    & 815,644       & 514,226       \\
\#VOCAB          & 472,834      & 1,069,061     & 270,027       \\
\hline
\end{tabular}
\caption{Dataset statistics.}\label{tab::data_stats}
\end{table}

\subsection{Experiment Settings}
\noindent\textbf{Preprocessing}
Wiki +Freebase and PubMed + UMLS come with entities and synonym entity annotations, we adopt the Stanford CoreNLP package to do the tokenization. For MedBook, a Chinese word segmentation tool Jieba\footnote{https://github.com/fxsjy/jieba} is used to segment the corpus into meaningful entities and phrases. We remove redundant contexts in the corpus and filter out entities if they appear in the corpus less than five times. For entity representations, the proposed model works with various unsupervised word embedding methods. Here for simplicity, we adopt skip-gram \cite{mikolov2013distributed} with a dimension of 200. Context window is set as 5 with a negative sampling of 5 words for training.

\noindent\textbf{Evaluation Metric}
For synonym detection using {\ACLModelName} and other alternatives, we train the models with existing synonym and randomly sampled entity pairs as negative samples. During testing, we also sample random entity pairs as negative samples to evaluate the performance. Note that all test synonym entities are from unobserved groups of synonym entities: none of the test entities is observed in the training data. Thus evaluations are done in a completely cold-start setting.

The area under the curve (AUC) and Mean Average Precision (MAP) are used to evaluate the model. AUC is used to measure how well the models assign high scores to synonym entities and low scores to non-synonym entities. An AUC of 1 indicates that there is a clear boundary between scores of synonym entities and non-synonym entities. Additionally, a single-tailed t-test is conducted to evaluate the significance of performance improvements when we compare the proposed {\ACLModelName} model with all the other baselines.

For synonym discovery during the inference phase, we obtain candidate entities $E_{NN}$ from K-nearest neighbors of the query entity in the entity embedding space, and rerank them based on the output score $s(e, e_{NN})$ of the {\ACLModelName} for each $e_{NN} \in E_{NN}$. We expect candidate entities in the top positions are more likely to be synonym with the query entity.
We report the precision at position K (P@K), recall at position K (R@K), and F1 score at position K (F1@K). 

\noindent\textbf{Baselines}
We compare the proposed model with the following alternatives.
\begin{itemize}
    \item \textbf{word2vec} \cite{mikolov2013distributed}: a word embedding approach based on entity representations learned from the skip-gram algorithm. We use the learned word embedding to train a classifier for synonym discovery. A scoring function $Score_D(u,v)=x_u\mathbf{W}x_v^T$ is used as the objective.
    \item \textbf{GloVe} \cite{pennington2014glove}: another word embedding approach. The entity representations are learned based on the GloVe algorithm. The classifier is trained with the same scoring function $Score_D$, but with the learned glove embedding for synonym discovery.
    \item \textbf{SRN} \cite{neculoiu2016learning}: a character-level approach that uses a siamese multi-layer bi-directional recurrent neural networks to encode the entity as a sequence of characters. The hidden states are averaged to get an entity representation. Cosine similarity is used in the objective.
    \item \textbf{MaLSTM} \cite{mueller2016siamese}: another character-level approach. We adopt MaLSTM by feeding the character-level sequence to the model. Unlike SRN that uses Bi-LSTM, MaLSTM uses a single direction LSTM and $l$-1 norm is used to measure the distance between two entities.
    \item \textbf{DPE} \cite{qu2017automatic}: a model that utilizes dependency parsing results as the structured annotation on a single piece of context for synonym discovery.
    \item \textbf{\ACLModelName} is the proposed model, we used siamese loss (\ref{eq::siamese}) and triplet loss (\ref{eq::triplet}) as the objectives, respectively.
\end{itemize}

\subsection{Experiment Results}
We report Area Under the Curve (AUC) and Mean Average Precision (MAP) on three datasets in \ref{tab::overall}.
\begin{table*}[bt!]
\centering
\resizebox{\textwidth}{!}{%
\begin{tabular}{l|cc|cc|cc}\hline
\multirow{2}{*}{\textbf{MODEL}} & \multicolumn{2}{l}{\textbf{Wiki + Freebase}} & \multicolumn{2}{l}{\textbf{PubMed + UMLS}} & \multicolumn{2}{l}{\textbf{MedBook + MKG}} \\\cline{2-3}\cline{4-5}\cline{6-7}
~              & AUC       & MAP      & AUC       & MAP      & AUC       & MAP  \\\hline
word2vec \cite{mikolov2013distributed} &0.9272 & 0.9371 &0.9301 & 0.9422 & 0.9393 & 0.9418 \\
GloVe \cite{pennington2014glove}       &0.9188 & 0.9295 &0.8890 & 0.8869 & 0.7250 & 0.7049 \\
SRN \cite{neculoiu2016learning}        &0.8864 & 0.9134 &0.9517 & 0.9559 & 0.9419 & 0.9545 \\
MaLSTM \cite{mueller2016siamese}       &0.9178 & 0.9413 &0.8151 & 0.8554 & 0.8532   & 0.8833 \\
DPE \cite{qu2017automatic}             &0.9461 & 0.9573 &0.9513 & 0.9623 & 0.9479 & 0.9559  \\
\hline
\textbf{\ACLModelName} (Pairwise)          &0.9831$^\dag$ & 0.9818$^\dag$ & \textbf{0.9838}$^\dag$ & \textbf{0.9872}$^\dag$ & \textbf{0.9685} & \textbf{0.9673}\\
\phantom{00} w/o Leaky Unit             &0.9827$^\dag$ & 0.9817$^\dag$ &0.9815$^\dag$ & 0.9847$^\dag$ & 0.9667 & 0.9651  \\
\phantom{00} w/o Confluence Encoder (Bi-LSTM)   & 0.9683$^\dag$ &0.9625$^\dag$ &0.9495 & 0.9456 & 0.9311& 0.9156 \\\hline
\textbf{\ACLModelName} (Triplet)           &\textbf{0.9877}$^\dag$ & \textbf{0.9892}$^\dag$ &{0.9788}$^\dag$ & 0.9800$^\dag$ & 0.9410 & 0.9230 \\
\phantom{00} w/o Leaky Unit             &0.9705$^\dag$ &0.9631$^\dag$ &0.9779$^\dag$ & {0.9821}$^\dag$ & 0.9359 & 0.9214   \\
\phantom{00} w/o Confluence Encoder (Bi-LSTM)   &0.9582$^\dag$ &0.9531$^\dag$ & 0.9412 & 0.9288 & 0.9047 & 0.8867  \\\hline     
\end{tabular}%
}
\caption{Test performance in AUC and MAP on three datasets.}
\label{tab::overall}
\end{table*}
\begin{figure*}[tb!]
    \centering
    \includegraphics[width=0.32\linewidth]{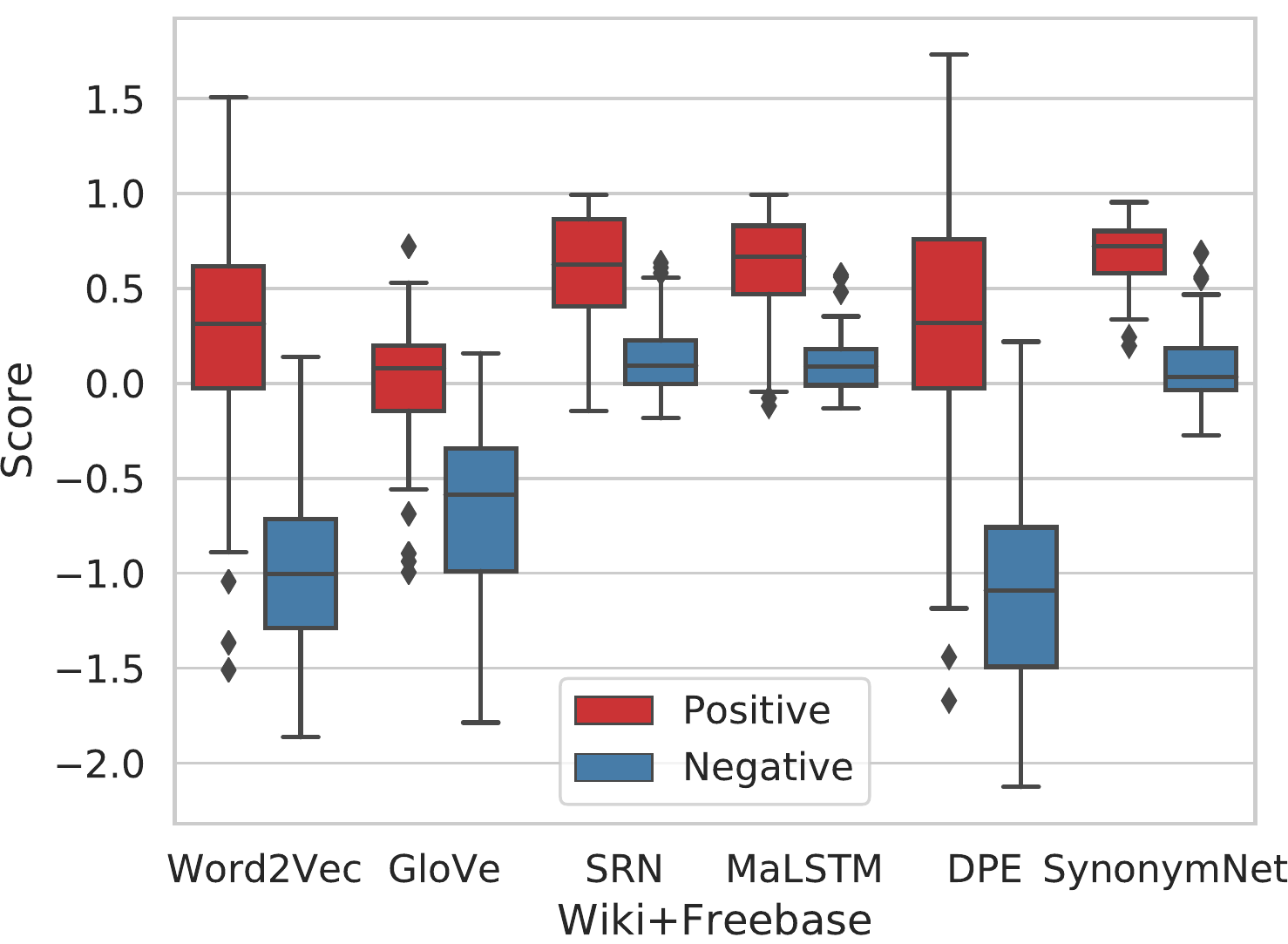}
     \includegraphics[width=0.32\linewidth]{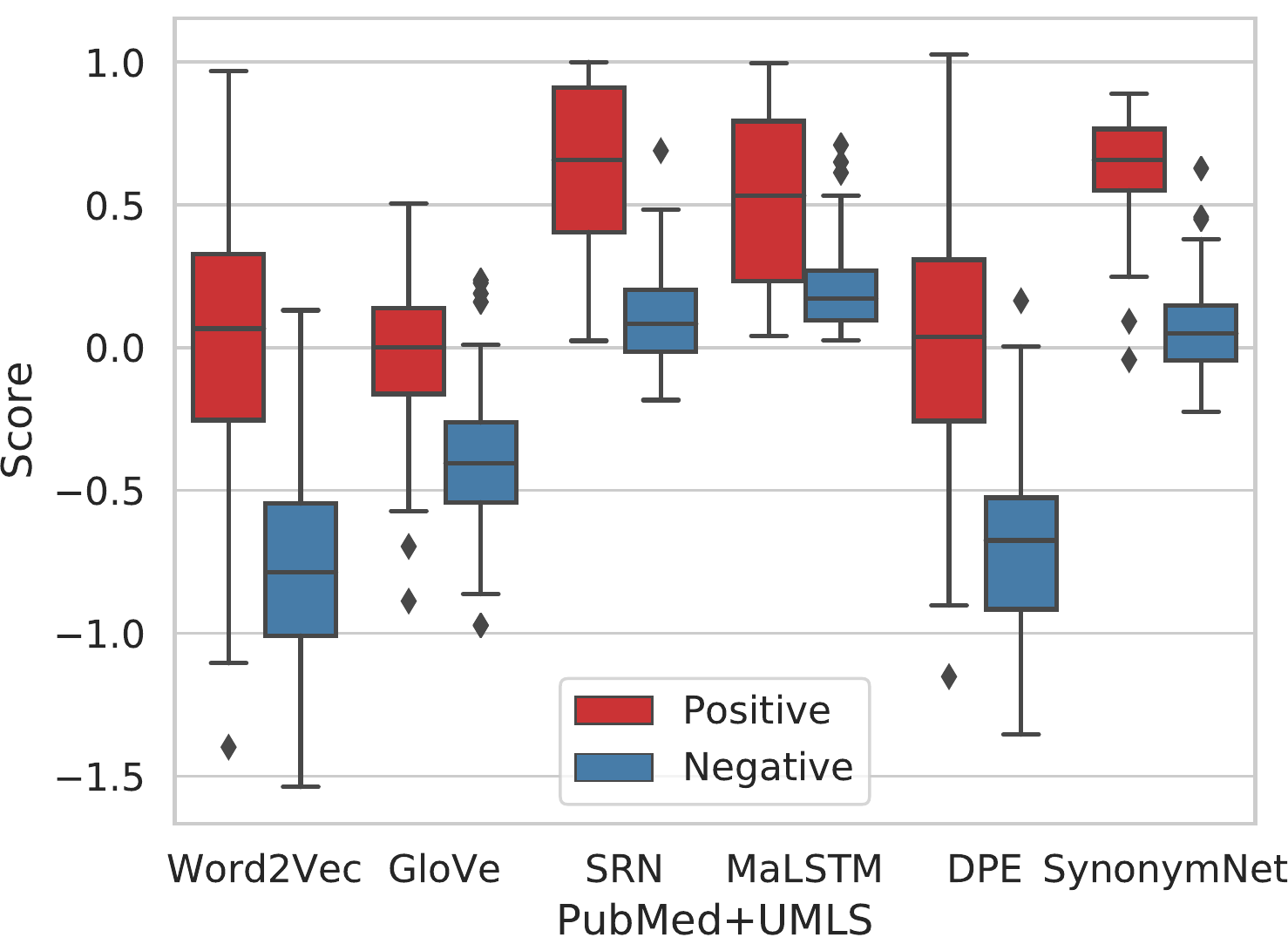}
     \includegraphics[width=0.32\linewidth]{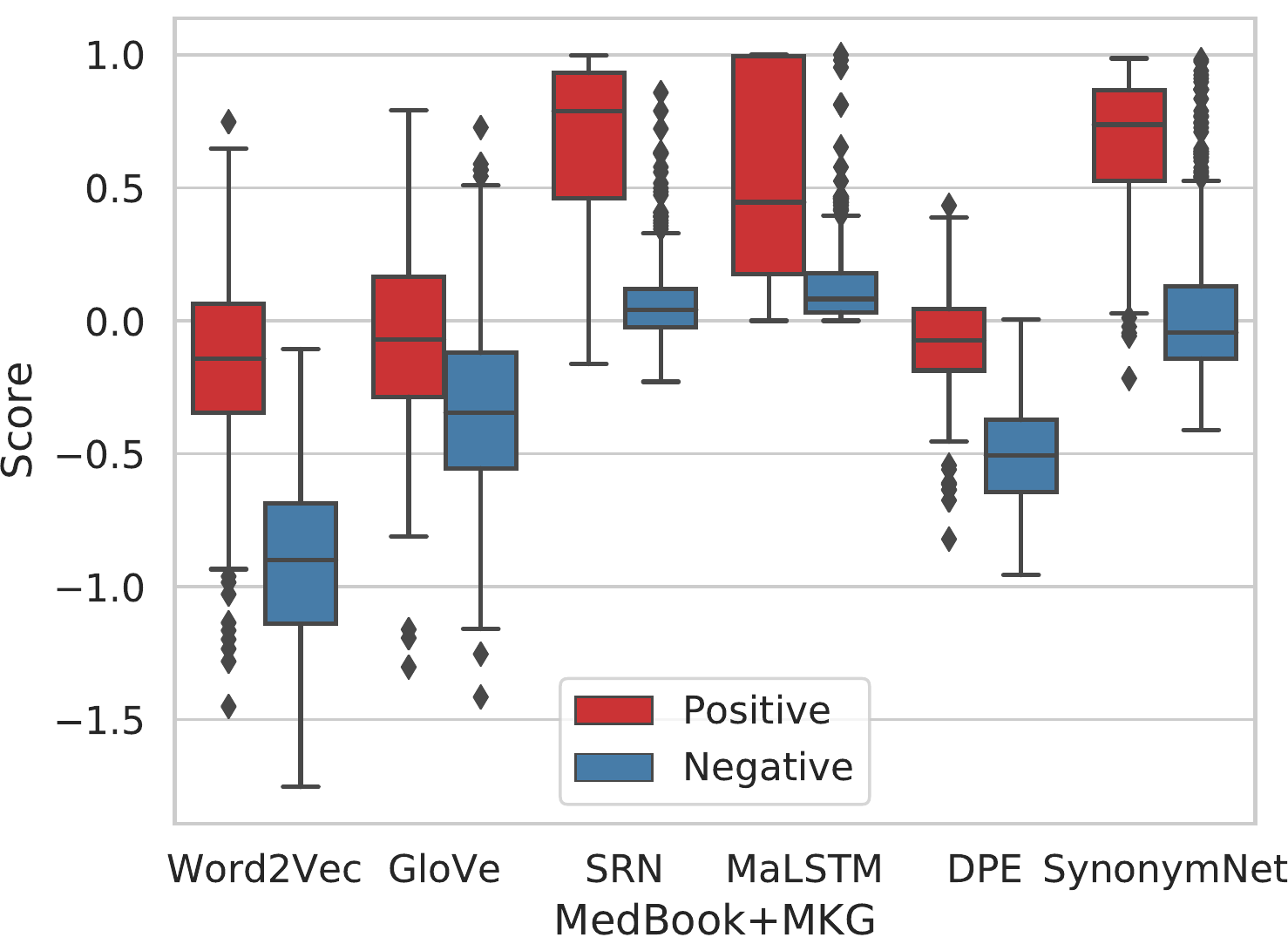}
    \caption{Test synonym score distributions on positive and negative entity pairs.}\label{fig::score_dist}
\end{figure*}

From the upper part of \ref{tab::overall} we can see that {\ACLModelName} performances consistently better than other baselines on three datasets. {\ACLModelName} with the triplet training objective achieves the best performance on Wiki +Freebase, while the Siamese objective works better on PubMed + UMLS and MedBook + MKG. Word2vec is generally performing better than GloVe. SRNs achieve decent performance on PubMed + UMLS and MedBook + MKG. This is probably because the synonym entities obtained from the medical domain tend to share more character-level similarities, such as \texttt{6-aminohexanoic acid} and \texttt{aminocaproic acid}. However, even if the character-level features are not explicitly used in our model, our model still performances better, by exploiting multiple pieces of contexts effectively. DPE has the best performance among other baselines, by annotating each piece of context with dependency parsing results. However, the dependency parsing results could be error-prone for the synonym discovery task, especially when two entities share the similar usage but with different semantics, such as \texttt{NBA finals} and \texttt{NFL playoffs}.

We conduct statistical significance tests to validate the performance improvement. The single-tailed t-test is performed for all experiments, which measures whether or not the results from the proposed model are significantly better than ones from baselines. The numbers with $\dag$ markers in \ref{tab::overall} indicate that the improvement is significant with p$<$0.05.

\ref{tab::discovery} reports the performance in P$@$K, R$@$K, and F1$@$K.
\begin{table*}[htbp!]
\centering
\begin{tabular}{l|lll|lll|lll}
\hline
~ & \multicolumn{3}{l}{\textbf{Wiki + Freebase}} & \multicolumn{3}{l}{\textbf{PubMed + UMLS}} & \multicolumn{3}{l}{\textbf{MedBook + MedKG}} \\ \hline
                 & P@K    & R@K    & F1@K   & P@K    & R@K    & F1@K   & P@K     & R@K     & F1@K    \\\hline
K=1              &  0.3455      & 0.3455       &0.3455        &  0.2400 & 0.0867 & 0.1253 & 0.3051 &	0.2294	& 0.2486 \\
K=5              &  0.1818      & 0.9091       &0.3030        &  0.2880 & 0.7967	& 0.3949  &   0.2388 &	0.8735 &	0.3536    \\
K=10             & 0.1000       & 1.0000       & 0.1818       &   0.1800     &  1.0000  & 0.2915       &  0.1418 & 1.0000 &	0.2360   \\\hline
\end{tabular}
\caption{Performance on Synonym Discovery.}\label{tab::discovery}
\end{table*}

Besides numeric metrics, we also use box plots to represent the score distributions for each method on all three datasets in \ref{fig::score_dist}. The red bars indicate scores on positive entity pairs that are synonym with each other, while the blue bars indicate scores on negative entity pairs. A general conclusion is that our model assigns higher scores for synonym entity pairs, marginally higher than other non-synonym entity pairs when compared with other alternatives.

\subsection{Ablation Study}
To study the contribution of different modules of {\ACLModelName} for synonym discovery, we also report ablation test results in the lower part of \ref{tab::overall}. ``w/o  Confluence Context Encoder" uses the Bi-LSTM as the context encoder. The last hidden states in both forward and backward directions in Bi-LSTM are concatenated; ``w/o Leaky Unit" does not have the ability to ignore uninformative contexts during the bilateral matching process: all contexts retrieved based on the entity, whether informative or not, are utilized in bilateral matching.
From the lower part of \ref{tab::overall} we can see that both modules (Leaky Unit and Confluence Encoder) contribute to the effectiveness of the model. The leaky unit contributes 1.72\% improvement in AUC and 2.61\% improvement in MAP on the Wiki dataset when trained with the triplet objective. The Confluence Encoder gives the model an average of 3.17\% improvement in AUC on all three datasets, and up to 5.17\% improvement in MAP. 

\subsection{Hyperparameters}
We train the proposed model with a wide range of hyperparameter configurations, which are listed in \ref{tab::hyperparameters}.
For the model architecture, we vary the number of randomly sampled contexts $P=Q$ for each entity from 1 to 20. 
Each piece of context is chunked by a maximum length of $T$. For the confluence context encoder, we vary the hidden dimension $d_{CE}$ from 8 to 1024. The margin value $m$ in triplet loss function is varied from 0.1 to 1.75.  
For the training, we try different optimizers (Adam \cite{kingma2014adam}, RMSProp \cite{tieleman2012lecture}, adadelta \cite{zeiler2012adadelta}, and Adagrad \cite{duchi2011adaptive}), with the learning rate varying from 0.0003 to 0.01. Different batch sizes are used to train the model. We apply random search to obtain the best-performing hyperparameter setting on the validation split for each dataset, as shown in \ref{tab::hyper_use}.

\begin{table}[h!]
\centering
\begin{tabular}{l|l}\hline
\textbf{HYPERPARAMETERS} & \textbf{VALUE} \\\hline
$P$ (context number) & \{1, 3, 5, 10, 15, 20\} \\
$T$ (maximum context length) & \{10, 30, 50, 80\} \\
$d_{CE}$ (layer size)     & \{8, 16, 32, 64, 128, 256, 512, 1024\}  \\
$m$ (margin)         & \{0.1, 0.25, 0.5, 0.75, 1.25, 1.5, 1.75\}  \\
Optimizer & \{Adam, RMSProp, Adadelta, Adagrad\}\\
Batch Size & \{4, 8, 16, 32, 64, 128\}\\
Learning Rate & \{0.0003, 0.0001, 0.001, 0.01\}\\
\hline
\end{tabular}
\caption{Hyperparameter settings.}\label{tab::hyperparameters}
\end{table}
\begin{table*}[hbt!]
\centering
\begin{tabular}{l|lllllll}\hline
        \textbf{DATASETS} & $P$ & $T$ & $d_{CE}$ & $m$ & Optimizer & Batch Size & Learning Rate \\\hline
        {Wiki + Freebase} & 20 & 50 & 256 & 0.75 & Adam & 16 & 0.0003 \\
        {PubMed + UMLS}   & 20 & 50 & 512 & 0.5 & Adam & 16 & 0.0003 \\
        {MedBook + MKG} & 5 & 80 & 256 & 0.75 & Adam & 16 & 0.0001 \\\hline
    \end{tabular}
    \caption{Hyperparameters.}
    \label{tab::hyper_use}
\end{table*}

\begin{figure*}[t!]
    \centering
    \minipage{0.29\textwidth}
    \includegraphics[width=\textwidth]{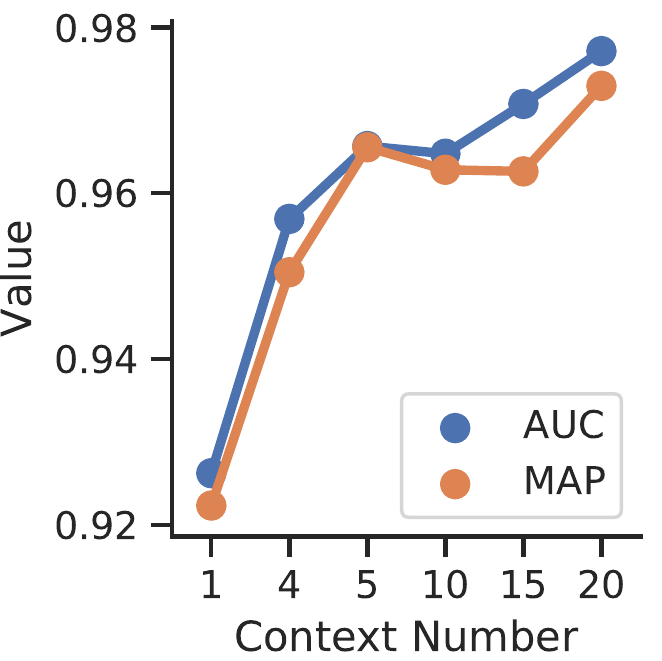}
    \endminipage\hfill
    \minipage{0.29\textwidth} 
    \includegraphics[width=\textwidth]{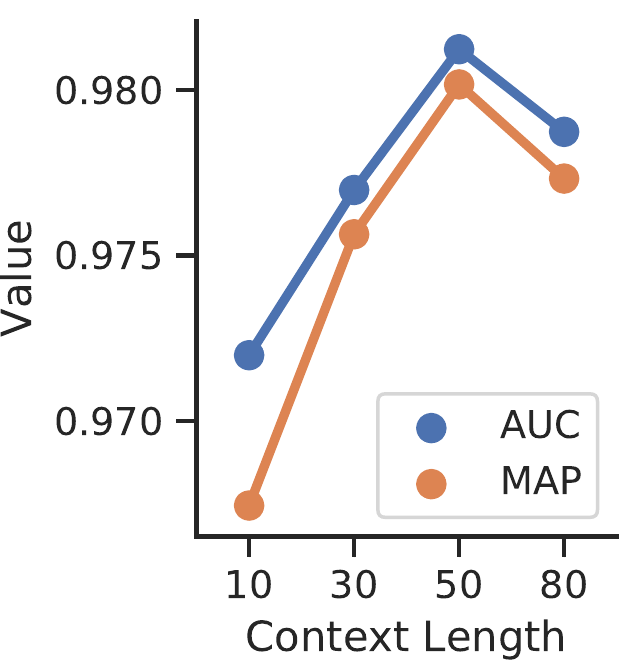}
    \endminipage\hfill
    \minipage{0.4\textwidth}
     \includegraphics[width=\textwidth]{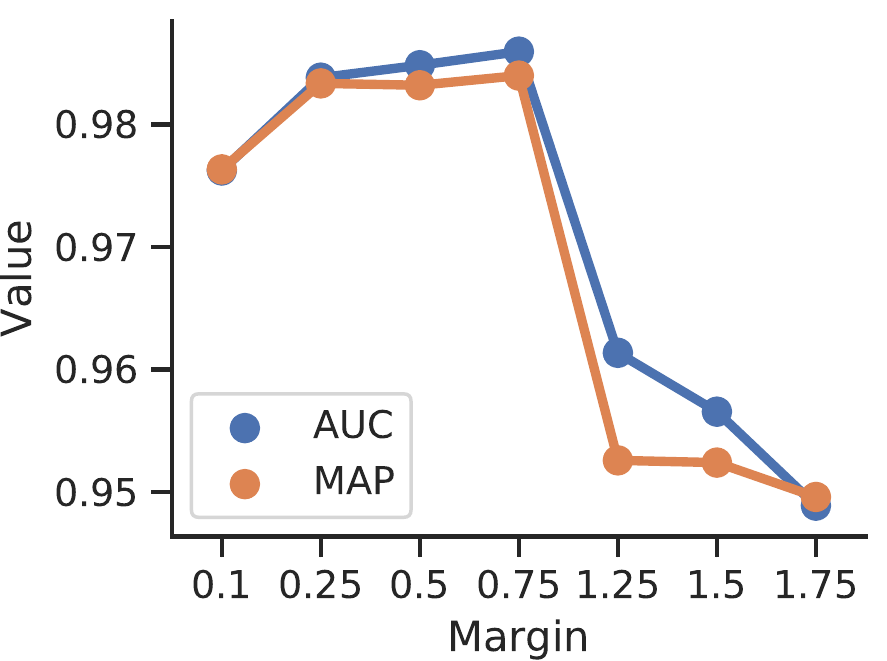}
    \endminipage\hfill
    \caption{Sensitivity analysis.}\label{fig::sensitivity}
\end{figure*}

Furthermore, we provide sensitivity analysis of the proposed model with different hyperparameters in Wiki + Freebase dataset in \ref{fig::sensitivity}. \ref{fig::sensitivity} shows the performance curves when we vary one hyperparameter while keeping the remaining fixed. As the number of contexts $P$ increases, the model generally performs better. Due to limitations on computing resources, we are only able to verify the performance of up to 20 pieces of randomly sampled contexts. The model achieves the best AUC and MAP when the maximum context length $T=50$: longer contexts may introduce too much noise while shorter contexts may be less informative.

\subsection{Case Studies}\label{appendix::cases}
\ref{tab::candidate} and \ref{tab::case} show a case for entity \texttt{UNGA}. The candidate entities in \ref{tab::candidate} are generated with pretrained word embedding using skip-gram. \ref{tab::case} shows the discovered synonym entities by the proposed {\ACLModelName} model, where a threshold of 0.8 on the {\ACLModelName} score is used.
\begin{table}[th!]
\centering
\begin{tabular}{ll}\hline
Candidate Entities & Cosine Similarity \\\hline
united\_nations\_general\_assembly$||$m.07vp7$||$ &0.847374\\
un\_human\_rights\_council &0.823727\\
the\_united\_nations\_general\_assembly &0.813736\\
un\_security\_council$||$m.07vnr$||$& 0.794973\\
palestine\_national\_council &0.791135\\
world\_health\_assembly$||$m.05\_gl9$||$ &0.790837\\ 
united\_nations\_security\_council$||$m.07vnr$||$ &0.787999\\
general\_assembly\_resolution& 0.784581\\
the\_un\_security\_council &0.784280\\
ctbt &0.777627\\
north\_atlantic\_council$||$m.05pmgy$||$ &0.775703\\
resolution\_1441 &0.773064\\
non-binding\_resolution$||$m.02pj22f$||$& 0.771475\\
unga$||$m.07vp7$||$& 0.770623\\ \hline                                             
\end{tabular}
\caption{Candidate entities retrieved for \texttt{UNGA}.}\label{tab::candidate}
\end{table}
\begin{table}[th!]
\centering
\begin{tabular}{ll}\hline
Final Entities & {\ACLModelName} Score \\\hline
united\_nations\_general\_assembly$||$m.07vp7$||$ & 0.842602\\
the\_united\_nations\_general\_assembly&  0.801745\\
unga$||$m.07vp7$||$&  0.800719\\ \hline
\end{tabular}
\caption{Discovered synonym entities for \texttt{UNGA} using {\ACLModelName}.}\label{tab::case}
\end{table}
\section{Related works}
\noindent\textbf{Synonym Discovery}
The synonym discovery focuses on detecting entity synonyms.
Most existing works try to achieve this goal by learning from structured information such as query logs \cite{ren2015synonym,chaudhuri2009exploiting,wei2009context}. While in this work, we focus on synonym discovery from free-text natural language contexts, which requires less annotation and is more challenging.

Some existing works try to detect entity synonyms by entity-level similarities \cite{lin2003identifying,roller2014inclusive,neculoiu2016learning,wieting2016charagram}. For example, \cite{roller2014inclusive} introduce distributional features for hypernym detection. \cite{neculoiu2016learning} use a Siamese structure that treats each entity as a sequence of characters, and uses a Bi-LSTM to encode the entity information. Such approach may be helpful for synonyms with similar spellings, or dealing with abbreviations. Without considering the context information, it is hard for the aforementioned methods to infer synonyms that share similar semantics but are not alike verbatim, such as \texttt{JD} and \texttt{law degree}.

Various approaches \cite{snow2005learning,sun2010semi,liao2017deep,cambria2018senticnet} are proposed to incorporate context information to characterize entity mentions. However, these models are not designed for synonym discovery.
\cite{qu2017automatic} utilize additional structured annotations, e.g. dependency parsing result, as the context of the entity for synonym discovery. 
While we aim to discover synonym entities from a free-text corpus without structured annotation. 

\noindent\textbf{Sentence Matching}
There is another related research area that studies sentence matching. Early works try to learn a meaningful single vector to represent the sentence \cite{tan2015lstm,mueller2016siamese}. These models do not consider the word-level interactions from two sentences during the matching. 
\cite{wang2016compare,wang2016multi,wang2017bilateral} introduce multiple instances for matching with varying granularities.
Although the above methods achieve decent performance on sentence-level matching, the sentence matching task is different from context modeling for synonym discovery in essence. Context matching focuses on local information, especially the words before and after the entity word; while the overall sentence could contain much more information, which is useful to represent the sentence-level semantics, but can be quite noisy for context modeling. We adopt a confluence encoder to model the context, which is able to aware of the location of an entity in the context while preserving information flow from both left and right contexts.

Moreover, sentence matching models do not explicitly deal with uninformative instances: max-pooling strategy and attention mechanism are introduced. The max-pooling strategy picks the most informative one and ignores all the other less informative ones. In context matching, such property could be unsatisfactory as an entity is usually associated with multiple contexts. We adopt a bilateral matching which involves a leaky unit to explicitly deal with uninformative contexts, so as to eliminate noisy contexts while preserving the expression diversity from multiple pieces of contexts.

\chapter{Conclusion}\label{chaper:conclusion}
(Part of the chapter was previously published in \cite{zhang2016mining,zhang2017bringing,zhang2018generative,zhang2018joint,zhang2018synonymnet}.)
In this dissertation, we have explored the structured knowledge discovery from the massive text corpus. More specifically, two general and strongly correlated research objectives are explored: one is to harness structured information for natural language understanding and modeling, and the other objective is to effectively expand and refine structured knowledge harnessing the massiveness of the text corpus. We thoroughly studied four different research problems: \TaskNameOne, \TaskNameTwo, {\TaskNameThree}, and \TaskNameFour. We have evaluated the effectiveness of the proposed approaches on various user-generated text corpora such as the question-answering corpus, web search queries, voice commands, and documents by extensive quantitative experiments and case studies. The main contributions of our works are summarized as follows:

\begin{itemize}
    \item 
    We studied the Structured Intent Detection problem that aims to understand complicated user intentions in online question-answering discussion forums. An Intent Graph is formulated to possess explicit constraints on concept mentions as nodes and semantic transitions among concepts as directed edges on the Intent Graph, which are key components to characterize Structured Intents. A neural network model named coCTI-MTL based on multi-task learning is introduced to extract concept mentions as well as semantic transitions collectively as a sub-graph of the Intent Graph to represent Structured Intents. Empirical results show that the proposed method can accurately detect complicated user intents from real-world information-seeking text corpora generated by users on an online medical question-answering discussion forum. Being able to detect complicated intents may further benefit other tasks such as dialogue management, recommendation, and question rewriting.
    
    \item 
    We presented a capsule neural network based model, namely {\ModelName}, to harness the hierarchical relationships among words, slots, and intents in the utterance for joint slot filling and intent detection. Unlike treating slot filling as a sequential prediction problem, the proposed model {\ModelName} assigns each word to its most appropriate slots in {\SecondCapsule} by a dynamic routing-by-agreement schema. The learned word-level slot representations are further aggregated to get the utterance-level intent representations via dynamic routing-by-agreement. A re-routing schema is proposed to further synergize the slot filling performance using the inferred intent representation. Experiments on two real-world datasets show the effectiveness of the proposed models when compared with other alternatives as well as existing NLU services.
    
    \item We introduce a generative perspective to study the {\KDDTaskNameFull} problem, which aims to expand the scale of high-quality yet novel structured knowledge from the massive text corpus with minimized annotation and additional data collection. We propose a model named {\KDDModelNameFull} (\KDDModelName) which capitalizes on rich semantic information learned unsupervisely from a large text corpus as entity representations. The proposed model defines each relationship by solely learning the expression commonalities and differences from existing entity pairs that are diversely expressed. It generates meaningful, novel entity pairs of a specific relationship by directly sampling from the learned latent space without the requirement of additional context information. The performance of the proposed method is evaluated on real-world data both quantitatively and qualitatively.
    
    \item 
    We developed a framework for synonym discovery from the text corpus without structured annotation. A novel neural network model {\ACLModelName} is introduced for synonym detection, which tries to determine whether or not two given entities are synonym with each other. The proposed model is able to automatically detect synonym entities from a large corpus, which could help remove duplicate entities in knowledge graphs and thus improve the quality of structured knowledge. {\ACLModelName} makes use of multiple pieces of contexts in which each entity is mentioned, and compares the context-level similarity via a bilateral matching schema to determine synonymity. Experiments on three real-world datasets show that the proposed method {\ACLModelName} can discover synonym entities effectively on both generic datasets (Wiki+Freebase in English), as well as domain-specific datasets (PubMed+UMLS in English and MedBook+MKG in Chinese) with an improvement up to 4.16\% in AUC and 3.19\% in MAP. 
\end{itemize}

\appendices
\newpage
\appendix
\section{ACM Copyright Letter}
``Authors can reuse any portion of their own work in a new work of their own (and no fee is expected) as long as a citation and DOI pointer to the Version of Record in the ACM Digital Library are included.

Contributing complete papers to any edited collection of reprints for which the author is not the editor, requires permission and usually a republication fee.

Authors can include partial or complete papers of their own (and no fee is expected) in a dissertation as long as citations and DOI pointers to the Versions of Record in the ACM Digital Library are included. Authors can use any portion of their own work in presentations and in the classroom (and no fee is expected).'' \footnote{\url{http://authors.acm.org/main.html}}

\section{IEEE Copyright Letter}
\epsfig{file=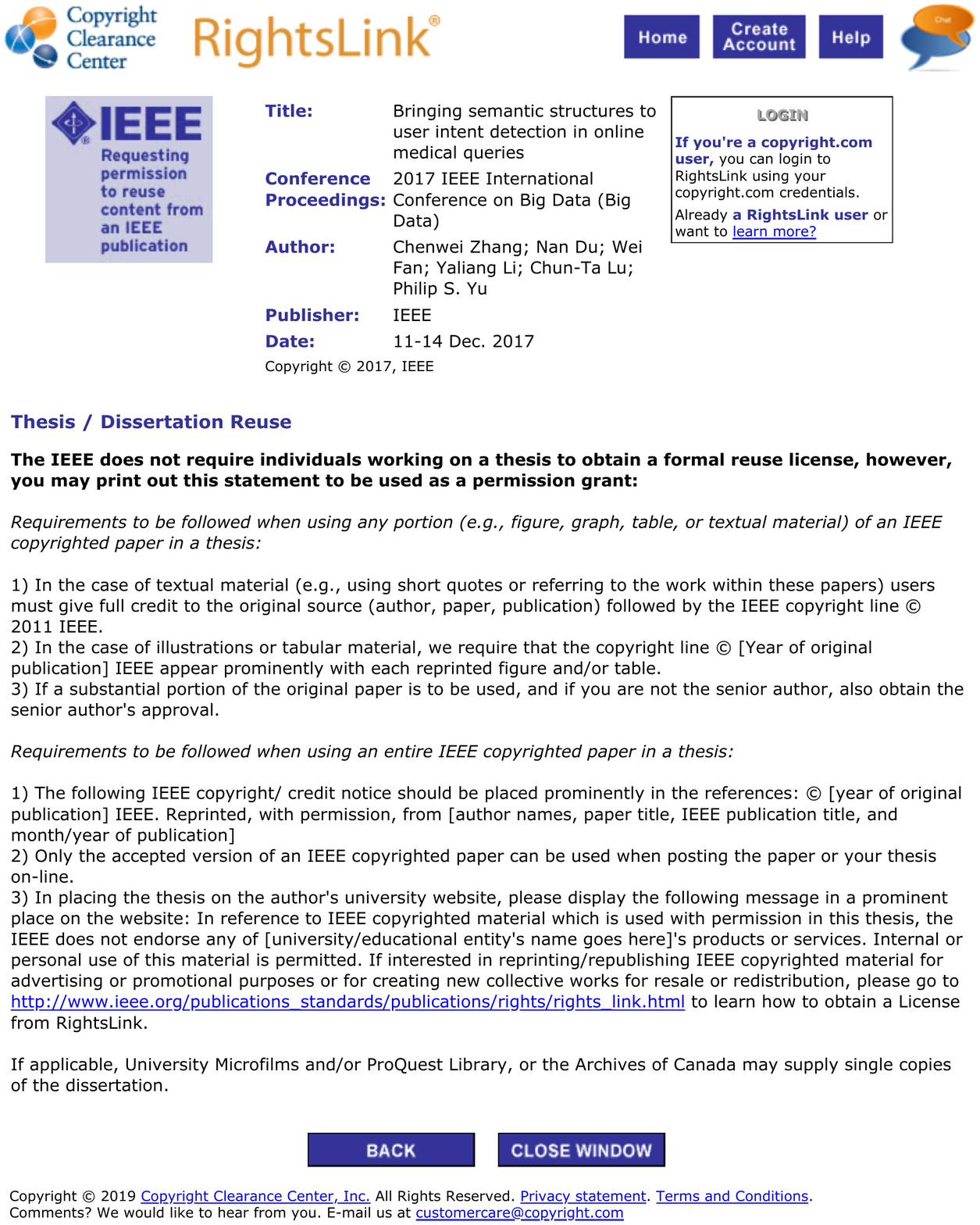, width=0.9\textwidth}

\section{arXiv.org Copyright Letter}
\epsfig{file=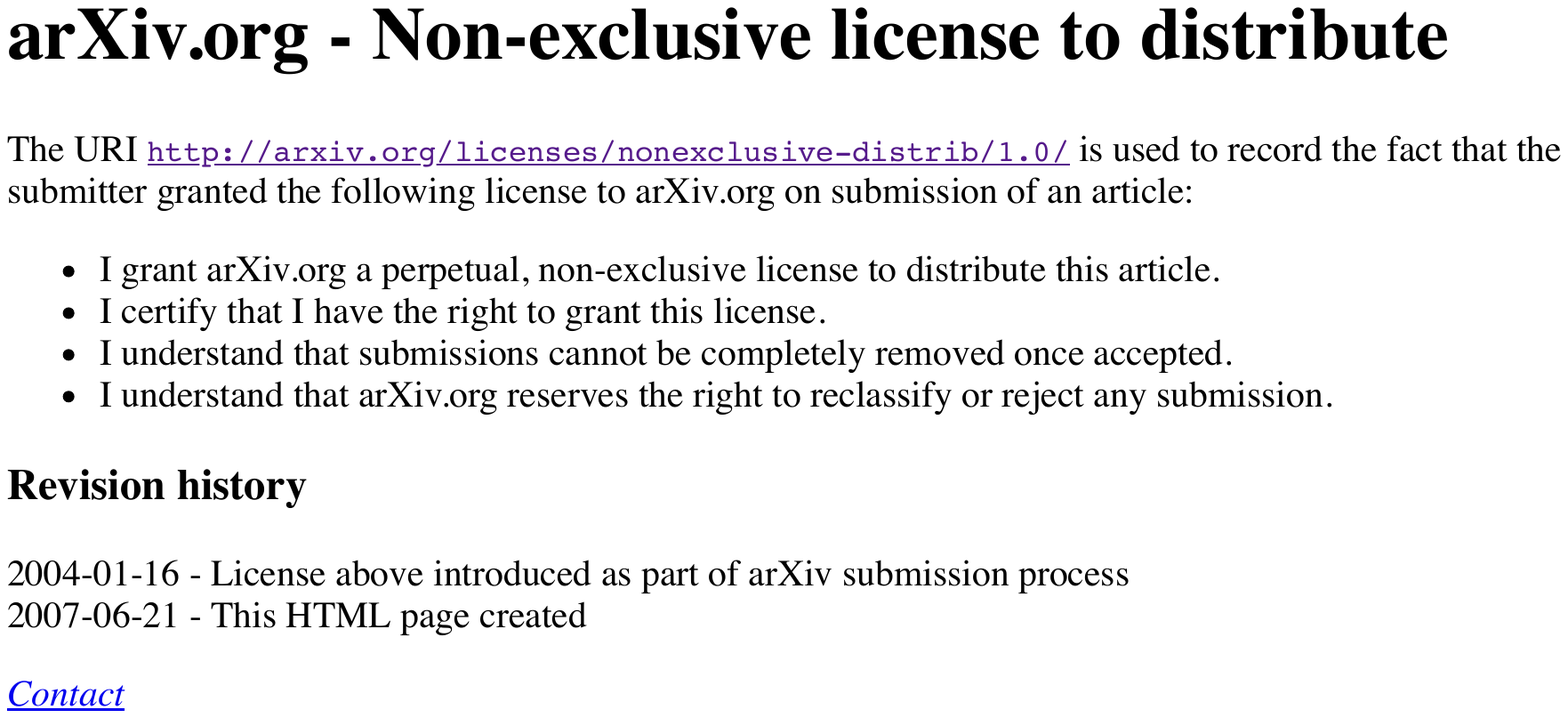, width=0.9\textwidth}

\section{ACL Copyright Letter}
\epsfig{file=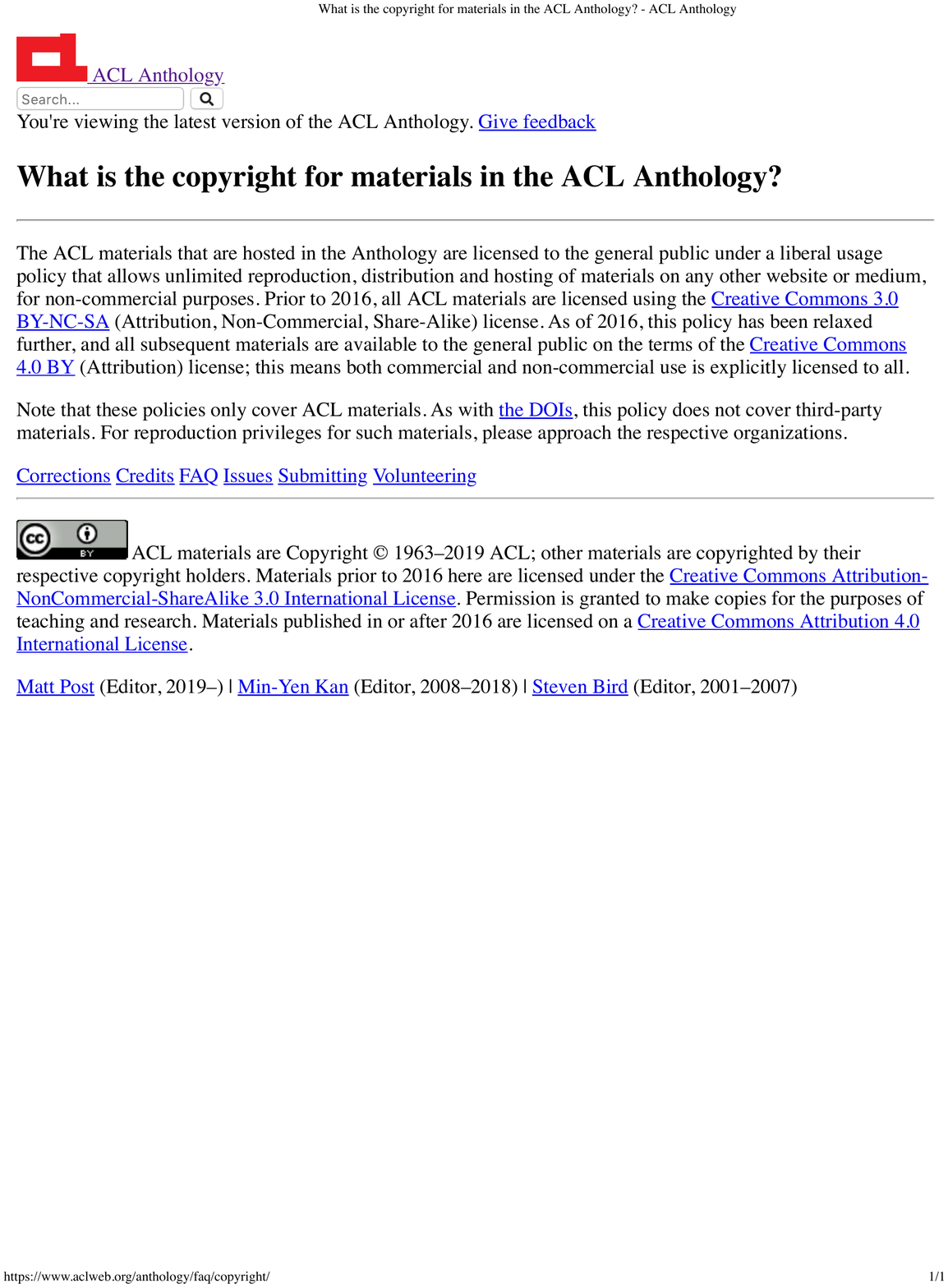, width=0.9\textwidth}

\bibforma
\bibliography{thesis}
\newpage
\vita
\textbf{Name:} Chenwei Zhang

\textbf{EDUCATION:}

~~~~~~\textbf{B.Eng. in Computer Science and Technology}, Southwest University, 2014.

\textbf{PUBLICATIONS:}
\begin{itemize}
\item \underline{Chenwei Zhang}, Yaliang Li, Nan Du, Wei Fan, Philip S. Yu. \textit{Joint Slot Filling and Intent Detection via Capsule Neural Networks}. In Proceedings of the 57th Annual Meeting of the Association for Computational Linguistics (ACL), 2019.
\item Congying Xia, \underline{Chenwei Zhang}, Tao Yang, Yaliang Li, Nan Du, Xian Wu, Wei Fan, Fenglong Ma and Philip Yu. \textit{Multi-grained Named Entity Recognition}. In Proceedings of the 57th Annual Meeting of the Association for Computational Linguistics (ACL), 2019.
\item Jiawei Zhang, \underline{Chenwei Zhang}, Bowen Dong, Yang Yang, Philip S. Yu. \textit{Missing Movie Synergistic Completion across Multiple Isomeric Online Movie Knowledge Libraries}. In Proceedings of the International Joint Conference on Neural Networks (IJCNN), 2019.
\item Fenglong Ma, Yaliang Li, \underline{Chenwei Zhang}, Jing Gao, Nan Du, Wei Fan. \textit{MCVAE: Margin-based Conditional Variational Autoencoder for Relation Classification and Pattern Generation}. In Proceedings of the 2019 World Wide Web Conference (WWW), 2019.
\item \underline{Chenwei Zhang}, Yaliang Li, Nan Du, Wei Fan, Philip S. Yu. \textit{On the Generative Discovery of Structured Medical Knowledge}. In Proceedings of the 24th ACM SIGKDD International Conference on Knowledge Discovery and Data Mining (KDD), 2018.
\item \underline{Chenwei Zhang}, Yaliang Li, Nan Du, Wei Fan, Philip S. Yu. \textit{SynonymNet: Multi-context Bilateral Matching for Entity Synonyms}. arXiv, 2018.
\item Congying Xia*, \underline{Chenwei Zhang}*, Xiaohui Yan, Yi Chang, Philip S. Yu. \textit{Zero-shot User Intent Detection via Capsule Neural Networks}. In Proceedings of the 2018 Conference on Empirical Methods in Natural Language Processing (EMNLP), 2018. (* equally contributed)
\item Shaika Chowdhury, \underline{Chenwei Zhang}, Philip S. Yu. \textit{Multi-Task Pharmacovigilance Mining from Social Media Posts}. In Proceedings of the 27th edition of The Web Conference (WWW), 2018.
\item Zhang-Meng Liu, \underline{Chenwei Zhang}, Philip S. Yu. \textit{Direction-of-Arrival Estimation based on Deep Neural Networks with Robustness to Array Imperfections}. In the IEEE Transactions on Antennas and Propagation, 2018.
\item Yue Wang, \underline{Chenwei Zhang}, Shen Wang, Philip S. Yu, Lu Bai, Lixin Cui. \textit{Market Abnormality Period Detection via Co-movement Attention Model}. In Proceedings of the IEEE International Conference on Big Data (Big Data), 2018.
\item Ye Liu, Jiawei Zhang, \underline{Chenwei Zhang}, Philip S. Yu. \textit{Data-driven Blockbuster Planning on Online Movie Knowledge Library}. In Proceedings of the IEEE International Conference on Big Data (Big Data), 2018.
\item Yue Wang, \underline{Chenwei Zhang}, Shen Wang, Philip S. Yu, Lu Bai, Lixin Cui. \textit{Deep Co-investment Network Learning for Financial Assets}. arXiv, 2018.
\item Yaliang Li, Liuyi Yao, Nan Du, Jing Gao, Qi Li, Chuishi Meng, \underline{Chenwei Zhang}, Wei Fan. \textit{Finding Similar Medical Questions from Question Answering Websites}. arXiv, 2018.
\item \underline{Chenwei Zhang}, Wei Fan, Nan Du, Yaliang Li, Chun-Ta Lu, and Philip S. Yu. \textit{Bringing Semantic Structures to User Intent Detection in Online Medical Queries}. In Proceedings of the IEEE International Conference on Big Data (Big Data), 2017.
\item Bokai Cao, Lei Zheng, \underline{Chenwei Zhang}, Philip S. Yu, Andrea Piscitello, John Zulueta, Olu Ajilore, Kelly Ryan and Alex Leow. \textit{DeepMood: Modeling Mobile Phone Typing Dynamics for Mood Detection}. In Proceedings of the 22nd ACM SIGKDD International Conference on Knowledge Discovery and Data Mining (KDD), 2017.
\item Jiawei Zhang, Congying Xia, \underline{Chenwei Zhang}, Limeng Cui, Yanjie Fu, and Philip S Yu. \textit{BL-MNE: Emerging Heterogeneous Social Network Embedding through Broad Learning with Aligned Autoencoder}. In Proceeding of the IEEE International Conference on Data Mining (ICDM), 2017.
\item Junxing Zhu, Jiawei Zhang, Lifang He, Quanyuan Wu, Bin Zhou, \underline{Chenwei Zhang} and Philip S. Yu. \textit{Broad Learning based Multi-Source Collaborative Recommendation}. In Proceedings of the 26th ACM International Conference on Information and Knowledge Management (CIKM), 2017.
\item Junxing Zhu, Jiawei Zhang, \underline{Chenwei Zhang}, Quanyuan Wu, Yan Jia, Bin Zhou, and Philip S. Yu. \textit{CHRS: Cold Start Recommendation across Multiple Heterogeneous Information Networks}. IEEE Access (2017).
\item \underline{Chenwei Zhang}, Wei Fan, Nan Du and Philip S. Yu. \textit{Mining User Intentions from Medical Queries: A Neural Network Based Heterogeneous Jointly Modeling Approach}. In Proceedings of the 25th International World Wide Web Conference (WWW), 2016.
\item \underline{Chenwei Zhang}, Sihong Xie, Yaliang Li, Jing Gao, Wei Fan and Philip S. Yu. \textit{Multi-source Hierarchical Prediction Consolidation}. In Proceedings of the 25th ACM International Conference on Information and Knowledge Management (CIKM), 2016.
\item Chaochun Liu, Huan Sun, Nan Du, Shulong Tan, Hongliang Fei, Wei Fan, Tao Yang, Hao Wu, Yaliang Li, and \underline{Chenwei Zhang}. \textit{An Augmented LSTM Framework to Construct Medical Self-diagnosis Android}. In Proceeding of the IEEE International Conference on Data Mining (ICDM), 2016.  
\item Yaliang Li, Chaochun Liu, Nan Du, Wei Fan, Qi Li, Jing Gao, \underline{Chenwei Zhang}, and Hao Wu. \textit{Extracting Medical Knowledge from Crowdsourced Question Answering Website}. IEEE Transactions on Big Data (2016).  
\item \underline{Chenwei Zhang}, Xiaoyan Su, Yong Hu, Zili Zhang, Yong Deng. \textit{An Evidential Spam-Filtering Framework}. Cybernetics and Systems (2016): 1-18.
\item Hongping Wang, Xi Lu, Yuxian Du, \underline{Chenwei Zhang}, Rehan Sadiq, Yong Deng. \textit{Fault Tree Analysis Based on TOPSIS and Triangular Fuzzy Number}. International Journal of System Assurance Engineering and Management (2014): 1-7.
\item  \underline{Chenwei Zhang}, Yong Hu, Felix T. S. Chan, Rehan Sadiq and Yong Deng. \textit{A New Method to Determine Basic Probability Assignment Using Core Samples}. Knowledge-Based Systems~(2014): 140-149.
\end{itemize}

\end{document}